\documentclass[10pt,twocolumn,letterpaper]{article}

\usepackage{cvpr}              %
\usepackage{caption}
\usepackage{graphicx}
\usepackage{amsmath}
\usepackage{amssymb}
\usepackage{booktabs}
\usepackage{multirow}
\usepackage{pifont}%
\newcommand{\cmark}{\ding{51}}%
\newcommand{\xmark}{\ding{55}}%

\usepackage[pagebackref,breaklinks,colorlinks]{hyperref}

\usepackage[capitalize]{cleveref}
\crefname{section}{Sec.}{Secs.}
\Crefname{section}{Section}{Sections}
\Crefname{table}{Table}{Tables}
\crefname{table}{Tab.}{Tabs.}

\def\cvprPaperID{10878} %
\def\confName{CVPR}
\def\confYear{2023}

\begin{document}

\title{Open Vocabulary Semantic Segmentation with Patch Aligned Contrastive Learning}

\author{Jishnu Mukhoti\thanks{Preliminary work. Contact: jishnu.mukhoti@eng.ox.ac.uk}\\
University of Oxford\\
Meta AI \\
\and
Tsung-Yu Lin\\
Meta AI\\
\and
Omid Poursaeed\\
Meta AI\\
\and
Rui Wang\\
Meta AI\\
\and
Ashish Shah\\
Meta AI\\
\and
Philip H.S. Torr\\
University of Oxford\\
\and
Ser-Nam Lim\\
Meta AI
}

\maketitle

\begin{abstract}
 We introduce \emph{Patch Aligned Contrastive Learning (PACL)}, a modified compatibility function for CLIP's contrastive loss, intending to train an alignment between the patch tokens of the vision encoder and the CLS token of the text encoder. With such an alignment, a model can identify regions of an image corresponding to a given text input, and therefore transfer seamlessly to the task of open vocabulary semantic segmentation without requiring any segmentation annotations during training. Using pre-trained CLIP encoders with PACL, we are able to set the state-of-the-art on the task of open vocabulary zero-shot segmentation on 4 different segmentation benchmarks: Pascal VOC, Pascal Context, COCO Stuff and ADE20K. Furthermore, we show that PACL is also applicable to image-level predictions and when used with a CLIP backbone, provides a general improvement in zero-shot classification accuracy compared to CLIP, across a suite of 12 image classification datasets.
\end{abstract}

\section{Introduction}
\label{sec:intro}

Understanding the semantic content in visual scenes has been one of the most important problems studied in computer vision at various levels of granularity. Work on this problem has led to significant improvements along several threads including image level predictions like image classification \cite{yu2022coca, wortsman2022model, dai2021coatnet}, object level predictions within an image like object detection  \cite{wei2022contrastive, wang2022image, zhang2022dino, liu2022swin, yuan2021florence, zhang2022glipv2}, as well as pixel level predictions in an image like semantic segmentation \cite{wang2022image, wei2022contrastive, li2022mask, chen2022vision}. Although in image classification we require only a single label per image for prediction, for scene understanding at a higher level of granularity like segmentation, training supervised models requires annotations at a pixel level. Such annotations require significant human effort and are often very expensive to obtain. This impedes training such supervised models on a large scale with millions of images.

One way to tackle this problem could be to train models in an unsupervised manner without requiring any segmentation annotations. The best methods \cite{cho2021picie, hamilton2022unsupervised} in this category exploit the similarity between internal representations of self-supervised image encoders \cite{caron2021emerging}. This similarity is then used to identify and cluster similar regions of the image as segmentations. These models however are significantly outperformed by their fully supervised counterparts on most segmentation benchmarks.

\begin{figure}[!t]
    \centering
    \includegraphics[width=0.9\linewidth]{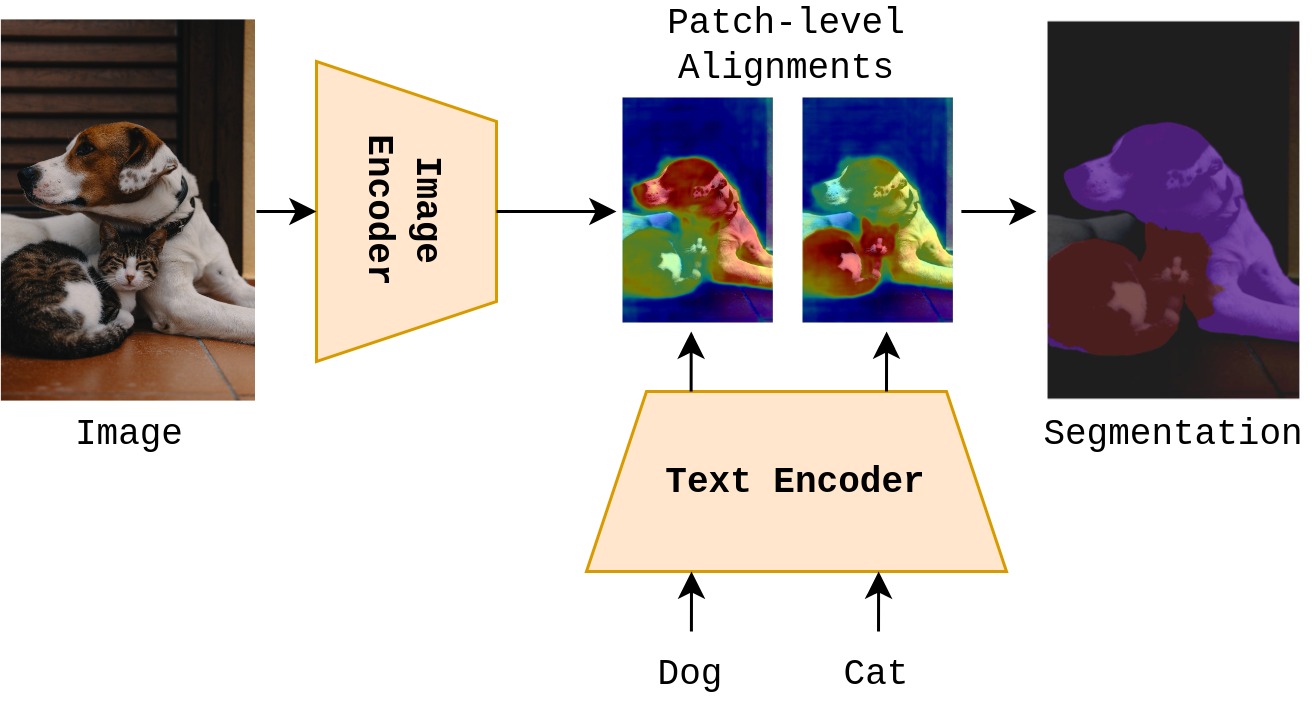}
    \caption{\textbf{High level overview of our model.} We train an alignment between the patch level embeddings from the image encoder and the CLS embedding from the text encoder. This alignment can then be used to perform open-vocabulary semantic segmentation in a zero-shot manner.}
    \vspace{-4mm}
    \label{fig:intro_fig}
\end{figure}

Recent improvements in multi-modal foundation models has led to the possibility of training on very large scale datasets scraped off the internet \cite{radford2021learning}. These datasets mostly contain pairs of images and their corresponding natural language text descriptions. Models like CLIP \cite{radford2021learning}, ALIGN \cite{jia2021scaling}, Florence \cite{yuan2021florence} and CoCa \cite{yu2022coca} trained on such large internet scale datasets have been shown to transfer very well to several downstream tasks. Furthermore, having been trained on natural language textual descriptions, these models are often expected to recognize a wide variety of real-world visual concepts which can be expressed in natural language, a setting better known as \emph{open vocabulary prediction}.

The natural question then is whether these multi-modal models can be used for pixel level predictions, i.e., semantic segmentation in the open vocabulary setting. Prior works on this topic \cite{li2022language, ghiasi2021open, xu2021simple, liu2022open, xu2022groupvit} show that this is indeed possible. However, 3 of these works use either fully supervised segmentation annotations \cite{liu2022open}, class-agnostic segmentation masks \cite{ghiasi2021open} or a region proposal model trained using segmentation annotations \cite{xu2021simple}, thereby being limited by the availability of expensive segmentation annotations/masks. To the best of our knowledge, only two models: ViL-Seg \cite{liu2022open} and GroupViT \cite{xu2022groupvit} perform the task of open-vocabulary semantic segmentation while being trained solely on image-text data. Among these two, the better performer, GroupViT, defines a modified vision transformer (ViT) \cite{dosovitskiy2020image} architecture to naturally find semantic clusters within an image. Due to a different architecture, their model has to be trained end-to-end from scratch using image-text datasets and cannot leverage pre-trained vision encoders.

In this work, we tackle the problem of open-vocabulary semantic segmentation without using any segmentation annotations or masks, with a model purely trained on image-text data. We start with the observation in \cite{hamilton2022unsupervised} that self-supervised ViT models like DINO \cite{caron2021emerging}, have similar patch representations for semantically similar regions of an image. We find this observation to be true for CLIP's ViT based vision encoders as well. However, we also find that CLIP does not exhibit a patch level alignment between its vision and text encoders, primarily owing to the fact that its contrastive loss only aligns the CLS image and text tokens.

Inspired from previous work on contrastive learning for weakly supervised phrase grounding \cite{gupta2020contrastive}, we define a new compatibility function for contrastive loss to train an alignment between the patch tokens of the vision encoder and the CLS token of the text encoder. In particular, we take the cosine similarity between the text CLS token and the vision patch tokens and use these similarities as weights to compute a weighted sum over vision tokens. The final compatibility function is then simply the cosine similarity between the weighted sum of the vision patch tokens thus obtained and the CLS text token. We find that models trained on our \emph{Patch Aligned Contrastive Learning} loss indeed exhibit the desired patch level fine-grained alignment. Thus, at inference time, the compatibility function can be used to make image level predictions and the patch level alignment can be used for zero-shot transfer to semantic segmentation. A high level overview of our model is shown in \Cref{fig:intro_fig}.

Note that unlike GroupViT, our PACL method is more flexible and general and can be used with any pre-trained ViT based encoders as well. We evaluate PACL with a pre-trained CLIP encoder on the task of zero-shot semantic segmentation using 4 different datasets: Pascal VOC \cite{Everingham10}, Pascal Context \cite{mottaghi_cvpr14}, COCO Stuff \cite{caesar2018coco} and ADE20K \cite{zhou2017scene}. On all 4 datasets, PACL consistently beats previous baselines \cite{li2022language, ghiasi2021open, liu2022open, xu2022groupvit}, even the ones which use segmentation annotations or segmentation masks for training. In addition, we find that PACL trained on top of a CLIP backbone leads to a general improvement in zero-shot classification performance across a suite of 12 different image classification datasets.

Thus, in a nutshell, our contributions are as follows. \textbf{Firstly,} we propose \emph{Patch Aligned Contrastive Learning} (PACL), a modified compatibility function for contrastive loss in order to train an alignment between the patch representations of a ViT based vision encoder and the CLS text representation of a text encoder. We show that this alignment can be used to find regions within an image corresponding to a given text input and hence, can be used for zero-shot transfer to open-vocabulary semantic segmentation. \textbf{Secondly,} we show that PACL with a pre-trained CLIP encoder obtains state-of-the-art scores on zero-shot semantic segmentation across 4 different segmentation benchmarks: Pascal VOC, Pascal Context, COCO Stuff and ADE20K. \textbf{Finally,} PACL with a CLIP backbone also shows a general improvement in performance on zero-shot classification tasks across 12 different image classification datasets.

\section{Related Work}
\label{sec:related_work}

In this section, we discuss some of the relevant works motivating our method.

\textbf{Supervised semantic segmentation:} Given an image, the task of semantic segmentation \cite{Long_2015_CVPR} involves classifying every pixel in the image to one of a fixed set of classes. Naturally, supervised datasets for semantic segmentation like Pascal VOC \cite{Everingham10}, ADE20K \cite{zhou2017scene} and Cityscapes \cite{cordts2016cityscapes} contain images with class annotations for every pixel. A significant amount of work \cite{ronneberger2015u, chen2017rethinking, wang2020deep, zhao2017pyramid} has been done to leverage these datasets and generate strong models for semantic segmentation. However, since annotating images at a pixel level is laborious and expensive, these datasets remain limited to a relatively small number of classes.

\textbf{Unsupervised semantic segmentation:} Identifying that the requirement of dense annotations is the problem, some works \cite{cho2021picie, hwang2019segsort, van2021unsupervised, hamilton2022unsupervised, melas2022deep, wang2022tokencut} have tried to leverage self-supervised learning techniques to train features which can be used for segmentation without requiring dense annotations. Notable among these works is STEGO \cite{hamilton2022unsupervised} which uses the localized feature correspondences in self-supervised models like DINO \cite{caron2021emerging} for the task of unsupervised segmentation. In our work, we study the existence of a similar feature correspondence in vision encoders of multi-modal models like CLIP \cite{radford2021learning} and use it to train a patch level alignment between image and text modalities. Note however, that it is still difficult for such unsupervised segmentation approaches to scale up to a large number of visual concepts.

\textbf{Natural language supervision:} Recently, the availability of datasets with millions of image-text pairs scraped from the internet has made it possible to train large-scale multi-modal fusion models on such datasets. Such models \cite{radford2021learning, jia2021scaling, kim2021vilt, singh2022flava, yu2022coca, yuan2021florence} are able to transfer well to several downstream tasks including vision-language pre-training (VLP) \cite{chen2022vlp} tasks like image-text retrieval \cite{wang2016comprehensive} and visual question answering \cite{antol2015vqa}, as well as vision specific tasks like zero-shot image classification \cite{radford2021learning, jia2021scaling, yu2022coca} and object detection \cite{zhang2022glipv2, kamath2021mdetr}. Given the large-scale training of such multi-modal fusion models, it is then natural to ask if these models can be leveraged to scale up the task of semantic segmentation and recognise a large number of visual concepts at a fine-grained level.

\textbf{Natural language supervision for zero-shot segmentation:} Some work has been done in this direction of using large-scale multi-modal models, like CLIP \cite{radford2021learning}, for the task of semantic segmentation. For instance, LSeg \cite{li2022language} trains a segmentation model as its vision encoder and uses the frozen text encoder from CLIP to align pixel level embeddings with text. The resulting model is able to recognise conceptually similar labels which are not present within the training set. However, it trains the vision encoder in a fully supervised manner using segmentation annotations. OpenSeg \cite{ghiasi2021open} on the other hand is based on the ALIGN \cite{jia2021scaling} model and trains using image-text data and class-agnostic segmentation annotations. ViL-Seg \cite{liu2022open} trains using only image-text data with a vision based contrasting and a cross-modal contrasting objective along with an online clustering head to segment visual embeddings. Finally, GroupViT \cite{xu2022groupvit} proposes a modified ViT architecture which allows grouping semantically similar tokens into clusters useful for open vocabulary segmentation. To the best of our knowledge, ViL-Seg and Group-ViT are the only existing methods which solely use image-text data for training an open vocabulary semantic segmentation model. In our work, we propose a simple modification to the CLIP compatibility function for contrastive loss, which enables training an alignment between the patch tokens of a ViT based vision encoder and the CLS token of a text encoder. This alignment can then be seamlessly utilized for the task of semantic segmentation without using any segmentation annotations or class-agnostic segmentation masks during training.

\section {Patch Level Alignment in CLIP}
\label{sec:clip_lacks_alignment}

The contrastive training of CLIP ensures that the CLS tokens obtained from CLIP's transformer based vision and text encoders are aligned for similar image-text pairs. However, such an alignment between image and text at a patch level does not necessarily exist. %
To empirically study this, we use a semantic segmentation dataset, Pascal VOC \cite{Everingham10}, and classify each patch in the dataset to one of a fixed set of classes. The patch level vision tokens are classified using the same zero-shot classification \cite{radford2021learning} method normally used on the CLS vision token. The classification accuracy, thus obtained, provides a measure of patch level alignment between the vision and text representations in the model, where a high classification accuracy indicates a high alignment and vice-versa.

More formally, let $\mathcal{D}_{\mathrm{seg}} = (\mathbf{x}, \mathbf{y})_{i=1}^N$ be the semantic segmentation dataset where $\mathbf{x} \in \mathbb{R}^{C, H, W}$ and $\mathbf{y} \in \mathbb{R}^{H, W}$. We represent CLIP's vision and text encoders as $f_v: \mathbb{R}^{C, H, W} \rightarrow \mathbb{R}^{D_v}$ and $f_t: \mathbb{R}^{l} \rightarrow \mathbb{R}^{D_t}$ respectively. Similarly, let $e_v: \mathbb{R}^{D_v} \rightarrow \mathbb{R}^{D}$ and $e_t: \mathbb{R}^{D_t} \rightarrow \mathbb{R}^{D}$ be the linear embedders to project the vision and text encodings to the joint $D$ dimensional space. Normally, for zero-shot classification, we measure the cosine similarity between the vision and text embeddings: $s(\mathbf{x}, c) = \frac{e_v(f_v(\mathbf{x}))}{|e_v(f_v(\mathbf{x}))|} \cdot \frac{e_t(f_t(c))}{|e_t(f_t(c))|}$ for each class name $c$ and compute the predictive probability as: $p(c|\mathbf{x}) = \frac{e^{s(\mathbf{x}, c)}}{\sum_{c'}e^{s(\mathbf{x}, c')}}$. A simple modification to the vision encoder: $\hat{f}_v: \mathbb{R}^{C, H, W} \rightarrow \mathbb{R}^{T, D_v}$, where $T$ is the number of tokens or patches, allows us to perform the same classification method on every patch.

\begin{figure}[!t]
    \centering
    \begin{subfigure}{0.25\linewidth}
        \centering
        \includegraphics[width=\linewidth]{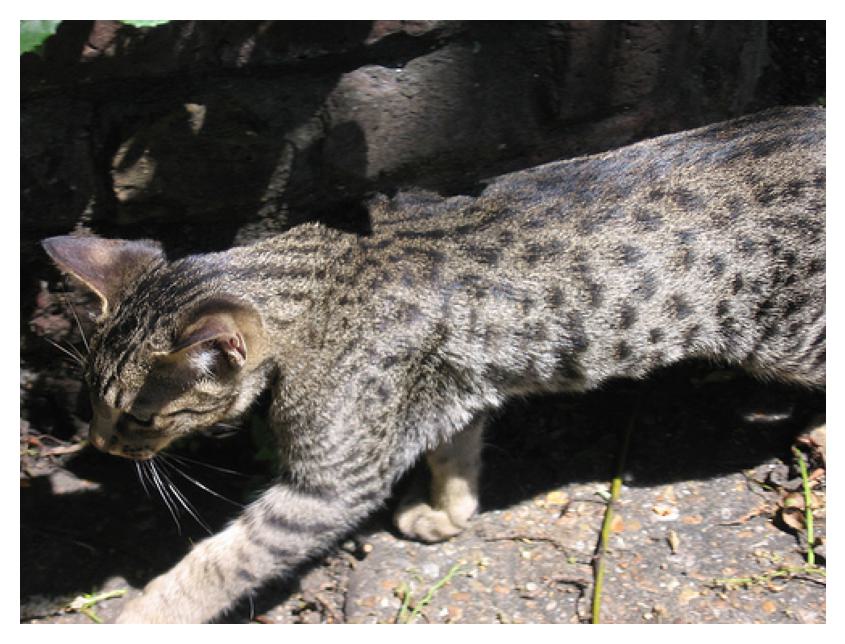}
        \vspace{-4mm}
    \end{subfigure}
    \begin{subfigure}{0.25\linewidth}
        \centering
        \includegraphics[width=\linewidth]{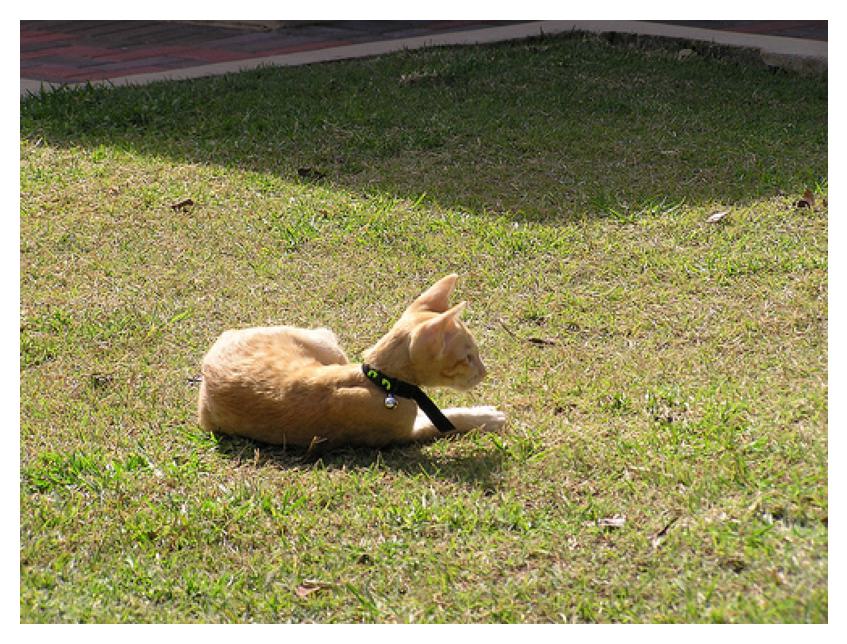}
        \vspace{-4mm}
    \end{subfigure}
    \begin{subfigure}{0.28\linewidth}
        \centering
        \includegraphics[width=\linewidth]{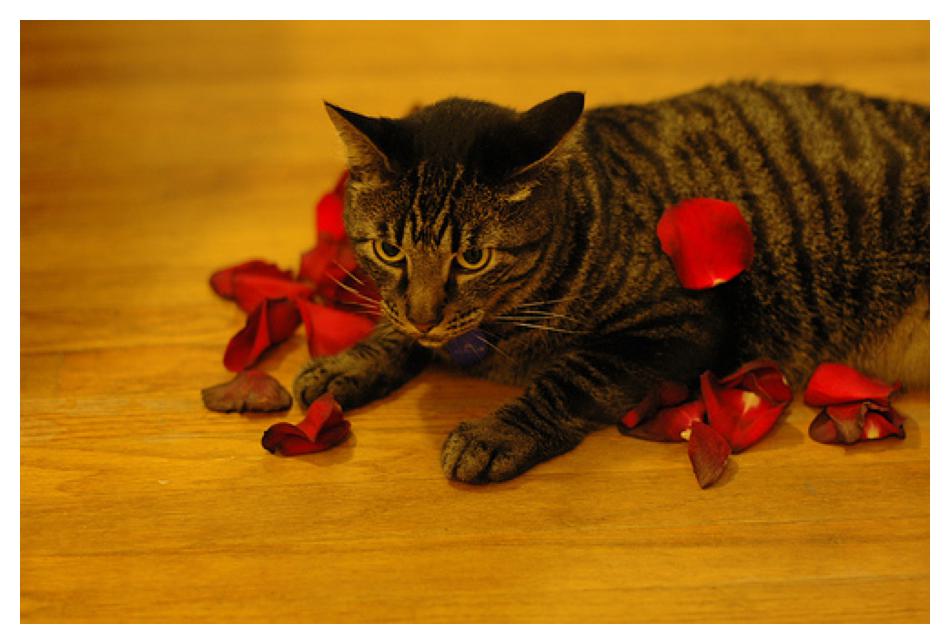}
        \vspace{-4mm}
    \end{subfigure}
    \begin{subfigure}{0.14\linewidth}
        \centering
        \includegraphics[width=\linewidth]{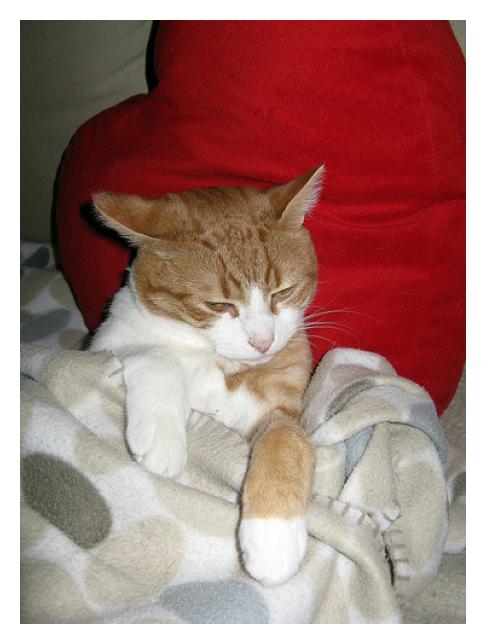}
        \vspace{-4mm}
    \end{subfigure}

    \begin{subfigure}{0.25\linewidth}
        \centering
        \includegraphics[width=\linewidth]{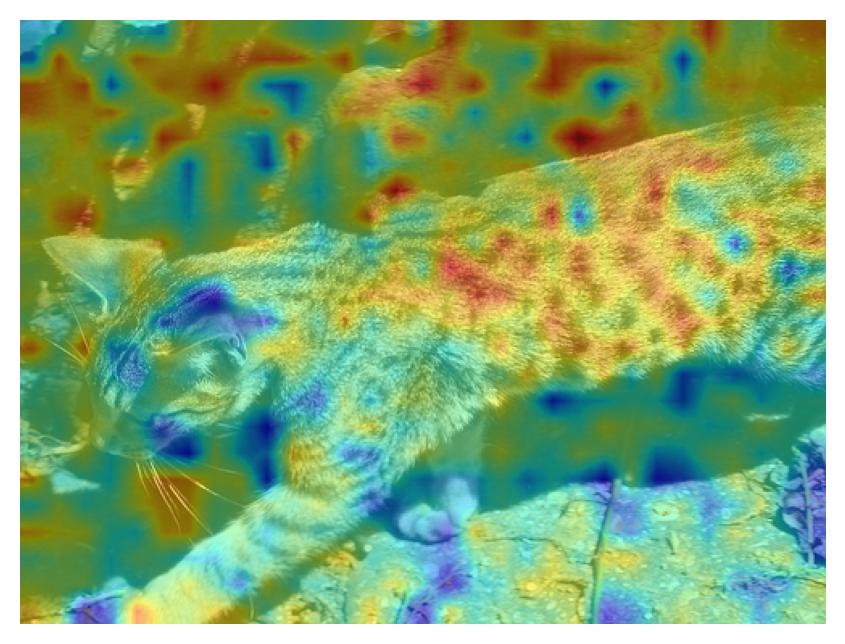}
        \vspace{-4mm}
    \end{subfigure}
    \begin{subfigure}{0.25\linewidth}
        \centering
        \includegraphics[width=\linewidth]{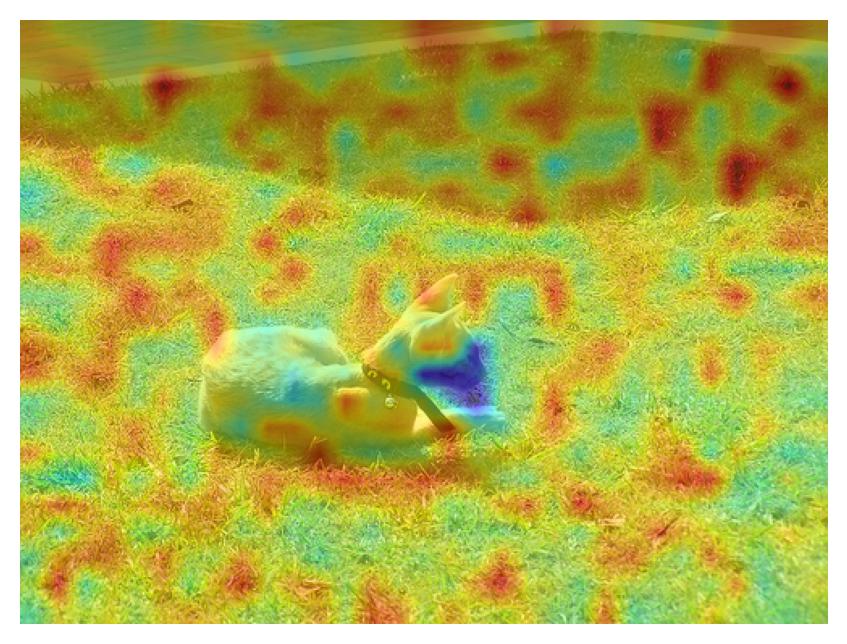}
        \vspace{-4mm}
    \end{subfigure}
    \begin{subfigure}{0.28\linewidth}
        \centering
        \includegraphics[width=\linewidth]{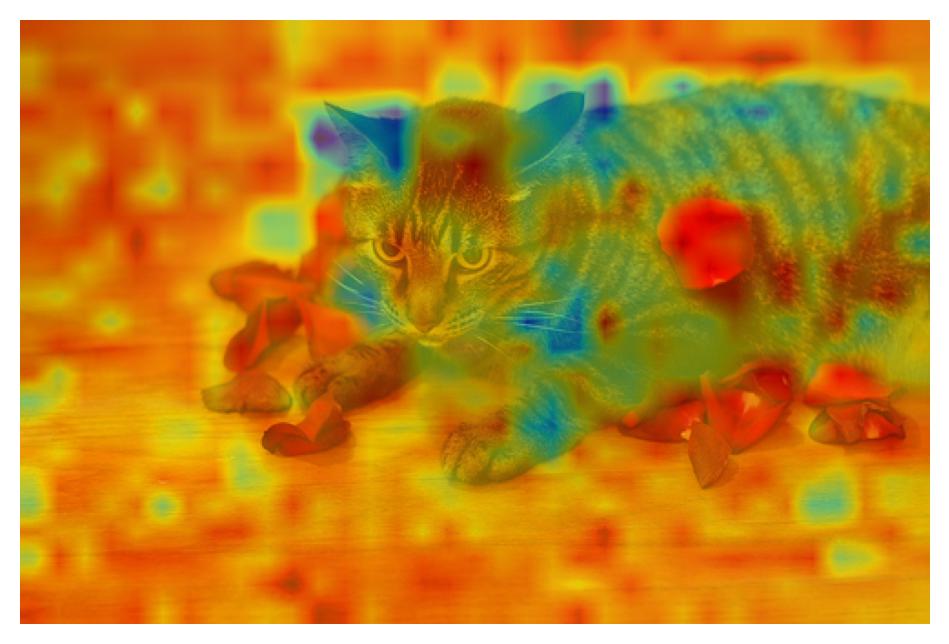}
        \vspace{-4mm}
    \end{subfigure}
    \begin{subfigure}{0.14\linewidth}
        \centering
        \includegraphics[width=\linewidth]{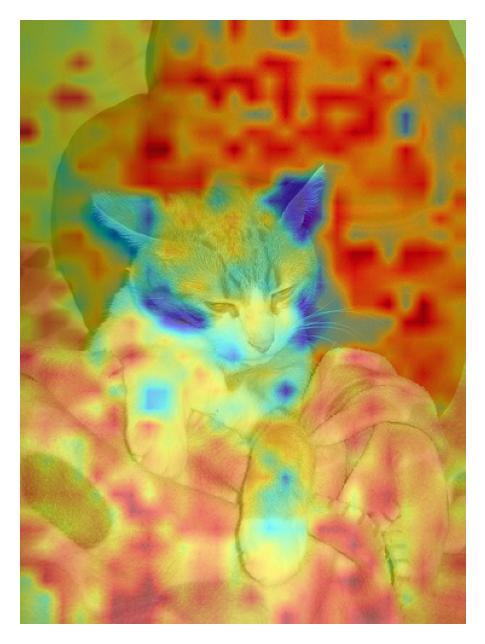}
        \vspace{-4mm}
    \end{subfigure}

    \begin{subfigure}{0.25\linewidth}
        \centering
        \includegraphics[width=\linewidth]{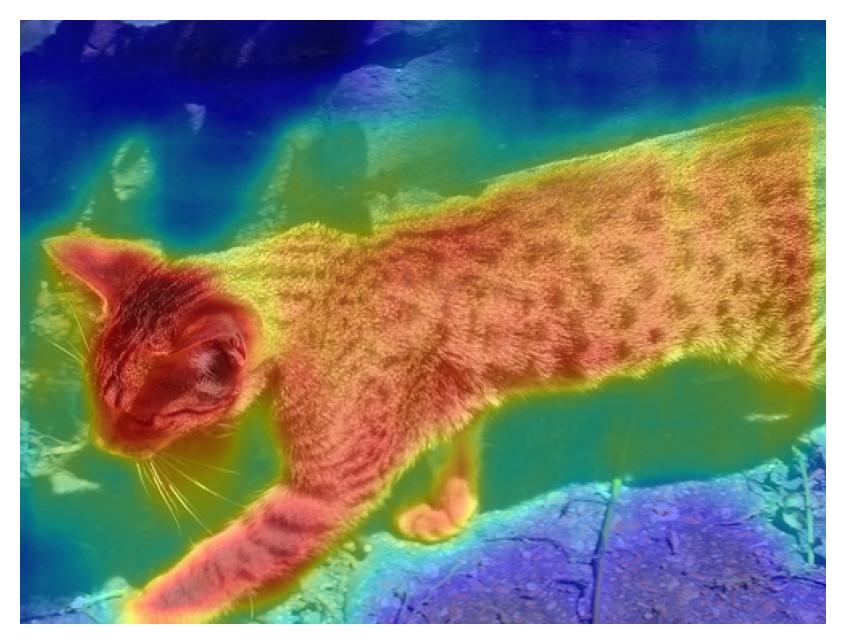}
    \end{subfigure}
    \begin{subfigure}{0.25\linewidth}
        \centering
        \includegraphics[width=\linewidth]{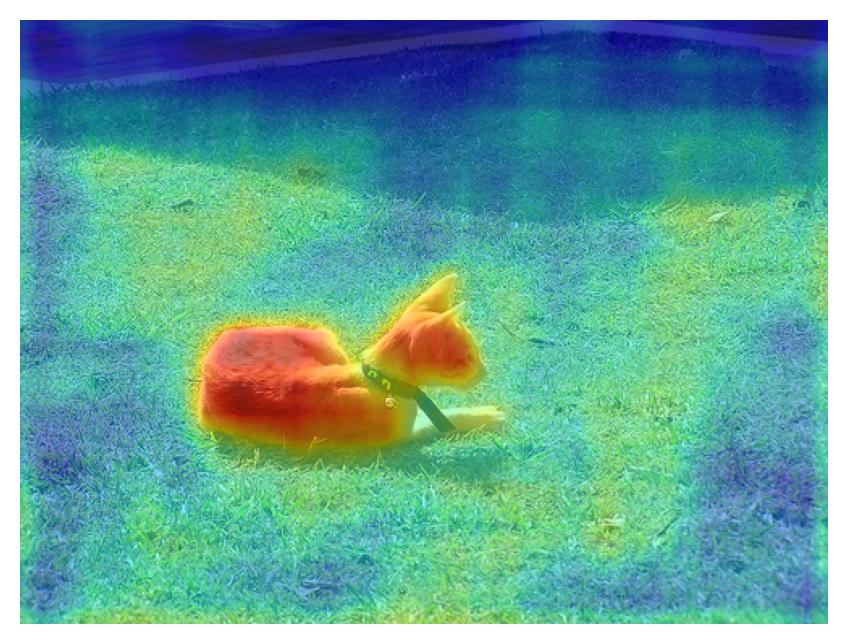}
    \end{subfigure}
    \begin{subfigure}{0.28\linewidth}
        \centering
        \includegraphics[width=\linewidth]{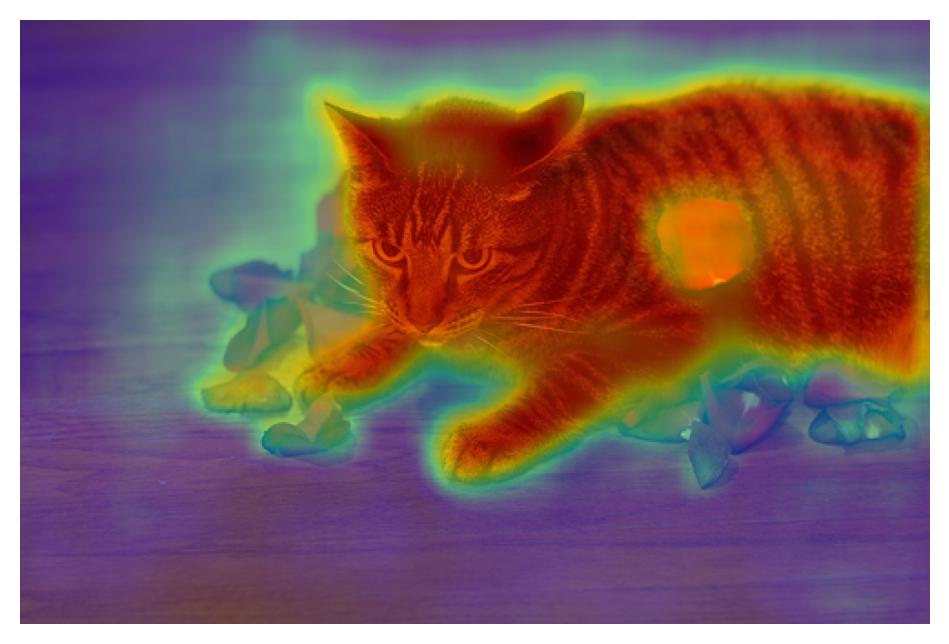}
    \end{subfigure}
    \begin{subfigure}{0.14\linewidth}
        \centering
        \includegraphics[width=\linewidth]{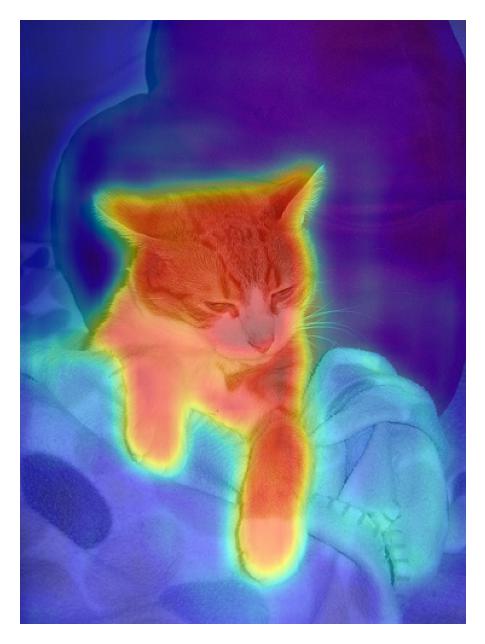}
    \end{subfigure}
    \vspace{-2mm}
    \caption{\textbf{Patch level alignment between the word ``cat" and images of cats.} In the first row, we show the original images, in the second row, we show the patch level alignment in CLIP ViT-B/16 and in the third row, we show the alignment for our method.}
    \vspace{-4mm}
    \label{fig:alignment_qualitative}
\end{figure}

\begin{table}
    \centering
    \begin{minipage}{0.6\linewidth}
    \centering
    \scriptsize
    \resizebox{\linewidth}{!}
    {
    \begin{tabular}{ccc}
    \toprule
    & \multicolumn{2}{c}{\textbf{Patch Classification Accuracy}} \\
    \textbf{CLIP Vision Encoder} & \textbf{\textit{Pre-Alignment}} & \textbf{\textit{Post-Alignment}} \\
    \midrule
    ViT-B-16 & $52.49$ & $96.51$ \\
    ViT-L/14 & $27.91$ & $95.33$ \\
    \bottomrule
    \end{tabular}
    }
    \caption{\textbf{Accuracy for patch level classification on Pascal VOC.} For a pre-trained CLIP model, the accuracy is low indicating low patch level alignment between image and text. On applying our PACL alignment method, the accuracy significantly increases for both CLIP encoders indicating higher image text patch level alignment.}
    \label{table:alignment_quantitative}
    \end{minipage} \hfill
    \begin{minipage}{0.37\linewidth}
        \centering
        \includegraphics[width=\linewidth]{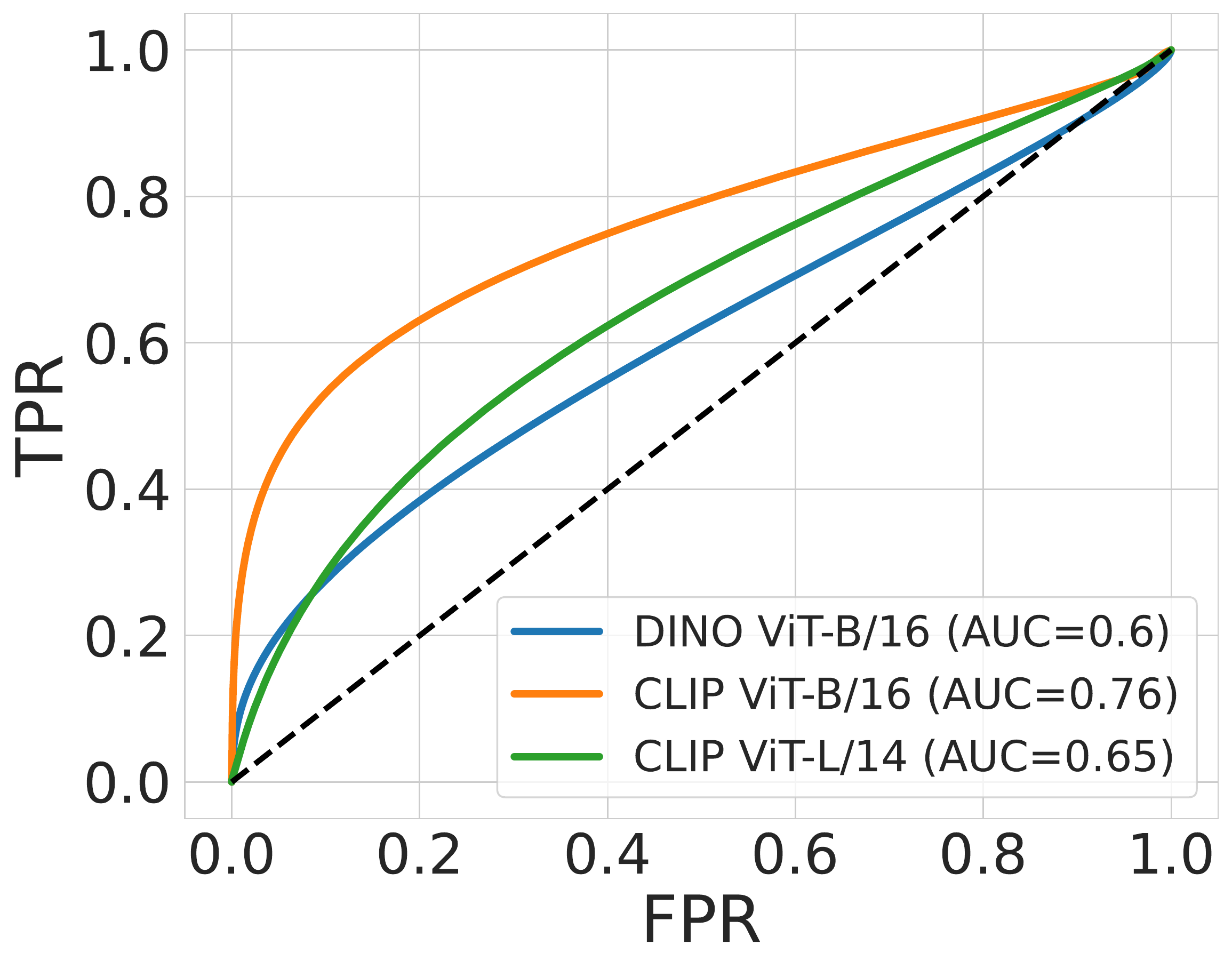}
        \captionof{figure}{\textbf{ROC curve indicating semantic coherence of CLIP and DINO vision encoders.} CLIP encoders outperform DINO.}
        \label{fig:roc_coherence}
    \end{minipage}
\end{table}

In \Cref{table:alignment_quantitative} second column (Pre-Alignment), we show the patch classification accuracy thus obtained for two CLIP models: ViT-B/16 and ViT-L/14. In \Cref{fig:alignment_qualitative}, first and second rows, we show qualitative samples of alignment on CLIP ViT-B/16, for 4 images of cats from Pascal VOC. With a patch classification accuracy of $52.49\%$ for ViT-B/16 and $27.91\%$ for ViT-L/14, it is clear that the alignment we seek is very poor at the patch level. Surprisingly, note that for ViT-L/14, a model known to provide better image level prediction performance than ViT-B/16, the patch level alignment is significantly worse. Hence, \emph{pre-trained CLIP models cannot be used for open vocabulary segmentation as the CLIP contrastive learning objective does not ensure patch level alignment between image and text modalities}.

\section{Semantic Coherence in Vision Encoders}
\label{sec:semantic_coherence}

Due to the poor patch level alignment between pre-trained CLIP image and text encoders, our next question is whether we can train such an alignment in CLIP. This would however require the pre-trained vision encoder to be \emph{sematically coherent}. In other words, semantically similar regions in an image should produce similar patch representations in the vision encoder. This property has been studied before in image self-supervised models like DINO \cite{hamilton2022unsupervised}. %
We use a similar test to quantify semantic coherence of CLIP's vision encoders.

In particular, we collect all patch representations from the vision encoder for each image in Pascal VOC and store the corresponding target classes using the segmentation labels. Let $\hat{f}_v(\mathbf{x_1})_{i, j} \in \mathbb{R}^{D_v}$ and $\hat{f}_v(\mathbf{x_2})_{p, q} \in \mathbb{R}^{D_v}$ be the patch representations obtained at index $(i, j)$ of image $\mathbf{x_1}$ and index $(p, q)$ of image $\mathbf{x_2}$ respectively. We compute the cosine similarity $\left(\frac{\hat{f}_v(\mathbf{x_1})_{i, j}}{|\hat{f}_v(\mathbf{x_1})_{i, j}|} \cdot \frac{\hat{f}_v(\mathbf{x_2})_{p, q}}{|\hat{f}_v(\mathbf{x_2})_{p, q}|}\right)$ between the patch representations and use this as a binary classifier to predict if the two patches have the same target label. Let the segmentation labels for the two patches be $l(\mathbf{x_1})_{i, j}$ and $l(\mathbf{x_2})_{p, q}$ respectively. Since we have labels for each pixel, we decide the label for each patch by majority-voting. The target value for binary classification is $1$ if $l(\mathbf{x_1})_{i, j} = l(\mathbf{x_2})_{p, q}$, else $0$. Note that performance on this binary classification task is indicative of semantic coherence, as a good classifier would require patch representations corresponding to same labels to have high cosine similarity and vice-versa.

\begin{figure}[!t]
    \centering
    \begin{subfigure}{0.18\linewidth}
        \centering
        \includegraphics[width=\linewidth]{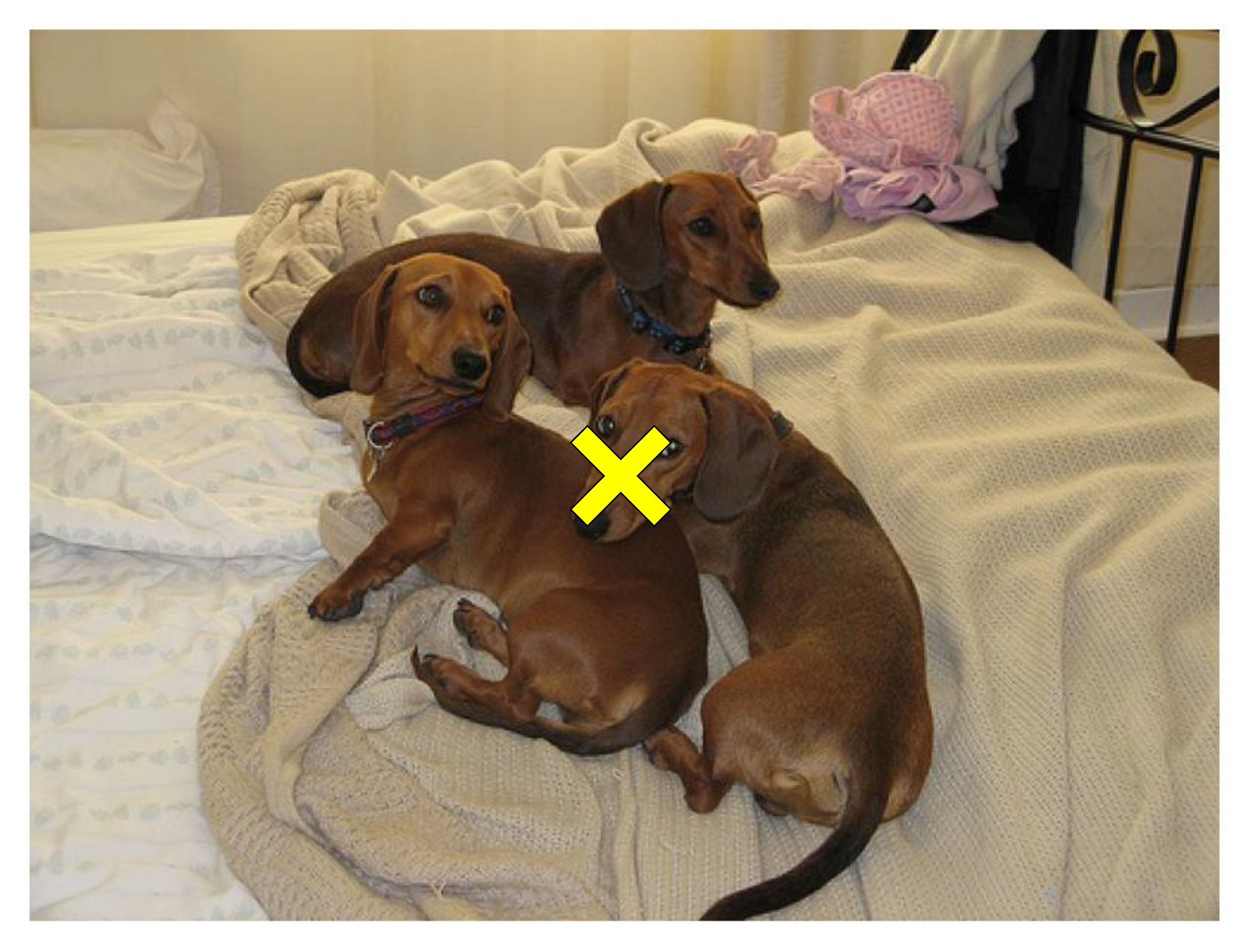}
        \caption{}
        \label{subfig:original_qualitative}
    \end{subfigure}
    \begin{subfigure}{0.18\linewidth}
        \centering
        \includegraphics[width=\linewidth]{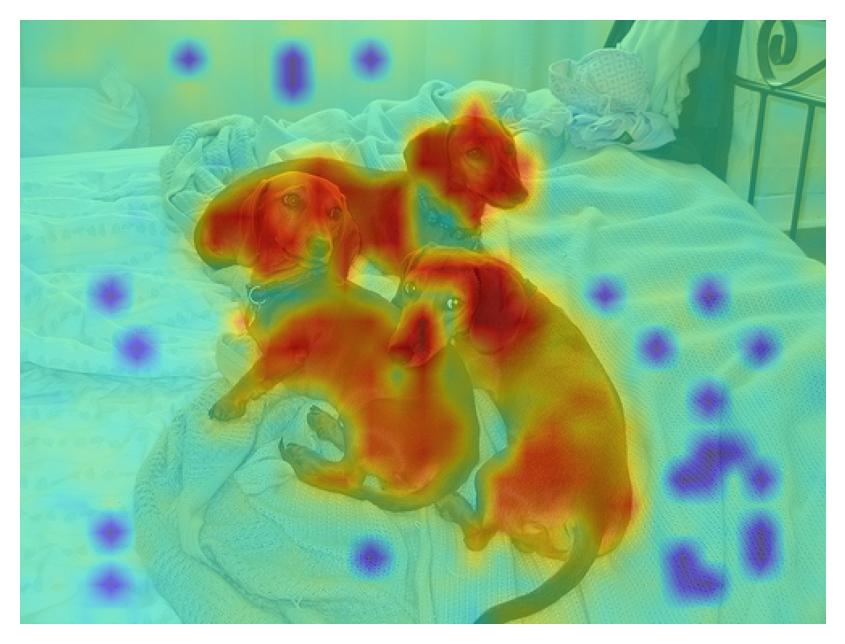}
        \caption{}
        \label{subfig:clip_coherence_same_image}
    \end{subfigure}
    \begin{subfigure}{0.20\linewidth}
        \centering
        \includegraphics[width=\linewidth]{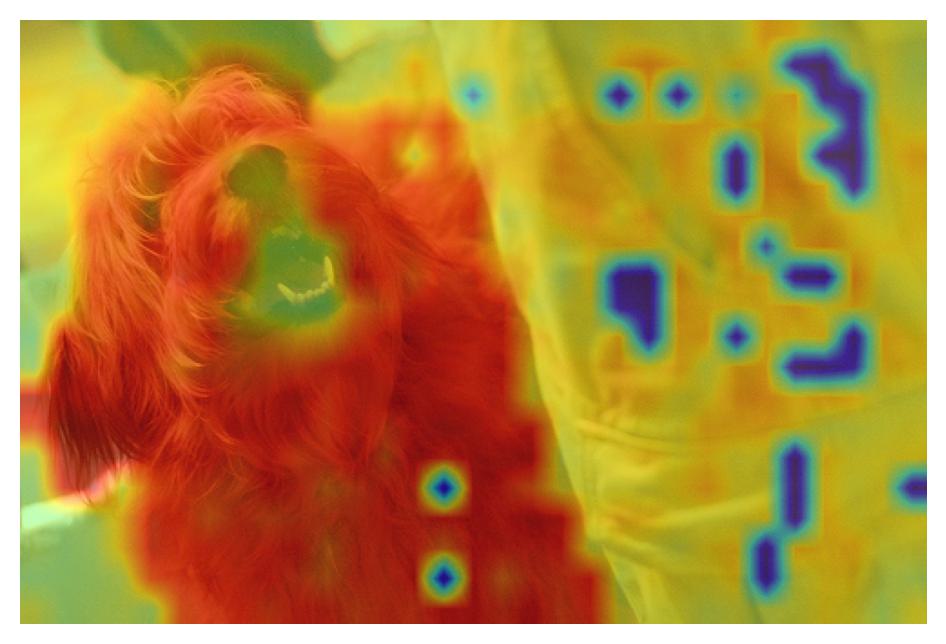}
        \caption{}
        \label{subfig:clip_coherence_diff_image}
    \end{subfigure}
    \begin{subfigure}{0.18\linewidth}
        \centering
        \includegraphics[width=\linewidth]{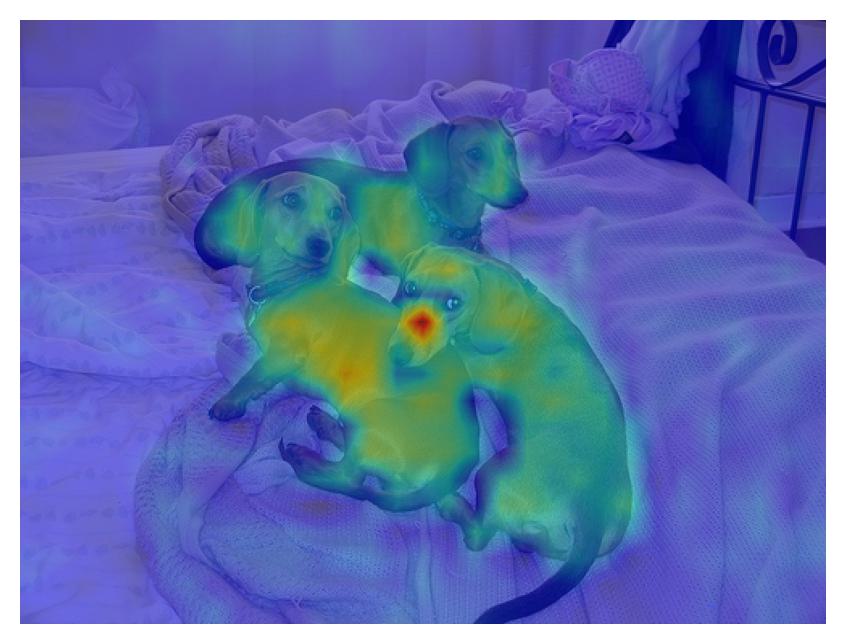}
        \caption{}
        \label{subfig:dino_coherence_same_image}
    \end{subfigure}
    \begin{subfigure}{0.20\linewidth}
        \centering
        \includegraphics[width=\linewidth]{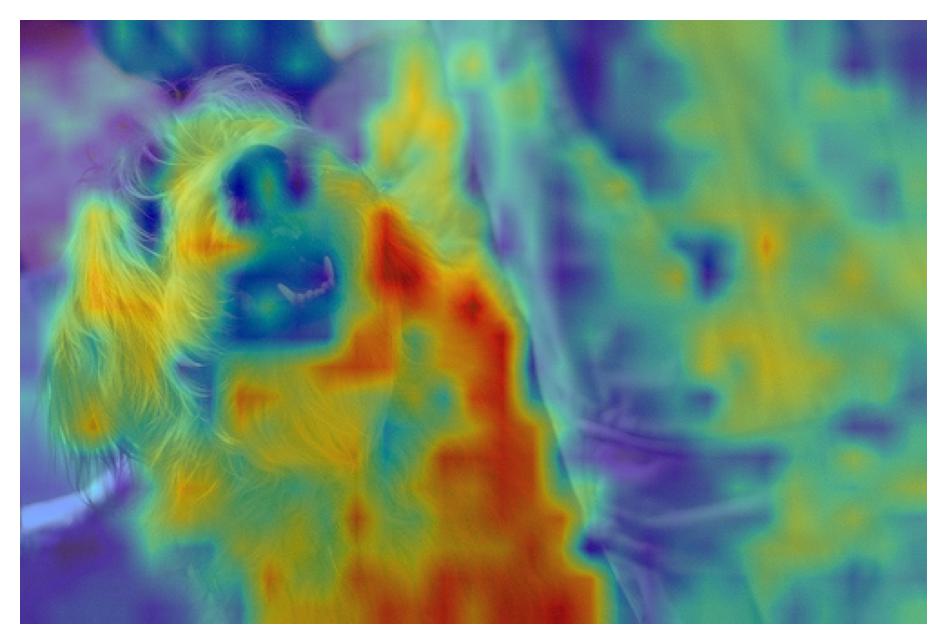}
        \caption{}
        \label{subfig:dino_coherence_diff_image}
    \end{subfigure}
    \vspace{-2mm}
    \caption{\textbf{Qualitative results on semantic coherence between CLIP and DINO ViT-B/16.} \textbf{a)}: we show the original image of a dog class with the patch marker (yellow X near the centre). \textbf{b, c)}: we show CLIP vision encoder cosine similarity across all patches for the same and a different image of a dog. \textbf{d, e)}: we show the same for DINO. See more examples in \Cref{app:semantic_coherence}.}
    \vspace{-4mm}
    \label{fig:coherence_qualitative}
\end{figure}

We present the ROC curve and the AUROC scores for CLIP and DINO in \Cref{fig:roc_coherence}. Surprisingly, we find that \emph{CLIP's vision encoders outperform DINO on semantic coherence}\footnote{CLIP outperforming DINO on semantic coherence indicates that CLIP's vision encoders are good candidates for unsupervised segmentation approaches like STEGO \cite{hamilton2022unsupervised}, but further study of this feature is beyond the scope of this work.}. This is encouraging as it indicates that we can indeed train a mapping between similar vision tokens and their corresponding text representations. We also present qualitative results in \Cref{fig:coherence_qualitative} where we plot the patch level cosine similarity between a chosen patch (marked in yellow X in \Cref{subfig:original_qualitative}) and the remaining patches in the same image as well as a different image having the same class (dog). We do this for CLIP ViT-B/16 in \Cref{subfig:clip_coherence_same_image} and \Cref{subfig:clip_coherence_diff_image} and for DINO ViT-B/16 in \Cref{subfig:dino_coherence_same_image} and \Cref{subfig:dino_coherence_diff_image}. In both cases, CLIP's encoder seems to perform at par or better than DINO. Motivated by these observations, in the next section, we discuss a method to train a patch level alignment between the vision tokens and the CLS text token in CLIP using purely image-text data.

\vspace{-2mm}
\section{Patch Aligned Contrastive Learning (PACL)}
\label{sec:pacl}

In the previous section, we showed that although CLIP lacks a patch level alignment between image and text representations, such an alignment can indeed be trained. However, note that this is a difficult problem as there is no ground-truth text data annotating each patch in an image-text dataset. Hence, training such an alignment can only be done in a weakly supervised fashion. Inspired from previous work on weakly supervised phrase grounding \cite{gupta2020contrastive}, in this section, we propose a modification on CLIP's contrastive loss, to learn an alignment between the vision patch tokens and the CLS text token.

\textbf{A modified compatibility function for contrastive loss:} Our method is simple in the sense that the only change we make to CLIP's training is in the compatibility function of its contrastive loss. Normally, for an image-text pair $(\mathbf{x}, \mathbf{y})$, CLIP computes the CLS vision and text embeddings as $e_v(f_v(\mathbf{x}))$ and $e_t(f_t(\mathbf{y}))$ respectively, where $f_v: \mathbb{R}^{C,H,W} \rightarrow \mathbb{R}^{D_v}$, $f_t: \mathbb{R}^{L} \rightarrow \mathbb{R}^{D_t}$ are the vision and text encoders and $e_v: \mathbb{R}^{D_v} \rightarrow \mathbb{R}^{D}$, $e_t: \mathbb{R}^{D_t} \rightarrow \mathbb{R}^{D}$ are the vision and text embedders to project the representations into the same dimensional space. The compatibility function $\phi(\mathbf{x}, \mathbf{y})$ is the cosine similarity between the vision and text CLS embeddings: $\phi(\mathbf{x}, \mathbf{y}) = \left(\frac{e_v(f_v(\mathbf{x}))}{|e_v(f_v(\mathbf{x}))|} \cdot \frac{e_t(f_t(\mathbf{y}))}{|e_t(f_t(\mathbf{y}))|}\right)$. Given this compatibility function, CLIP uses the InfoNCE \cite{oord2018representation} contrastive loss to learn vision and text representations which are aligned for similar image-text pairs:
\begin{equation}
\begin{split}
    & \mathcal{L}_{\mathbf{x}} = \frac{1}{k} \sum_{i=1}^{k} \left(\frac{e^{\phi(\mathbf{x_i}, \mathbf{y_i})}}{\sum_{j=1}^k e^{\phi(\mathbf{x_i},\mathbf{y_j})}}\right) \\
    & \mathcal{L}_{\mathbf{y}} = \frac{1}{k} \sum_{i=1}^{k} \left(\frac{e^{\phi(\mathbf{x_i}, \mathbf{y_i})}}{\sum_{j=1}^k e^{\phi(\mathbf{x_j},\mathbf{y_i})}}\right)
\end{split}
\end{equation}
with the contrastive loss being $\mathcal{L}_\mathrm{InfoNCE} = 1/2 (\mathcal{L}_{\mathbf{x}} + \mathcal{L}_{\mathbf{y}})$.

Note that the above loss function produces an alignment between the CLS image and text tokens but as we observed in \Cref{sec:clip_lacks_alignment}, it does not produce the desired alignment at patch level between vision and text encoders. In order to then train this alignment, we make the following changes to CLIP's loss. First, we use the patch tokens instead of the CLS token from the vision encoder, $\hat{f}_v: \mathbb{R}^{C, H, W} \rightarrow \mathbb{R}^{T, D_v}$, where $T$ is the number of tokens or patches. Next, we use a modified vision embedder $\hat{e}_v: \mathbb{R}^{T, D_v} \rightarrow \mathbb{R}^{T, D}$ to generate embeddings in the shared D-dimensional space for all patch tokens. We compute the patch level similarity
\begin{equation}
\label{eq:patch_similarity}
    s(\mathbf{x}, \mathbf{y}) = \hat{e}_v(\hat{f}_v(\mathbf{x}))e_t(f_t(\mathbf{y}))
\end{equation}
between all vision patch embeddings and the CLS text embedding, where $s(\mathbf{x}, \mathbf{y}) \in \mathbb{R}^T$. We normalize the patch level similarity to the range $[0, 1]$ by applying a softmax function across tokens, $a(\mathbf{x}, \mathbf{y}) = \mathrm{softmax}(s(\mathbf{x}, \mathbf{y}))$. Finally, we take a weighted sum across all vision patch embeddings where the weights of the tokens are obtained from the patch level similarities $a(\mathbf{x}, \mathbf{y})$ as:
\begin{equation}
    \hat{v} = \hat{e}_v(\hat{f}_v(\mathbf{x}))^\intercal a(\mathbf{x}, \mathbf{y})
\end{equation}
where $\hat{v} \in \mathbb{R}^D$. The updated compatibility function $\hat{\phi}(\mathbf{x}, \mathbf{y})$ is then computed as the following dot product:
\begin{equation}
    \hat{\phi}(\mathbf{x}, \mathbf{y}) = \left(\frac{\hat{v}}{|\hat{v}|} \cdot \frac{e_t(f_t(\mathbf{y}))}{|e_t(f_t(\mathbf{y}))|}\right). 
\end{equation}

\begin{figure*}[!t]
    \centering
    \includegraphics[width=0.6\linewidth]{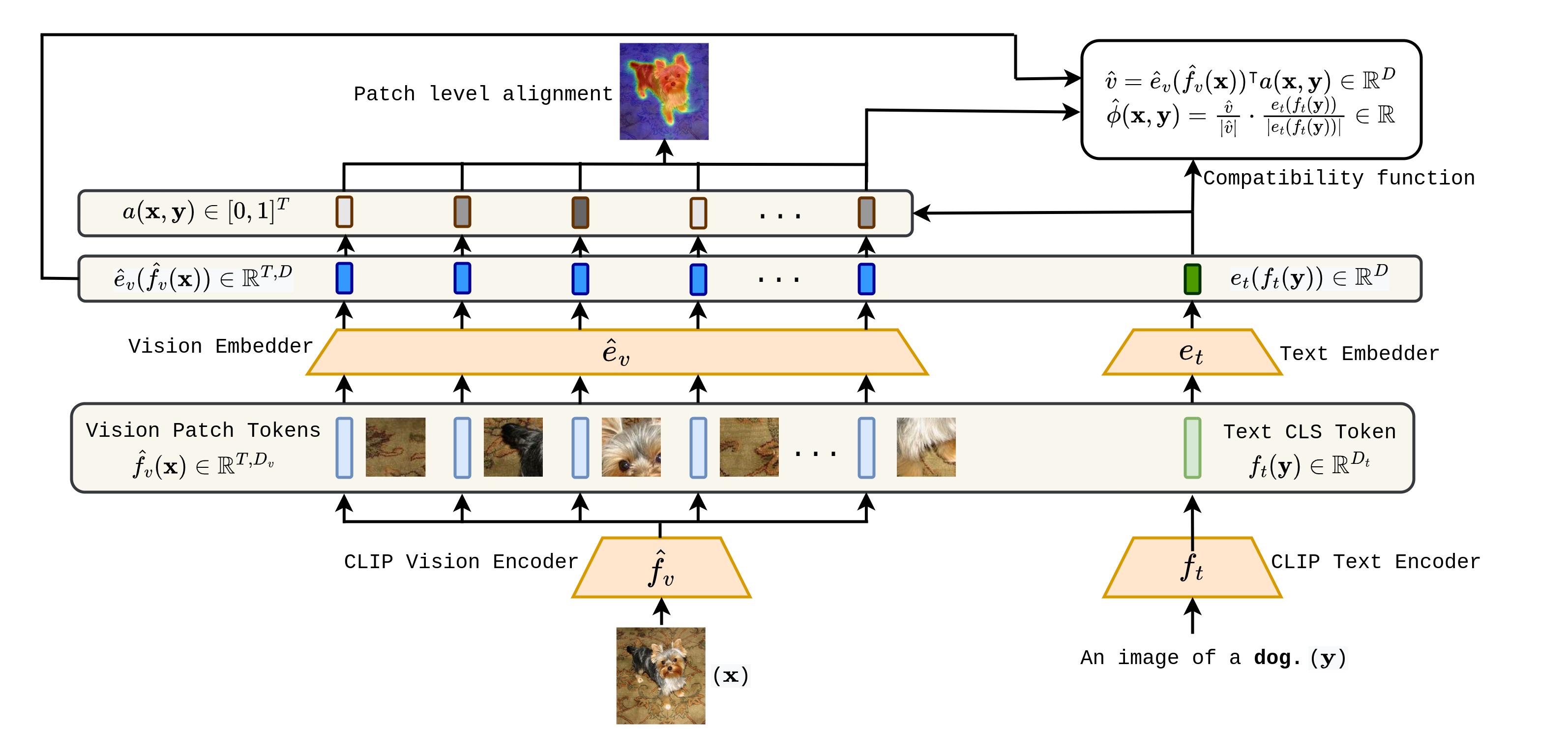}
    \caption{\textbf{Compatibility function $\phi(\mathbf{x}, \mathbf{y})$ for Patch Aligned Contrastive Learning (PACL).} The image encoder $\hat{f}_v$ and embedder $\hat{e}_v$ produce patch level representations for each image whereas the text encoder $f_t$ and embedder $e_t$ produce the CLS representation for a given text. We compute the cosine similarity between the CLS text embedding and the vision patch embeddings and use them as weights to take a weighted sum over vision patch tokens. We use the cosine similarity between the weighted sum and the CLS text token as our compatibility $\hat{\phi}(\mathbf{x}, \mathbf{y})$.}
    \vspace{-4mm}
    \label{fig:pacl_compatibility_function}
\end{figure*}

We use this modified compatibility function with InfoNCE contrastive loss for training and we call this method \emph{Patch Aligned Contrastive Learning}. \Cref{fig:pacl_compatibility_function}, shows a diagrammatic representation of the steps involved in computing the compatibility function for an image-text pair.

\textbf{Grounded in Mutual Information: } To understand how our compatibility function $\hat{\phi}(\mathbf{x}, \mathbf{y})$ works, we go back to the relation of the InfoNCE \cite{oord2018representation} loss with mutual information (MI). Let $x \in \mathcal{X}$ and $y \in \mathcal{Y}$ be two multivariate random variables with a joint probability function $p(x, y)$. MI between $x$ and $y$, computed as $\mathbb{I}[x, y] = \mathbb{E}_{(x, y) \sim p(x, y)}\log \left[\frac{p(x, y)}{p(x)p(y)}\right]$, captures the amount of information shared between $x$ and $y$. However, MI is computationally intractable and hence requires approximations in order to be estimated. The InfoNCE loss $\mathcal{L}_{\mathrm{InfoNCE}}(\theta)$ defined using a compatibility function $\phi_{\theta}(x, y)$ with model parameters $\theta$ provides such an estimate and is a lower bound on MI as: $\mathbb{I}[x, y] \geq \log(k) - \mathcal{L}_{\mathrm{InfoNCE}}(\theta)$, where $k$ is the batch size in InfoNCE loss with one positive sample and $k-1$ negative samples per batch. Hence, minimizing $\mathcal{L}_{\mathrm{InfoNCE}}$ maximises the lower bound estimate of MI.

In vanilla CLIP training, the random variables $x$ and $y$ are images $\mathbf{x}$ and texts $\mathbf{y}$ respectively, the compatibility function $\phi_{\theta}(x, y)$ is $ \phi(\mathbf{x}, \mathbf{y})$, i.e., the cosine similarity between CLS vision and text token embeddings, and the model parameters are $\theta = \{f_v, e_v, f_t, e_t\}$.
Since, we modify the compatibility function $\hat{\phi}(\mathbf{x}, \mathbf{y})$ using a weighted sum over vision tokens, to maximise MI $\mathbb{I}[x, y]$ between image and text, $\mathcal{L}_{\mathrm{InfoNCE}}$ will have to attend to regions of the image which correspond to the text and assign such regions a higher value in $s(\mathbf{x}, \mathbf{y})$. This indicates that $s(\mathbf{x}, \mathbf{y})$ intuitively captures patch level alignment between image and text modalities. To empirically verify this, we conduct the same patch level classification task described in \Cref{sec:clip_lacks_alignment} where for each patch, we compute the similarity $s(\mathbf{x}, \mathbf{y})$ for all classes and predict the class with the highest similarity. Results are in \Cref{table:alignment_quantitative}, third column (Post-Alignment) with qualitative results in \Cref{fig:alignment_qualitative}, third row. In both cases, we indeed observe a stark improvement in patch level alignment compared to vanilla CLIP using our modified compatibility function.

It is worth noting here that a similar contrastive learning approach has been used for the problem of weakly supervised phrase grounding in \cite{gupta2020contrastive}. Their approach learns a mapping between ROI features from an object detector and word representations from a language model using an attention based weakly supervised contrastive learning. Although similar to our approach, they require the use of an object detector to provide ROI features, whereas we use CLIP's vision encoder patch tokens as region features, having shown (see \Cref{sec:semantic_coherence}) that such features indeed are semantically coherent. Furthermore, they also use a contextualised language model to generate negative samples for contrastive loss, whereas our method fits in seamlessly with the contrastive setting in CLIP. Finally, whereas they target weakly supervised phrase grounding, we aim to learn a multi-modal model which is zero-shot transferable to the task of open vocabulary semantic segmentation.

\textbf{Inference:} At inference time, we can compute both image level as well as dense predictions. For image level predictions, similar to CLIP, we simply use our compatibility function $\hat{\phi}(\mathbf{x}, \mathbf{y})$ to compute similarity between an image and text. For semantic segmentation, given an image $\mathbf{x}$ and a set of classnames $Y = \{\mathbf{y}_1, ..., \mathbf{y}_C\}$, we compute $s(\mathbf{x}, \mathbf{y_c}) {\ } \forall c \in \{1, ..., C\}$ as a mask for each class and then use a softmax across classes. In the next section, we provide a detailed set of experiments to show the performance of our approach at both zero-shot semantic segmentation as well as image classification tasks.

\section{Experiments \& Discussion}

\subsection{Zero-shot Semantic Segmentation}
\label{sec:zeroshot_semantic_segmentation}

In the previous section, we described PACL, a multi-modal contrastive objective to train an alignment between vision patch embeddings and CLS text embeddings in CLIP. In this section, we evaluate the quality of this alignment through zero-shot transfer to semantic segmentation. We present implementation and training details for PACL, evaluation settings for zero-shot segmentation, and finally, results and a discussion on the same.

\textbf{Training a small vision embedder:} 
In \Cref{sec:semantic_coherence}, we have shown that CLIP's pre-trained vision encoders $\hat{f}_v$ have a relatively strong semantic coherence. In order to leverage this coherence and the large scale pre-training of CLIP, we keep the image encoder $\hat{f}_v$, the text encoder $f_t$ and the text embedder $e_t$ frozen from a pre-trained CLIP model. We only train the vision embedder, i.e., $\theta = \{\hat{e}_v\}$. Note that the modification of the vision encoder from $f_v$ (outputs the CLS vision token) to $\hat{f}_v$ (outputs the patch tokens) does not require any re-training. For $\hat{e}_v$, we use a residual block with two linear layers in the main branch and a single linear layer in the residual connection, There is a ReLU non-linearity between the two linear layers (see \Cref{app:vision_embedder_architecture}). We find this simple architecture to work well for our applications.

\textbf{Image-text datasets for training:} We train our model purely on publicly available image-text datasets. In particular, we use Google Conceptual Captions (GCC) 3M \cite{sharma2018conceptual}, Google Conceptual Captions (GCC) 12M \cite{changpinyo2021cc12m} and YFCC-15M, a subset of YFCC-100M \cite{thomee2016yfcc100m} provided by CLIP \cite{radford2021learning}, with a total number of approximately 30M training samples. Similar to GroupViT \cite{xu2022groupvit}, in addition to the text descriptions in the datasets, we extract nouns from these descriptions, and randomly select one of 7 CLIP prompts (like ``\emph{itap of a ().}", see \Cref{app:prompt_engg}), to form sentences with these nouns. We add these sentences to the text descriptions as well. More details on the datasets can be found in \Cref{app:training_eval_datasets}. Note that \textbf{\emph{we do not use any segmentation annotations or class-agnostic segmentation masks during training}}. Further training details are in \Cref{app:training_details}.

\textbf{Stride trick at inference:} Since CLIP ViT-B/16 and ViT-L/14 use either $16 \times 16$ or $14 \times 14$ patches, the number of tokens generated is much smaller than the number of pixels, which is a problem for fine-grained predictions in segmentation. One workaround is to upscale the image at inference time to a larger size. We however find instead that a change to the stride of the convolutional layer to extract image patches in ViT can provide better fine-grained patches at inference time. In particular, we use a stride of $4 \times 4$ and upscale the resulting segmentations to image size using bi-linear interpolation.

\textbf{Segmentation datasets for evaluation: } Similar to recent works \cite{xu2022groupvit, ghiasi2021open} on zero-shot semantic segmentation, we use the following datasets for evaluation: \textbf{a)} \emph{Pascal VOC} \cite{Everingham10} (PV-20): 20 foreground classes with 1449 validation images, \textbf{b)} \emph{Pascal Context} \cite{mottaghi_cvpr14} (PC-59): 59 classes with 5k validation images, \textbf{c)} \emph{COCO Stuff} \cite{caesar2018coco} (CS-171): 171 ``thing" or ``stuff" classes and 5k validation images, \textbf{d)} \emph{ADE20K} \cite{zhou2017scene} (A-150): 150 classes with 2k validation images. Further details on these datasets can be found in \Cref{app:training_eval_datasets}. For all datasets, we report the mean intersection over union (mIoU) \cite{Everingham10}, the most popular evaluation metric for semantic segmentation.

\textbf{Comparative Baselines:} We compare PACL with some of the most recently published methods on zero-shot semantic segmentation. In particular, we use \emph{LSeg} \cite{li2022language}, \emph{ViL-Seg} \cite{liu2022open}, \emph{GroupViT} \cite{xu2022groupvit} and \emph{OpenSeg} \cite{ghiasi2021open} as comparative baselines. In addition, we also compare with two relatively older approaches in zero-shot segmentation: \emph{SPNet} \cite{xian2019semantic} and \emph{ZS3Net} \cite{bucher2019zero}. Note that some of these methods work under relatively relaxed constraints. In particular, SPNet, ZS3Net and LSeg use full segmentation annotations during training and OpenSeg uses class-agnostic segmentation masks. Furthermore, unlike us, ViL-Seg, SPNet and ZS3Net evaluate on a small subset of ``unseen" classes from Pascal VOC, Pascal Context and COCO Stuff. To our knowledge, GroupViT and ViL-Seg are the only two methods which solely use image-text data for training. We also add a baseline using vanilla CLIP by taking the alignment between the vision patch embeddings and the text CLS embedding from CLIP's pre-trained model.

\begin{figure}[!t]
    \centering
    \begin{subfigure}{0.17\linewidth}
        \centering
        \includegraphics[width=\linewidth]{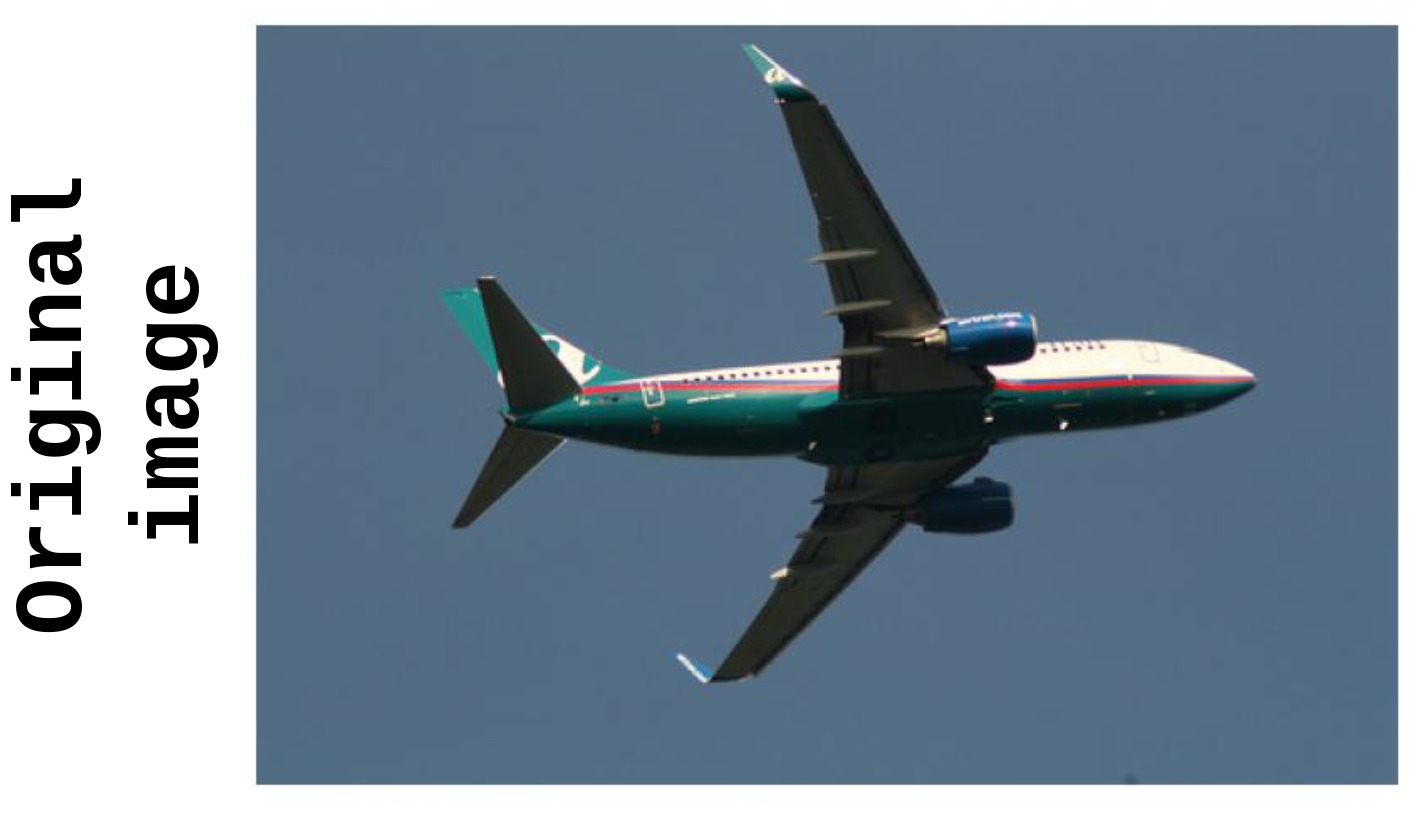}
        \vspace{-5mm}
    \end{subfigure}
    \begin{subfigure}{0.132\linewidth}
        \centering
        \includegraphics[width=\linewidth]{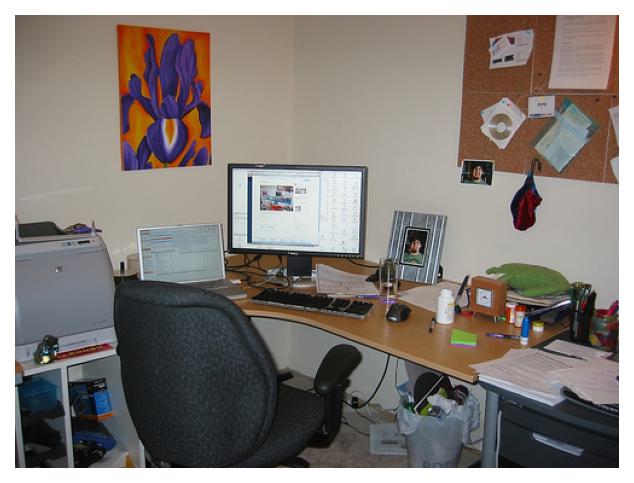}
        \vspace{-5mm}
    \end{subfigure}
    \begin{subfigure}{0.142\linewidth}
        \centering
        \includegraphics[width=\linewidth]{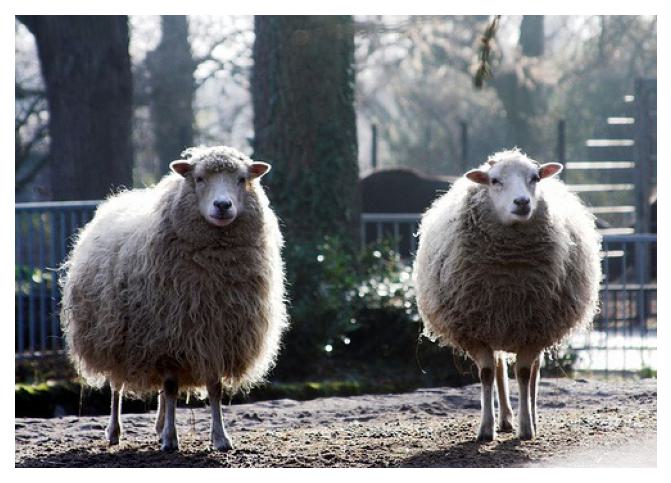}
        \vspace{-5mm}
    \end{subfigure}
    \begin{subfigure}{0.142\linewidth}
        \centering
        \includegraphics[width=\linewidth]{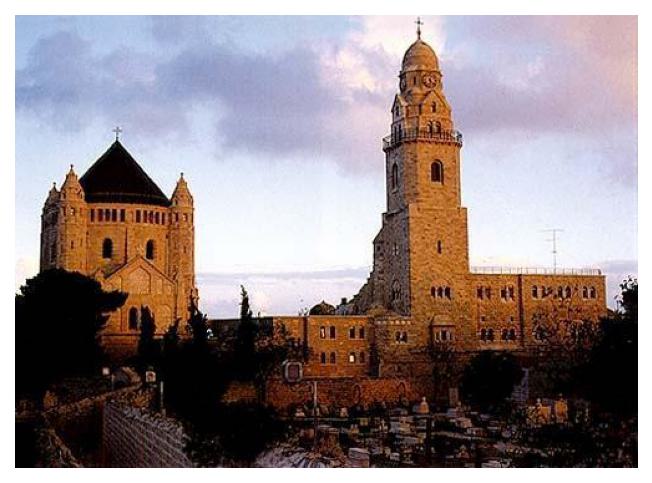}
        \vspace{-5mm}
    \end{subfigure}
    \begin{subfigure}{0.136\linewidth}
        \centering
        \includegraphics[width=\linewidth]{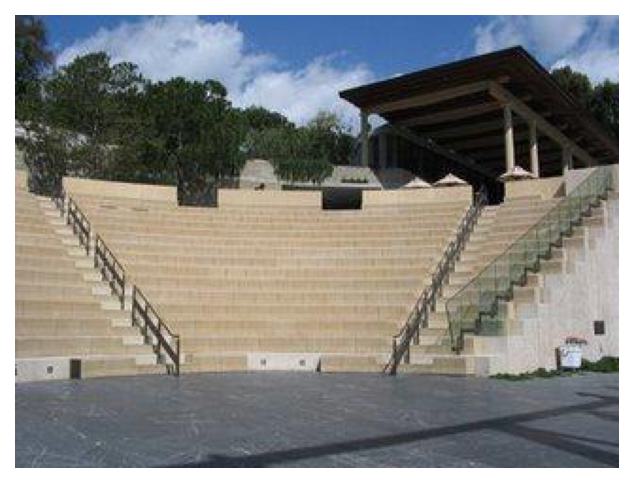}
        \vspace{-5mm}
    \end{subfigure}
    \begin{subfigure}{0.135\linewidth}
        \centering
        \includegraphics[width=\linewidth]{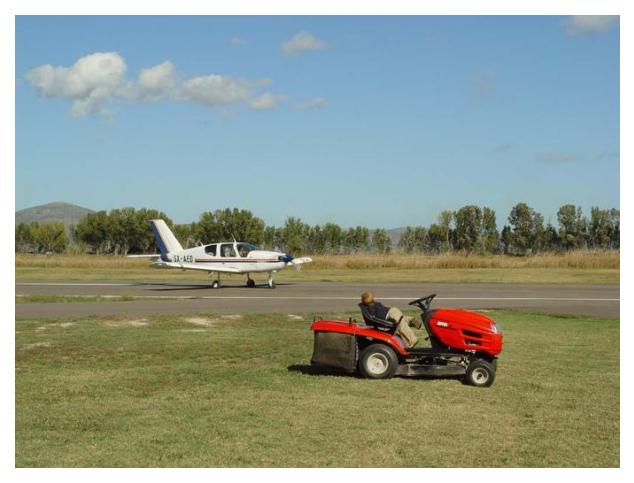}
        \vspace{-5mm}
    \end{subfigure}

    \begin{subfigure}{0.17\linewidth}
        \centering
        \includegraphics[width=\linewidth]{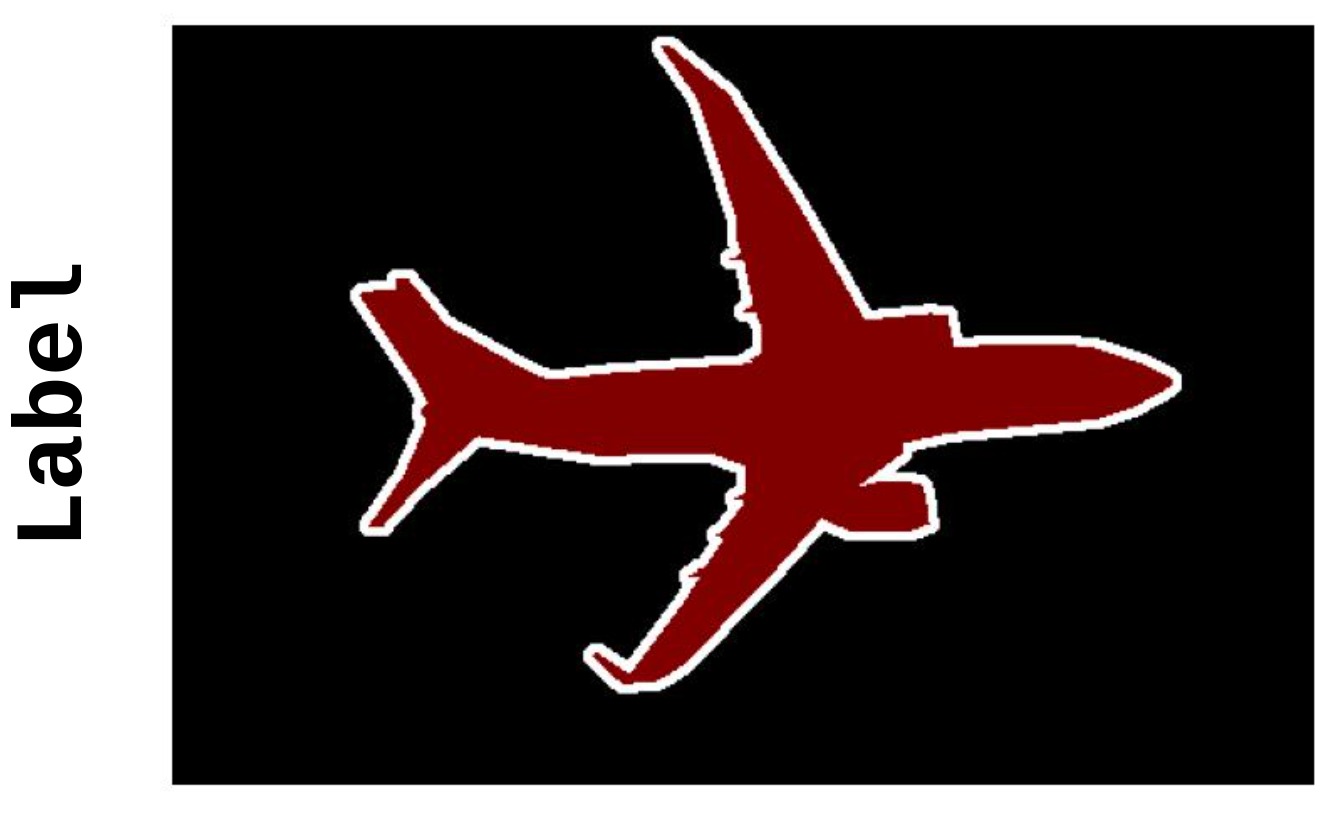}
        \vspace{-5mm}
    \end{subfigure}
    \begin{subfigure}{0.132\linewidth}
        \centering
        \includegraphics[width=\linewidth]{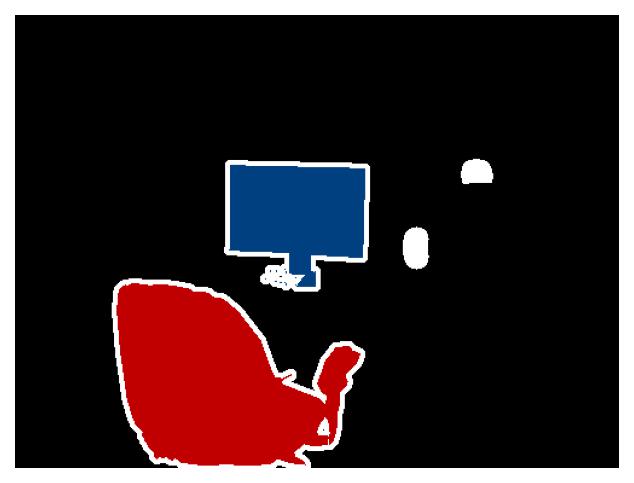}
        \vspace{-5mm}
    \end{subfigure}
    \begin{subfigure}{0.142\linewidth}
        \centering
        \includegraphics[width=\linewidth]{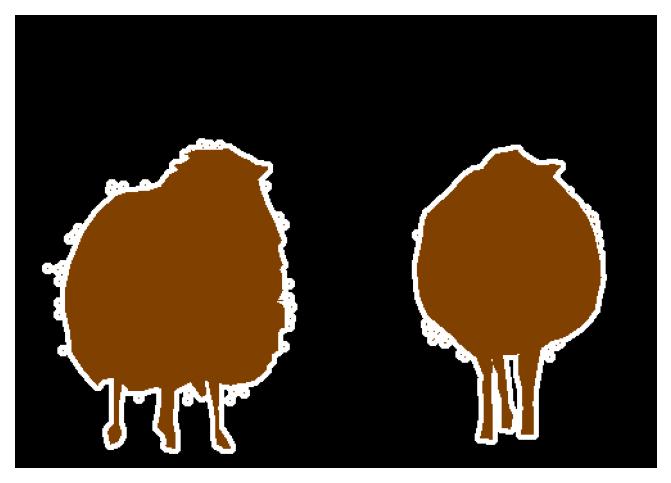}
        \vspace{-5mm}
    \end{subfigure}
    \begin{subfigure}{0.142\linewidth}
        \centering
        \includegraphics[width=\linewidth]{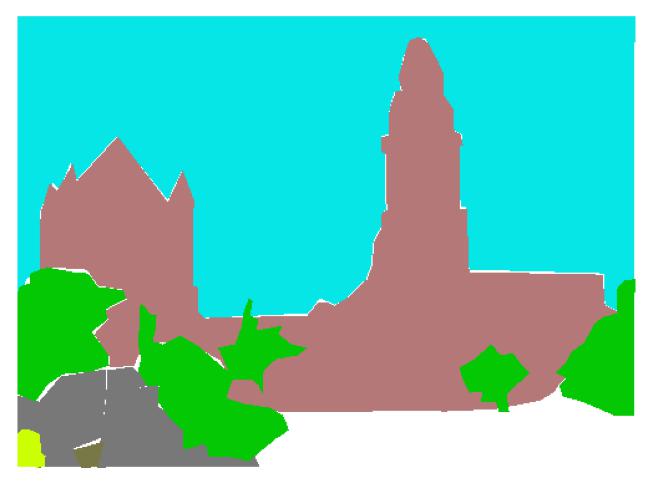}
        \vspace{-5mm}
    \end{subfigure}
    \begin{subfigure}{0.136\linewidth}
        \centering
        \includegraphics[width=\linewidth]{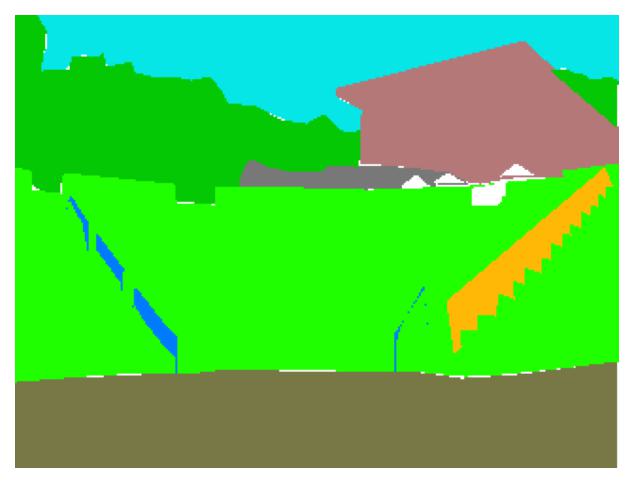}
        \vspace{-5mm}
    \end{subfigure}
    \begin{subfigure}{0.135\linewidth}
        \centering
        \includegraphics[width=\linewidth]{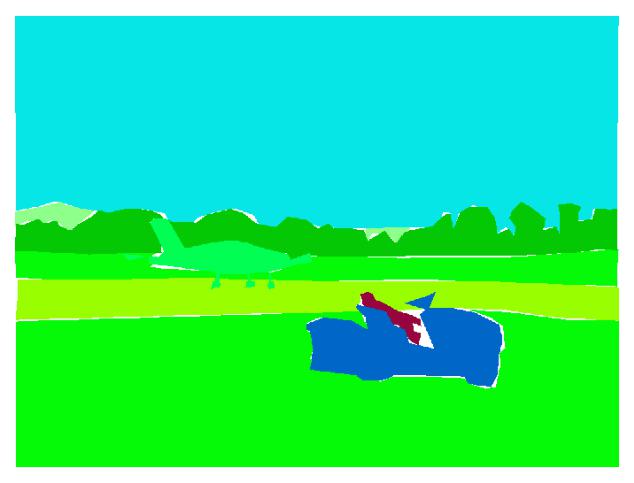}
        \vspace{-5mm}
    \end{subfigure}

    \begin{subfigure}{0.17\linewidth}
        \centering
        \includegraphics[width=\linewidth]{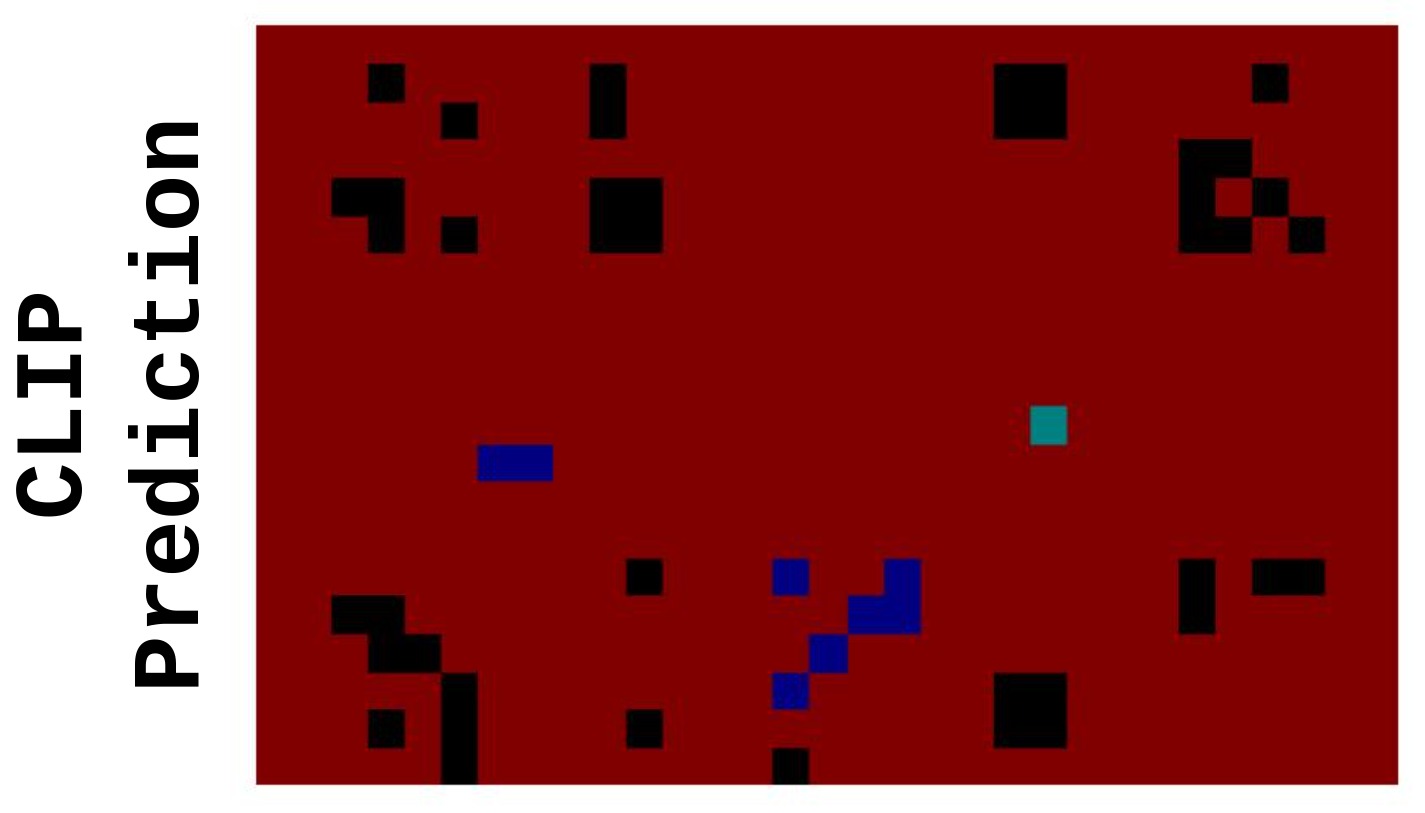}
        \vspace{-5mm}
    \end{subfigure}
    \begin{subfigure}{0.132\linewidth}
        \centering
        \includegraphics[width=\linewidth]{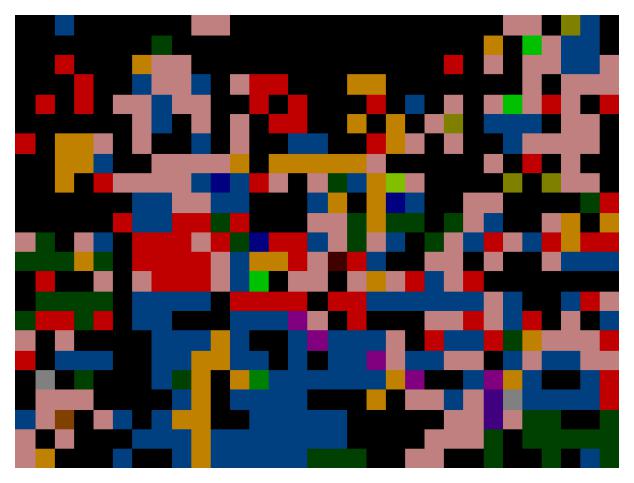}
        \vspace{-5mm}
    \end{subfigure}
    \begin{subfigure}{0.142\linewidth}
        \centering
        \includegraphics[width=\linewidth]{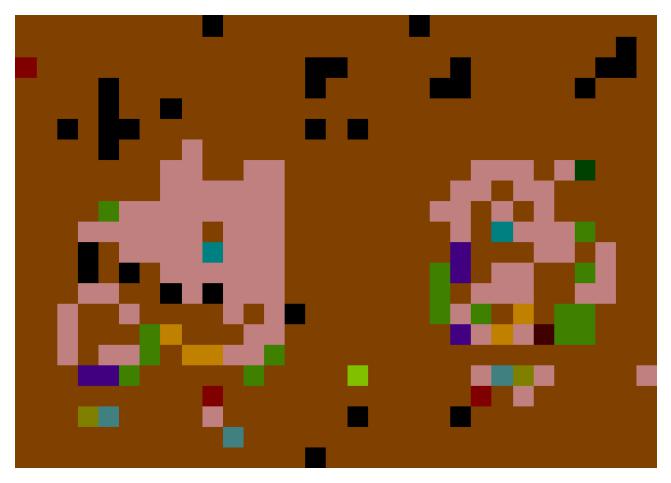}
        \vspace{-5mm}
    \end{subfigure}
    \begin{subfigure}{0.142\linewidth}
        \centering
        \includegraphics[width=\linewidth]{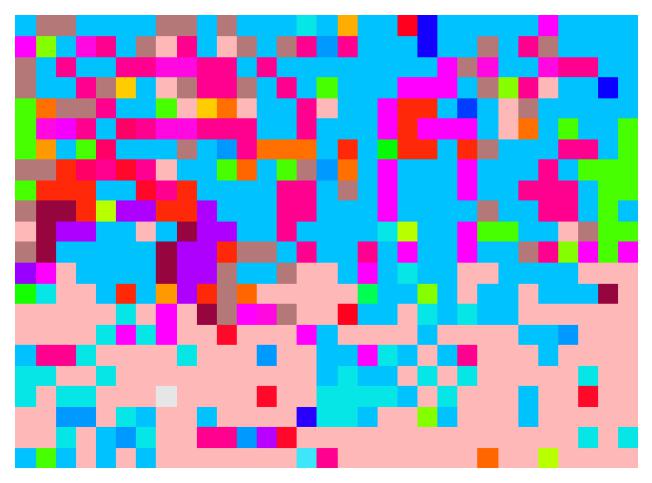}
        \vspace{-5mm}
    \end{subfigure}
    \begin{subfigure}{0.136\linewidth}
        \centering
        \includegraphics[width=\linewidth]{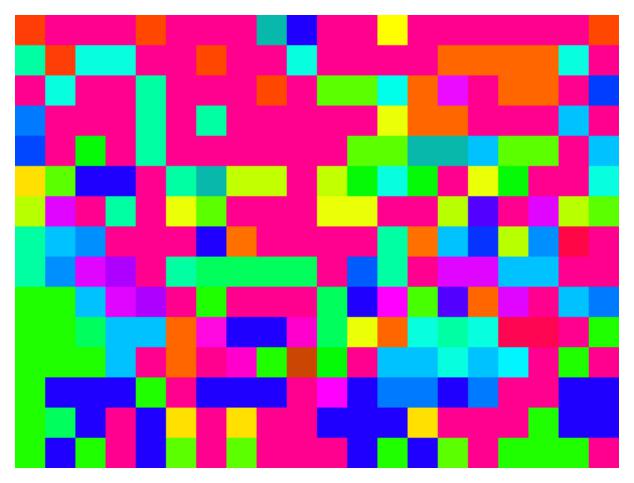}
        \vspace{-5mm}
    \end{subfigure}
    \begin{subfigure}{0.135\linewidth}
        \centering
        \includegraphics[width=\linewidth]{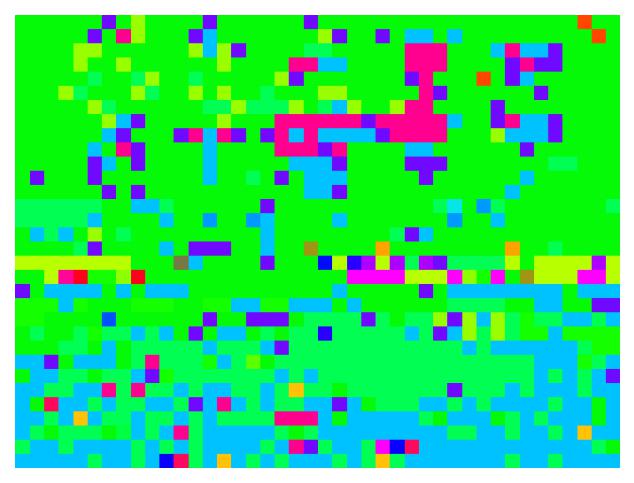}
        \vspace{-5mm}
    \end{subfigure}

    \begin{subfigure}{0.17\linewidth}
        \centering
        \includegraphics[width=\linewidth]{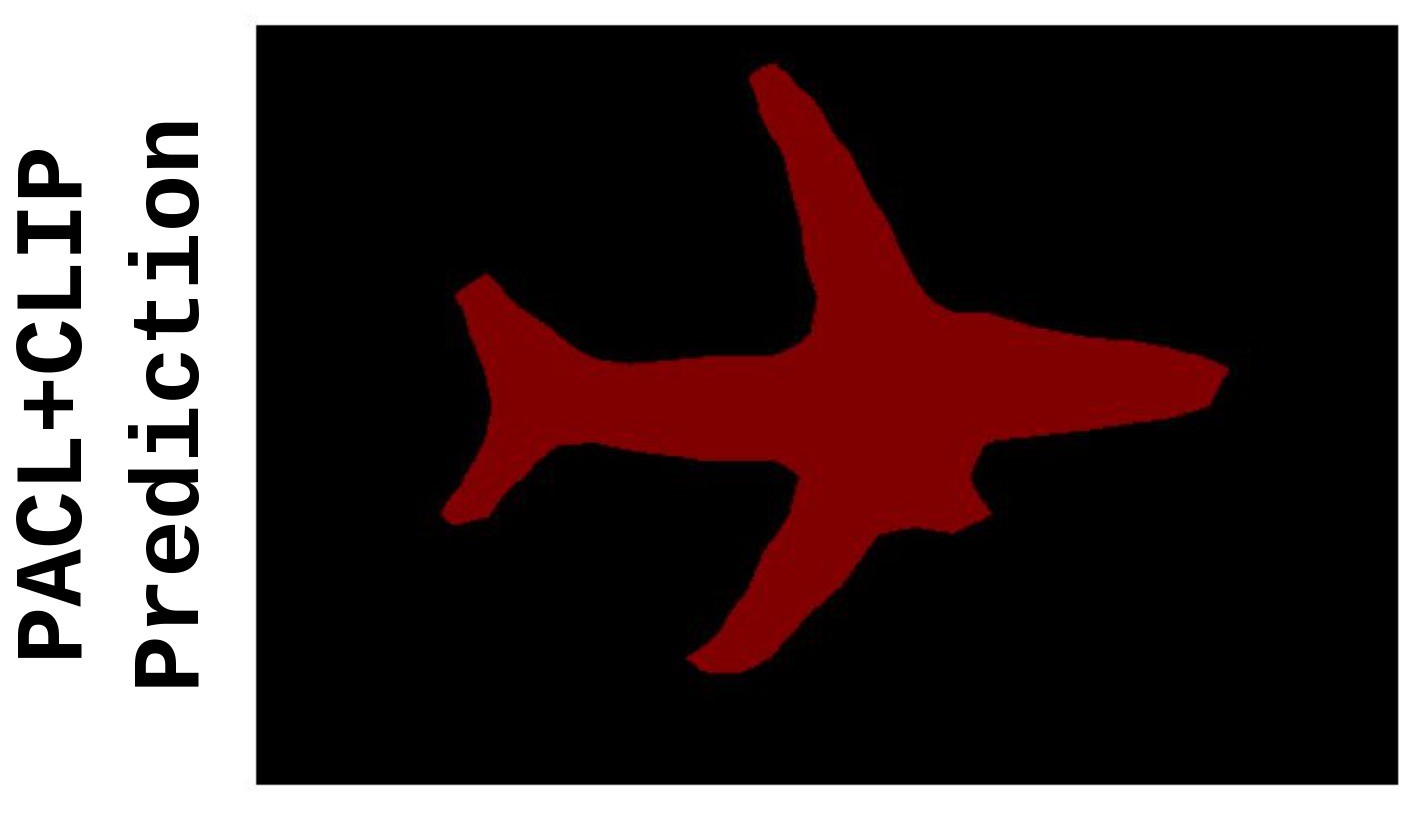}
        \vspace{-5mm}
    \end{subfigure}
    \begin{subfigure}{0.132\linewidth}
        \centering
        \includegraphics[width=\linewidth]{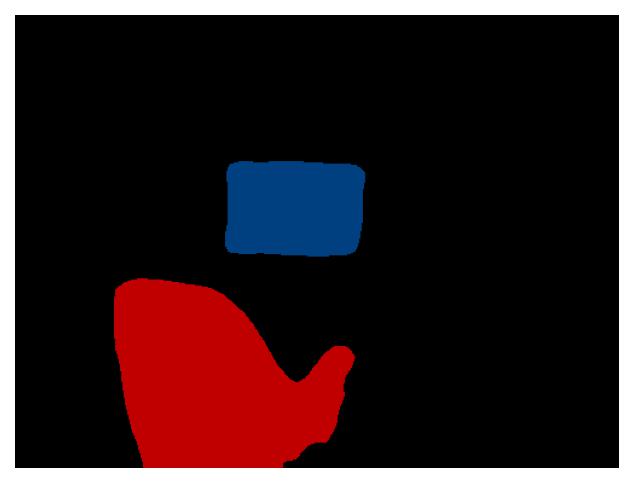}
        \vspace{-5mm}
    \end{subfigure}
    \begin{subfigure}{0.142\linewidth}
        \centering
        \includegraphics[width=\linewidth]{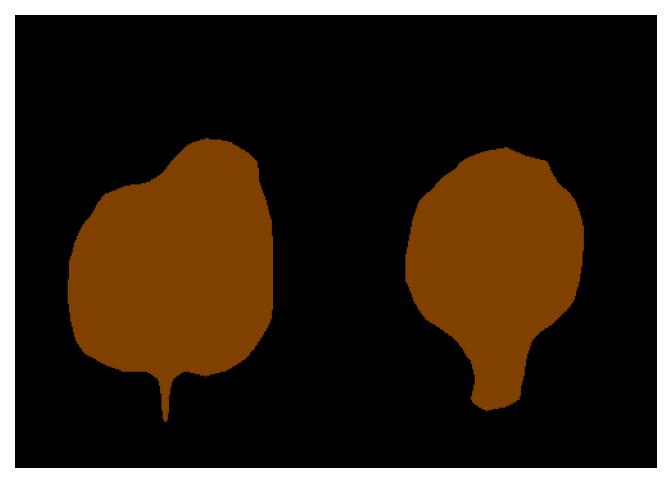}
        \vspace{-5mm}
    \end{subfigure}
    \begin{subfigure}{0.142\linewidth}
        \centering
        \includegraphics[width=\linewidth]{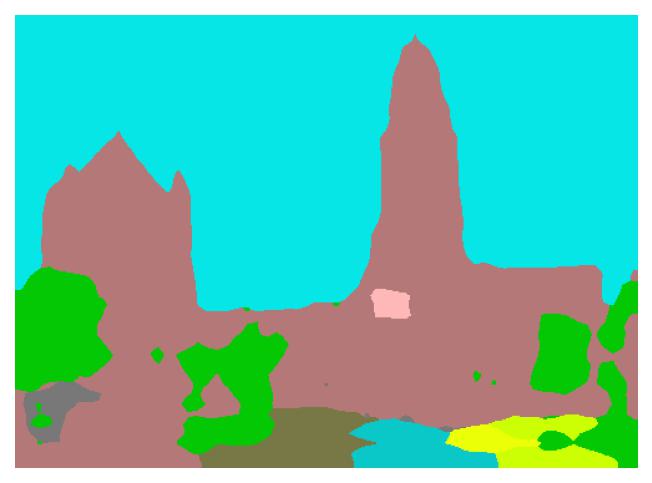}
        \vspace{-5mm}
    \end{subfigure}
    \begin{subfigure}{0.136\linewidth}
        \centering
        \includegraphics[width=\linewidth]{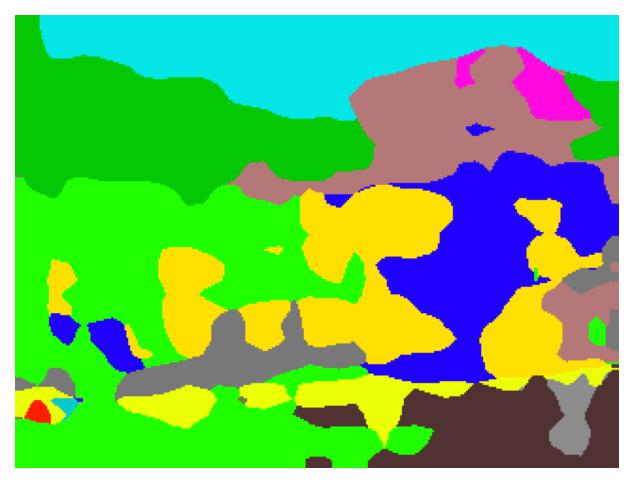}
        \vspace{-5mm}
    \end{subfigure}
    \begin{subfigure}{0.135\linewidth}
        \centering
        \includegraphics[width=\linewidth]{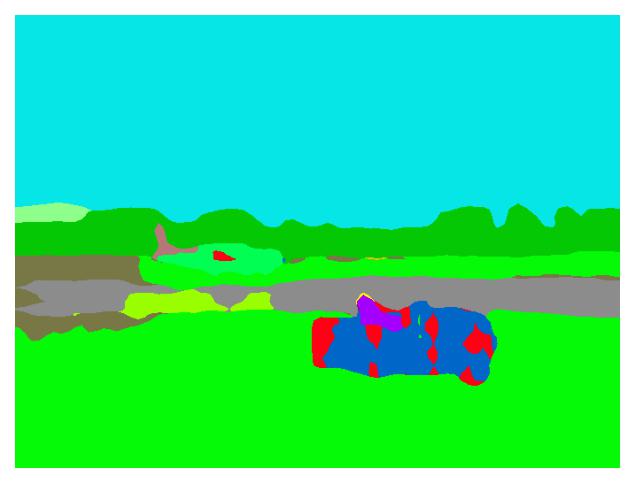}
        \vspace{-5mm}
    \end{subfigure}

    \caption{\textbf{Qualitative results on zero-shot semantic segmentation.} The first row denotes the original images, the second row shows the corresponding labels, the third row shows results obtained from a vanilla CLIP ViT-B/16, and the fourth row shows results of our method, PACL trained on a CLIP ViT-B/16 encoder. The first 3 images from the left are from Pascal VOC and the next 3 images are from ADE20K.}
    \vspace{-4mm}
    \label{fig:zeroshot_segmentation_qualitative}
\end{figure}

\begin{table*}[!t]
    \centering
    \scriptsize
    \resizebox{0.8\linewidth}{!}
    {
    \begin{tabular}{ccc|cc|cccc}
    \toprule
    & & \textbf{External} & \multicolumn{2}{c}{\textbf{Constraints}} & \multicolumn{4}{c}{\textbf{mIoU}} \\
    \textbf{Method} & \textbf{Encoder} & \textbf{Training Set} & \textbf{\textit{Annotation}} & \textbf{\textit{Mask}} & \textbf{\textit{PV-20}} \cite{Everingham10} & \textbf{\textit{PC-59}} \cite{mottaghi_cvpr14} & \textbf{\textit{CS-171}} \cite{caesar2018coco} & \textbf{\textit{A-150}} \cite{zhou2017scene} \\
    \midrule
    SPNet \cite{xian2019semantic} & ResNet-101 & \xmark & \cmark & \xmark & $15.6$ & $4.0$ & $8.7$ & - \\
    ZS3Net \cite{bucher2019zero} & ResNet-101 & \xmark & \cmark & \xmark & $17.7$ & $7.7$ & $9.6$ & - \\
    LSeg \cite{li2022language} & ViT-L/16 & \xmark & \cmark & \xmark & $52.3$ & - & - & - \\
    OpenSeg \cite{ghiasi2021open} & EfficientNet-B7 & COCO \cite{chen2015microsoft} + Loc. Narr. \cite{PontTuset_eccv2020} & \xmark & \cmark & $72.2$ & $48.2$ & - & $28.6$ \\
    ViL-Seg \cite{liu2022open} & ViT-B/16 & GCC12M \cite{changpinyo2021cc12m} & \xmark & \xmark & $34.4$ & $16.3$ & $16.4$ & - \\
    GroupViT \cite{xu2022groupvit} & ViT-S/16 & GCC12M \cite{changpinyo2021cc12m} + YFCC15M \cite{thomee2016yfcc100m, radford2021learning} & \xmark & \xmark & $52.3$ & $22.4$ & $24.3$ & - \\
    \midrule
    CLIP \cite{radford2021learning} & ViT-B/16 & WIT-400M \cite{radford2021learning} & \xmark & \xmark & $8.4$ & $2.3$ & $2.6$ & $1.3$ \\
    \midrule
    \textbf{CLIP + PACL (Ours)} & ViT-B/16 & GCC3M \cite{sharma2018conceptual} + GCC12M \cite{changpinyo2021cc12m} + YFCC15M \cite{thomee2016yfcc100m, radford2021learning} & \xmark & \xmark & $\mathbf{72.3}$ & $\mathbf{50.1}$ & $\mathbf{38.8}$ & $\mathbf{31.4}$ \\
    \bottomrule
    \end{tabular}
    }
    \vspace{-2mm}
    \caption{\textbf{Results on zero-shot semantic segmentation} on Pascal VOC (PV-20), Pascal Context (PC-59) and COCO Stuff (CS-171) and ADE20K (A-150) datasets. We provide the encoder architecture, external training dataset (if any) as well as if those methods use segmentation annotations or class-agnostic segmentation masks. Our method (CLIP + PACL) consistently outperforms all previous approaches.}
    \label{table:zeroshot_seg_quantitative}
    \vspace{-4mm}
\end{table*}

\textbf{Results \& discussion:} In \Cref{table:zeroshot_seg_quantitative}, we report the mIoU for each baseline on the 4 segmentation datasets mentioned above. Note that the numbers shown for SPNet, ZS3Net and ViL-Seg are obtained from the ViL-Seg paper \cite{liu2022open} and the numbers for all other baselines are obtained from their respective papers (cited in the table). In \Cref{fig:zeroshot_segmentation_qualitative}, we show qualitative results of our method (i.e., PACL + CLIP) on PascalVOC and ADE20K images (more in \Cref{app:qualitative_segmentation_results}). With mIoU scores of $72.3$, $50.1$, $38.8$ and $31.4$ on Pascal VOC, Pascal Context, COCO Stuff and ADE20K respectively, it is clear that \emph{PACL outperforms all other baselines consistently even though it works under a stricter set of assumptions}, i.e., it does not use any segmentation annotations and is evaluated on all classes of the segmentation datasets. This is further corroborated from our qualitative results in \Cref{fig:zeroshot_segmentation_qualitative}. It is interesting to note from \Cref{fig:zeroshot_segmentation_qualitative} that vanilla CLIP mostly seems to identify the correct classes in its predictions, just not the locations of those classes within the image. This relates to the problem of a lack of alignment between the CLS text token and the vision patch tokens which we had seen earlier (see \Cref{fig:alignment_qualitative}) and this problem is solved through the introduction of the PACL contrastive objective. Since PACL, as an approach, is not tied to any particular encoder, we next test its performance using different pre-trained encoders as well as different datasets on the zero-shot segmentation task.

\textbf{Ablations on datasets and encoders:} We perform an ablation by training PACL on a combination of different image-text training sets and different pre-trained vision encoders. For vision encoders we use CLIP ViT-B/16, CLIP ViT-L/14 and also use DINO's \cite{caron2021emerging} ViT-B/16 models. For training sets, we use GCC12M, (GCC12M + YFCC15M) and (GCC3M + GCC12M + YFCC15M). We report the mIoU obtained on Pascal VOC from each of the (model, dataset) combinations in \Cref{table:zeroshot_seg_ablations}.

\begin{figure}[!t]
    \centering
    \begin{subfigure}{0.21\linewidth}
        \centering
        \includegraphics[width=\linewidth]{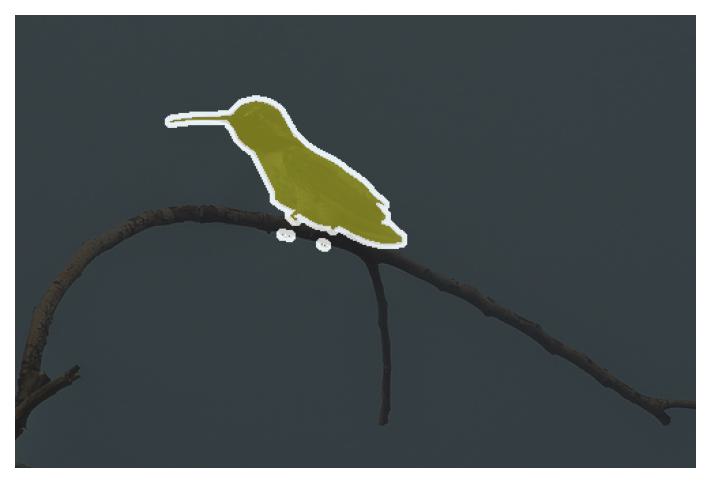}
        \vspace{-5mm}
    \end{subfigure}
    \begin{subfigure}{0.21\linewidth}
        \centering
        \includegraphics[width=\linewidth]{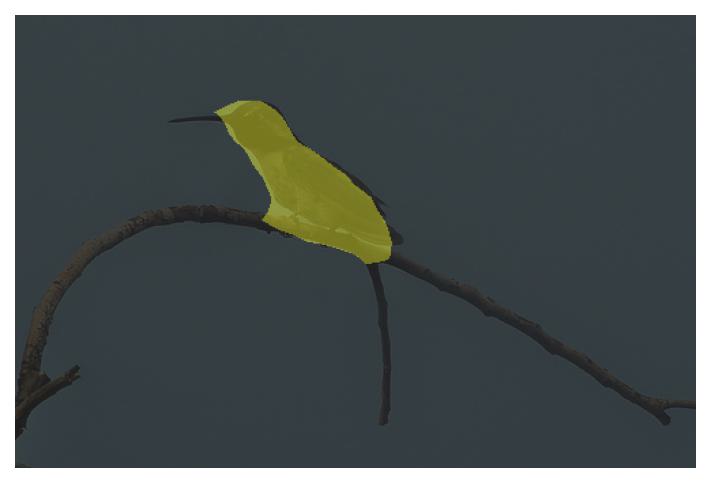}
        \vspace{-5mm}
    \end{subfigure}
    \begin{subfigure}{0.21\linewidth}
        \centering
        \includegraphics[width=\linewidth]{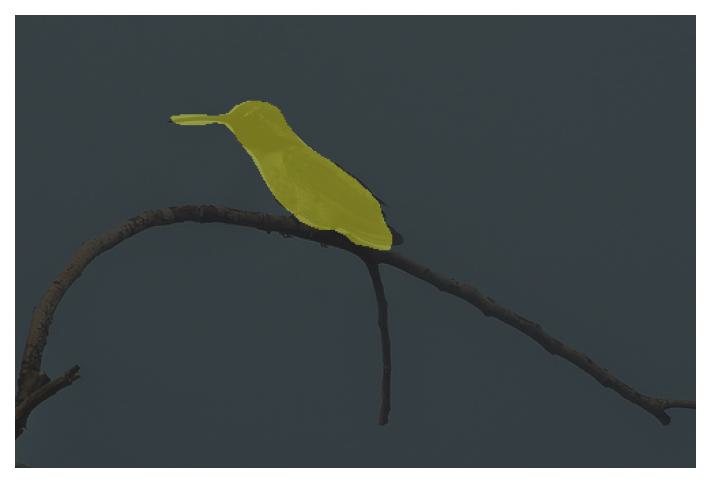}
        \vspace{-5mm}
    \end{subfigure}
    \begin{subfigure}{0.21\linewidth}
        \centering
        \includegraphics[width=\linewidth]{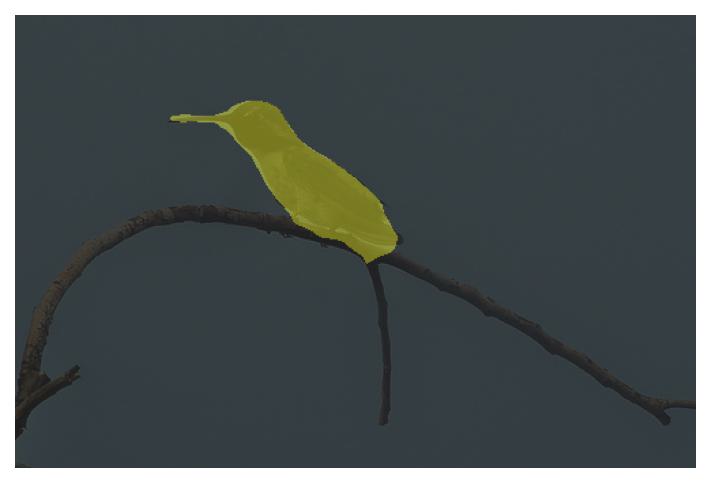}
        \vspace{-5mm}
    \end{subfigure}

    \begin{subfigure}{0.21\linewidth}
        \centering
        \includegraphics[width=\linewidth]{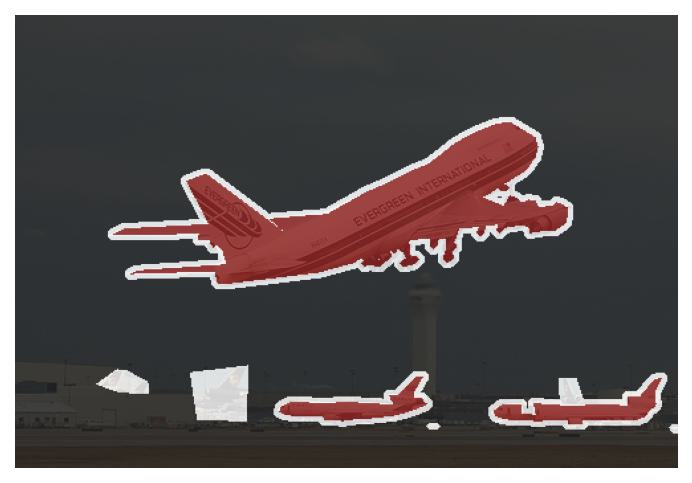}
        \vspace{-5mm}
    \end{subfigure}
    \begin{subfigure}{0.21\linewidth}
        \centering
        \includegraphics[width=\linewidth]{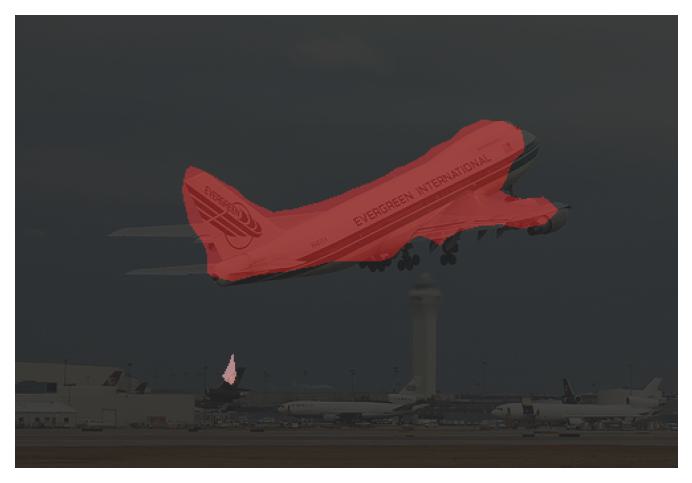}
        \vspace{-5mm}
    \end{subfigure}
    \begin{subfigure}{0.21\linewidth}
        \centering
        \includegraphics[width=\linewidth]{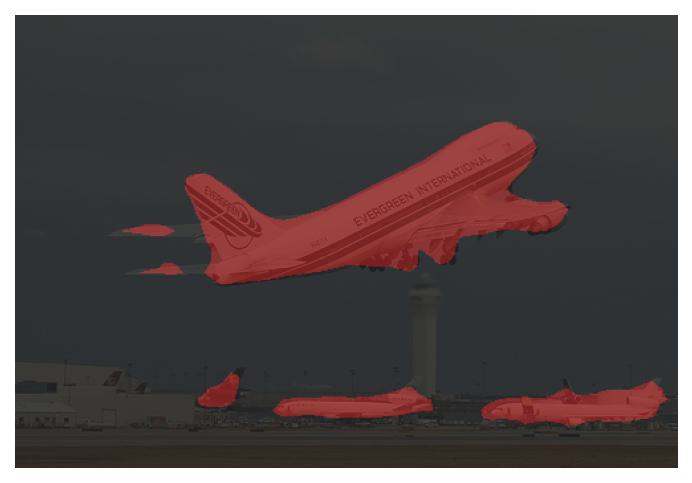}
        \vspace{-5mm}
    \end{subfigure}
    \begin{subfigure}{0.21\linewidth}
        \centering
        \includegraphics[width=\linewidth]{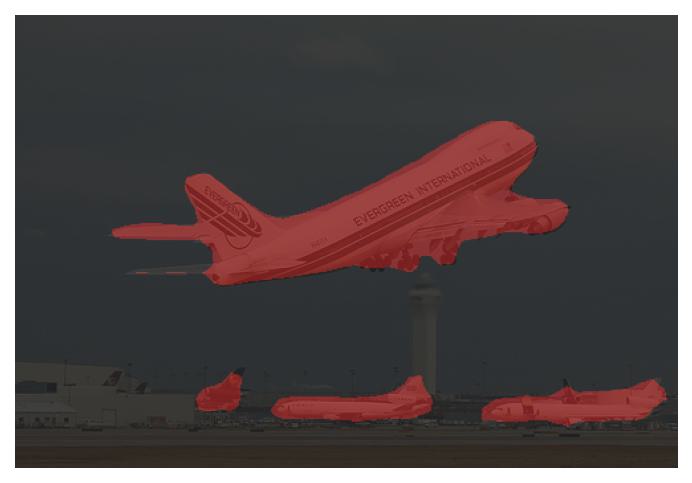}
        \vspace{-5mm}
    \end{subfigure}

    \begin{subfigure}{0.21\linewidth}
        \centering
        \includegraphics[width=\linewidth]{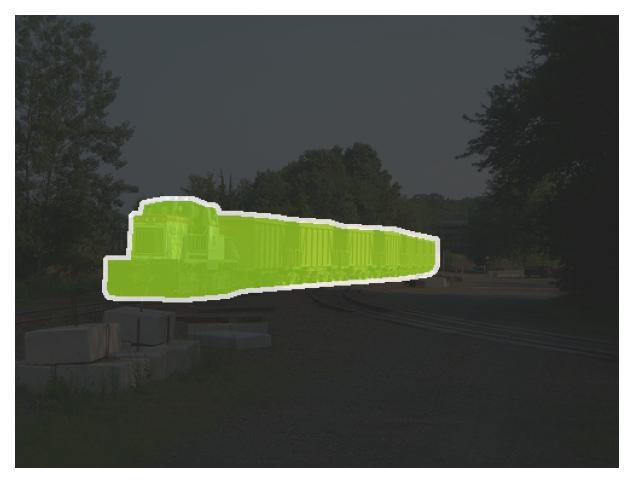}
        \caption{GT}
        \vspace{-3mm}
    \end{subfigure}
    \begin{subfigure}{0.21\linewidth}
        \centering
        \includegraphics[width=\linewidth]{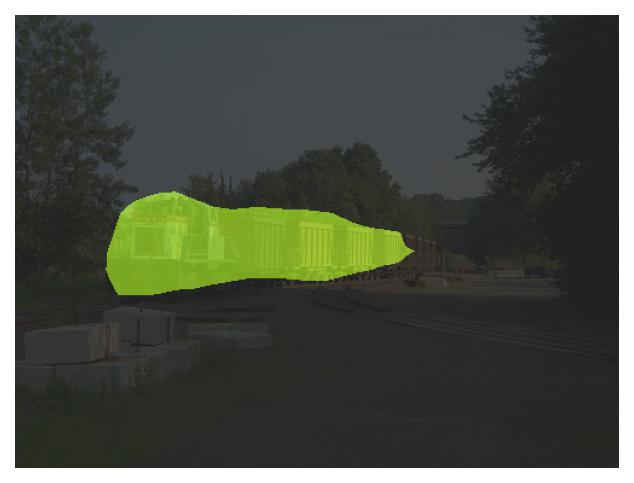}
        \caption{DINO B/16}
        \vspace{-3mm}
    \end{subfigure}
    \begin{subfigure}{0.21\linewidth}
        \centering
        \includegraphics[width=\linewidth]{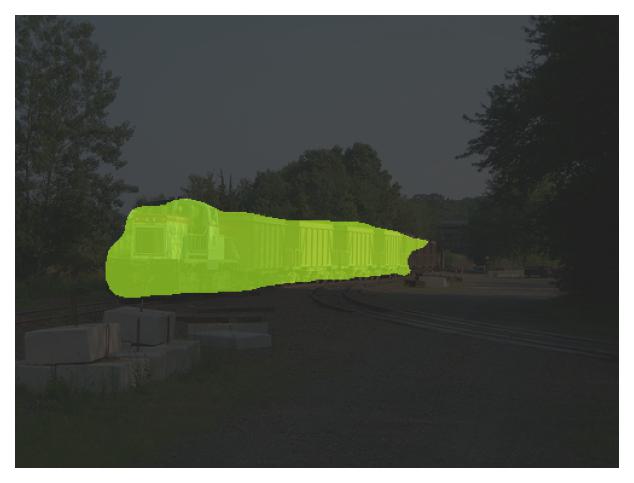}
        \caption{CLIP L/14}
        \vspace{-3mm}
    \end{subfigure}
    \begin{subfigure}{0.21\linewidth}
        \centering
        \includegraphics[width=\linewidth]{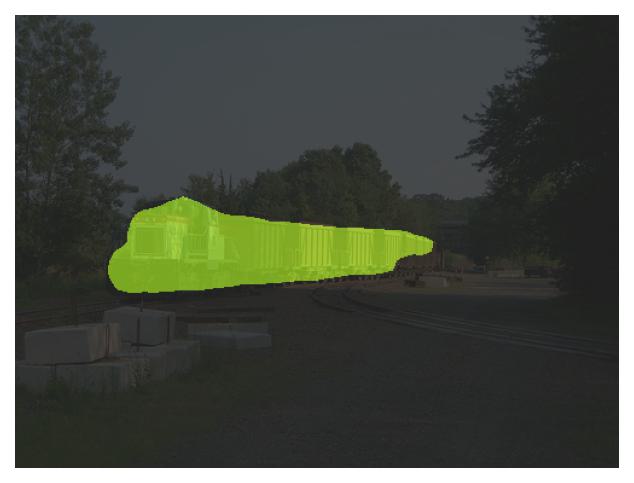}
        \caption{CLIP B/16}
        \vspace{-3mm}
    \end{subfigure}

    \caption{\textbf{Qualitative results comparing segmentation of different encoders using PACL.} We use 3 images from PASCAL VOC val set and show their segmentations for DINO ViT-B/16, CLIP ViT-L/14 and CLIP ViT-B/16.}
    \vspace{-2mm}
    \label{fig:segmentation_ablation_qualitative}
\end{figure}

These results provide two surprising observations. Firstly, PACL seems to generate an alignment even between DINO's pre-trained vision encoder and CLIP's text encoder, although these encoders have been trained separately and independently of each other. With an mIoU of 55.4, even the worst performing DINO baseline outperforms all competitive zero-shot segmentation baselines in \Cref{table:zeroshot_seg_quantitative} except OpenSeg. Secondly, PACL trained using CLIP's ViT-B/16 consistently outperforms ViT-L/14 even though ViT-L/14 is known to be a clear winner in terms of image level zero-shot tasks. In fact, there is a trend in performance where CLIP ViT-B/16 outperforms CLIP ViT-L/14 which outperforms DINO ViT-B/16. This is also noticeable in \Cref{fig:segmentation_ablation_qualitative} where CLIP encoders generate relatively better segmentation masks than DINO. This observation is strongly reminiscent of the one in \Cref{sec:semantic_coherence} and \Cref{fig:roc_coherence}, where we note that semantic coherence\footnote{Semantic coherence is the property which enables a vision encoder to generate similar patch/token level representations for semantically similar regions of an image.} is strongest in CLIP ViT-B/16 followed by CLIP ViT-L/14 and finally by DINO ViT-B/16. These empirical observations \emph{suggest that PACL is a general contrastive learning method which can be used to train a patch level alignment and works independent of vision and text encoders as long as the vision encoders exhibit the property of semantic coherence}. Indeed, semantic coherence seems to be the most important factor behind the success of PACL.

\begin{table}[!t]
    \centering
    \scriptsize
    \resizebox{0.8\linewidth}{!}
    {
    \begin{tabular}{cccc}
    \toprule
    \textbf{Dataset} & \textbf{Vision Encoder} & \textbf{Text Encoder} & \textbf{mIoU PV-20} \\
    \midrule
    \multirow{3}{*}{GCC12M} & CLIP B/16 & B/16 & $64.1$ \\
                            & CLIP L/14 & L/14 & $62.7$ \\
                            & DINO B/16 & B/16 & $55.4$ \\
    \midrule
    \multirow{3}{*}{GCC12M + YFCC15M} & CLIP B/16 & B/16 & $69.2$ \\
                                      & CLIP L/14 & L/14 & $68.4$ \\
                                      & DINO B/16 & B/16 & $62.6$ \\
    \midrule
    \multirow{3}{*}{\textbf{GCC3M + GCC12M + YFCC15M}} & \textbf{CLIP B/16} & \textbf{B/16} & $\mathbf{72.3}$ \\
                                              & CLIP L/14 & L/14 & $71.7$ \\
                                              & DINO B/16 & B/16 & $64.8$ \\
    \bottomrule
    \end{tabular}
    }
    \vspace{-2mm}
    \caption{\textbf{Ablation on zero-shot segmentation across encoder architectures and datasets} on Pascal VOC (PV-20). In the Text Encoder column, B/16(L/14) indicates the pre-trained text encoder trained for CLIP ViT-B/16(L/14).}
    \label{table:zeroshot_seg_ablations}
    \vspace{-3mm}
\end{table}

\subsection {Image Classification \& Future Work}
\label{sec:zeroshot_classification}

In \Cref{sec:pacl}, we mention that the modified compatibility function of PACL can be used to make image level predictions, similar to CLIP. In this section, we test our PACL models on zero-shot image classification. We then end with a discussion of possible future avenues from our work.

\begin{figure}[!t]
    \centering
    \begin{subfigure}{0.48\linewidth}
        \centering
        \includegraphics[width=\linewidth]{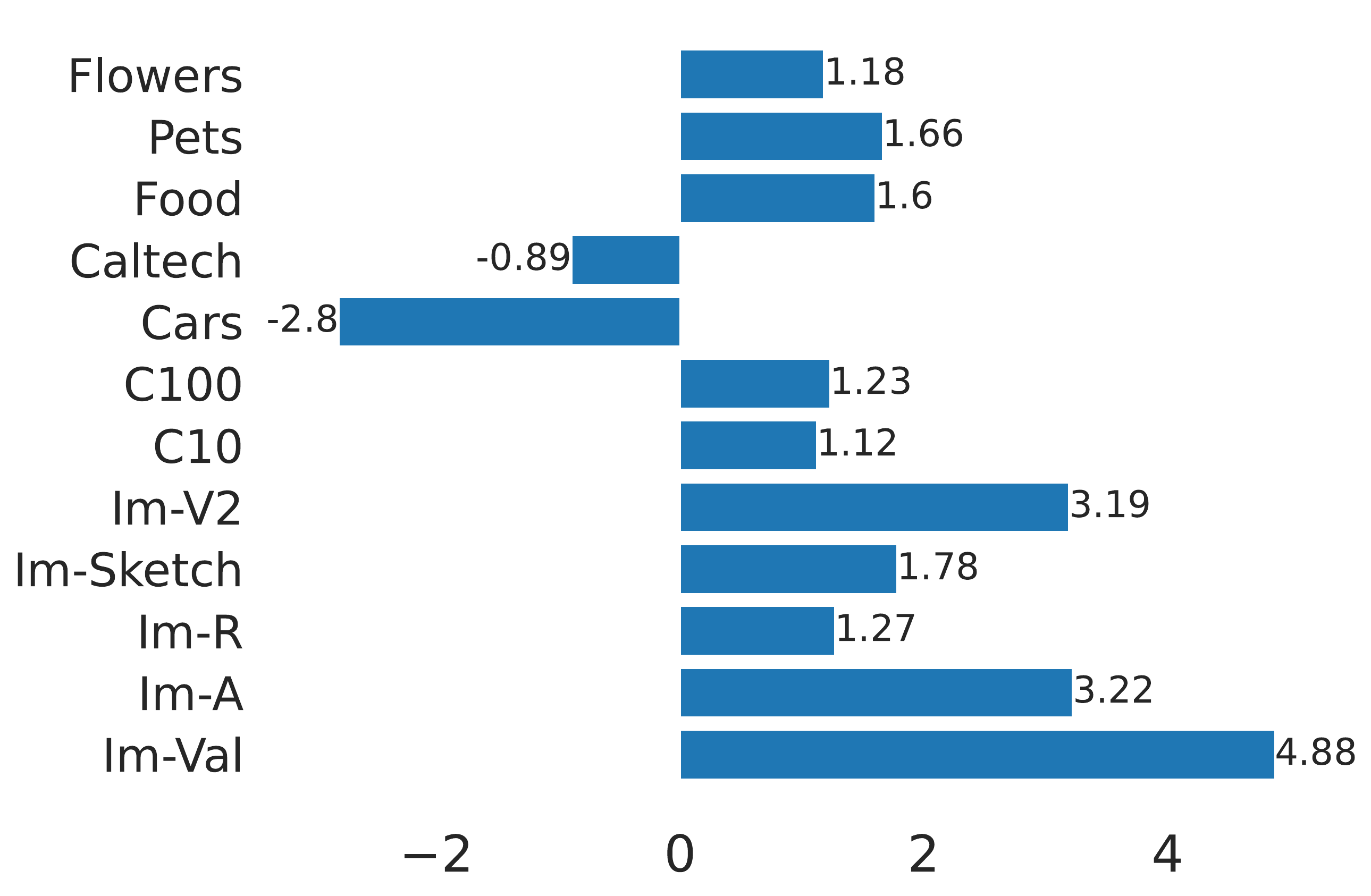}
        \caption{ViT-B/16}
        \label{subfig:diff_plot_vit_b}
    \end{subfigure}
    \begin{subfigure}{0.48\linewidth}
        \centering
        \includegraphics[width=\linewidth]{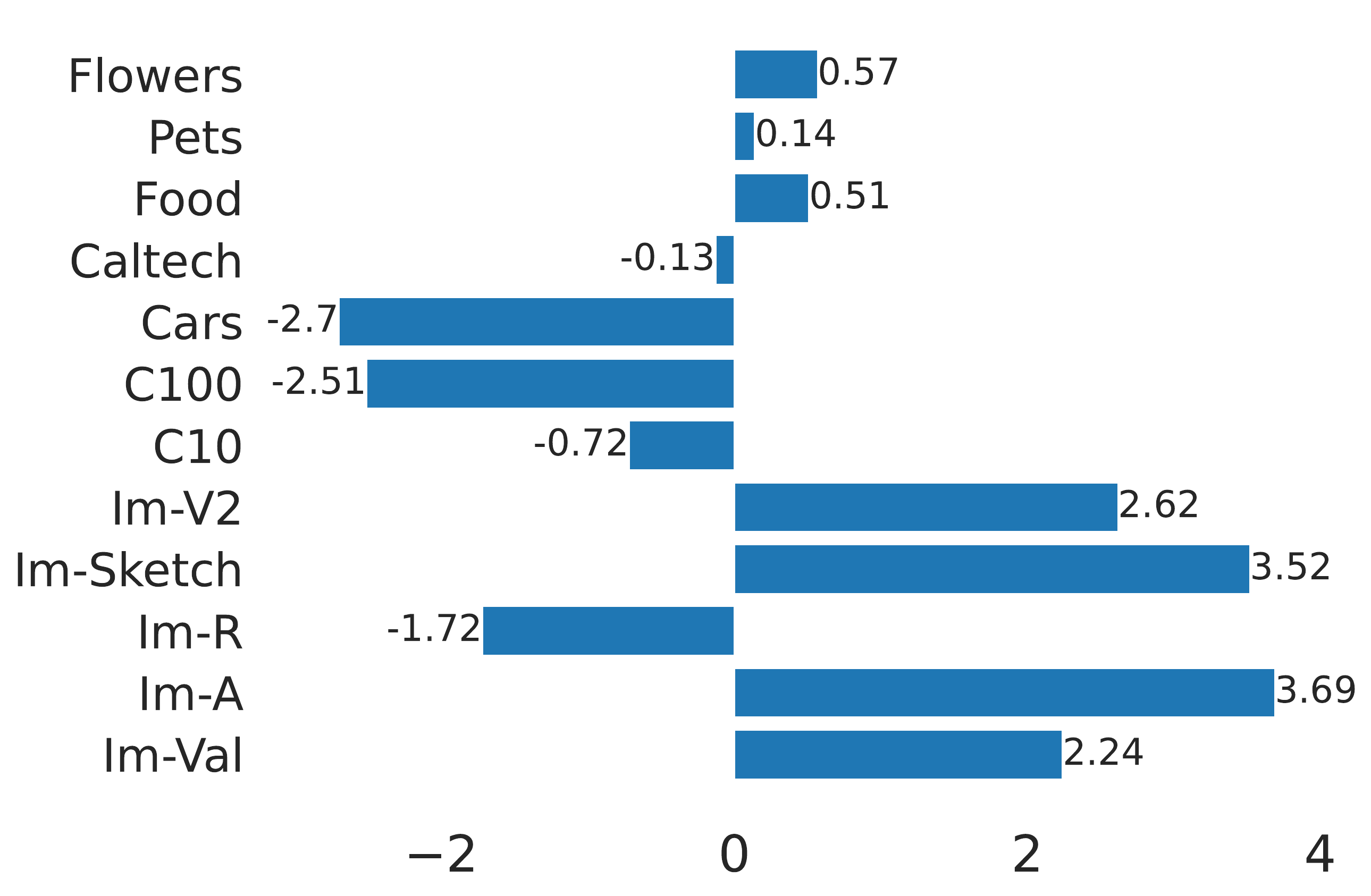}
        \caption{ViT-L/14}
        \label{subfig:diff_plot_vit_l}
    \end{subfigure}
    \vspace{-2mm}
    \caption{\textbf{Zero-shot image classification performance of PACL + CLIP vs vanilla CLIP on 12 datasets.} PACL + CLIP is competitive with or outperforms CLIP on most datasets.}
    \vspace{-5mm}
    \label{fig:diff_plot_classification}
\end{figure}

\textbf{Zero-shot image classification results:} We apply PACL trained using CLIP ViT-B/16 and ViT-L/14 encoders on (GCC3M + GCC12M + YFCC15M) to zero-shot image classification on 12 different datasets including ImageNet \cite{deng2009imagenet}, 4 datasets considered to be standard distribution shifts on ImageNet: ImageNet-A \cite{hendrycks2021natural}, ImageNet-R \cite{hendrycks2021many}, ImageNet-Sketch \cite{wang2019learning} and ImageNet-V2 \cite{recht2019imagenet}, as well as 7 other standard classification datasets, detailed in \Cref{app:training_eval_datasets}. We report the difference in classification accuracy between PACL + CLIP and vanilla CLIP for all the datasets in \Cref{fig:diff_plot_classification} (all classification accuracies in \Cref{app:zeroshot_image_classification}). PACL + CLIP outperforms vanilla CLIP on 10 and 7 out of the 12 classification datasets for ViT-B/16 and ViT-L/14 encoders respectively. Also note that except on ImageNet-R for ViT-L/14, PACL consistently outperforms vanilla CLIP on ImageNet and its distribution shifts. This observation is encouraging as it provides evidence in favour of our approach being applicable for image level applications in addition to segmentation. In the remainder of this section, we discuss possible avenues for future research from our work.

\textbf{Exploring PACL for image-level applications:} As seen above, since PACL is a general compatibility function for contrastive loss, it can be applied to all image level tasks. We show this through zero-shot image classification. However, it would be interesting to further explore PACL as an independent contrastive learning method. In particular, training models from scratch on PACL instead of the standard CLIP loss might provide additional benefits in the context of general VLP tasks like image-text retrieval \cite{wang2016comprehensive}. Since our work is focused around zero-shot semantic segmentation, we keep this exploration out of the scope of this work and as potential avenue for future research.

\textbf{Exploring other ways to generate patch level alignment:} All the current methods on zero-shot open vocabulary segmentation, including ours, use CLIP like models, i.e., models with individual vision and text encoders with a fusion of modalities at the end of the encoders. However, there could be other ways of fusing modalities which could also lead to a generation of patch level alignment between image and text. In particular, one of the seemingly likely candidates of multi-modal fusion for generating patch level alignment could be cross-attention between image and text tokens, often seen in architectures used in VLP training \cite{kim2021vilt, li2021align, singh2022flava} etc. Studying the patch level alignment in these models to see if they can be transferred to dense prediction tasks is also an interesting area of future exploration. 

\section{Conclusion}
In this work, we explored \emph{Patch Aligned Contrastive Learning} (PACL), a modified compatibility function for image-text contrastive loss which learns an alignment between patch tokens obtained from a ViT based vision encoder and the CLS token from a text encoder. We show that such an alignment allows a model to identify regions of an image corresponding to a given text input, thereby enabling a seamless zero-shot transfer to the task of semantic segmentation, without requiring any segmentation annotations or masks during training. On 4 different segmentation datasets, we beat previous approaches on zero-shot open vocabulary segmentation, including the ones which use expensive segmentation annotations or masks for training. Finally, we show that PACL can also be used to make image level predictions and, when used with a pre-trained CLIP encoder, provides a general improvement in classification accuracy across 12 different image classification datasets.

{\small
\bibliographystyle{ieee_fullname}
\bibliography{egbib}

\begin{thebibliography}{10}\itemsep=-1pt

\bibitem{antol2015vqa}
Stanislaw Antol, Aishwarya Agrawal, Jiasen Lu, Margaret Mitchell, Dhruv Batra,
  C~Lawrence Zitnick, and Devi Parikh.
\newblock Vqa: Visual question answering.
\newblock In {\em Proceedings of the IEEE international conference on computer
  vision}, pages 2425--2433, 2015.

\bibitem{bossard14}
Lukas Bossard, Matthieu Guillaumin, and Luc Van~Gool.
\newblock Food-101 -- mining discriminative components with random forests.
\newblock In {\em European Conference on Computer Vision}, 2014.

\bibitem{bucher2019zero}
Maxime Bucher, Tuan-Hung Vu, Matthieu Cord, and Patrick P{\'e}rez.
\newblock Zero-shot semantic segmentation.
\newblock {\em Advances in Neural Information Processing Systems}, 32, 2019.

\bibitem{caesar2018coco}
Holger Caesar, Jasper Uijlings, and Vittorio Ferrari.
\newblock Coco-stuff: Thing and stuff classes in context.
\newblock In {\em Proceedings of the IEEE conference on computer vision and
  pattern recognition}, pages 1209--1218, 2018.

\bibitem{caron2021emerging}
Mathilde Caron, Hugo Touvron, Ishan Misra, Herv{\'e} J{\'e}gou, Julien Mairal,
  Piotr Bojanowski, and Armand Joulin.
\newblock Emerging properties in self-supervised vision transformers.
\newblock In {\em Proceedings of the IEEE/CVF International Conference on
  Computer Vision}, pages 9650--9660, 2021.

\bibitem{changpinyo2021cc12m}
Soravit Changpinyo, Piyush Sharma, Nan Ding, and Radu Soricut.
\newblock {Conceptual 12M}: Pushing web-scale image-text pre-training to
  recognize long-tail visual concepts.
\newblock In {\em CVPR}, 2021.

\bibitem{chen2022vlp}
Feilong Chen, Duzhen Zhang, Minglun Han, Xiuyi Chen, Jing Shi, Shuang Xu, and
  Bo Xu.
\newblock Vlp: A survey on vision-language pre-training.
\newblock {\em arXiv preprint arXiv:2202.09061}, 2022.

\bibitem{chen2017rethinking}
Liang-Chieh Chen, George Papandreou, Florian Schroff, and Hartwig Adam.
\newblock Rethinking atrous convolution for semantic image segmentation.
\newblock {\em arXiv preprint arXiv:1706.05587}, 2017.

\bibitem{chen2015microsoft}
Xinlei Chen, Hao Fang, Tsung-Yi Lin, Ramakrishna Vedantam, Saurabh Gupta, Piotr
  Doll{\'a}r, and C~Lawrence Zitnick.
\newblock Microsoft coco captions: Data collection and evaluation server.
\newblock {\em arXiv preprint arXiv:1504.00325}, 2015.

\bibitem{chen2022vision}
Zhe Chen, Yuchen Duan, Wenhai Wang, Junjun He, Tong Lu, Jifeng Dai, and Yu
  Qiao.
\newblock Vision transformer adapter for dense predictions.
\newblock {\em arXiv preprint arXiv:2205.08534}, 2022.

\bibitem{cho2021picie}
Jang~Hyun Cho, Utkarsh Mall, Kavita Bala, and Bharath Hariharan.
\newblock Picie: Unsupervised semantic segmentation using invariance and
  equivariance in clustering.
\newblock In {\em Proceedings of the IEEE/CVF Conference on Computer Vision and
  Pattern Recognition}, pages 16794--16804, 2021.

\bibitem{cordts2016cityscapes}
Marius Cordts, Mohamed Omran, Sebastian Ramos, Timo Rehfeld, Markus Enzweiler,
  Rodrigo Benenson, Uwe Franke, Stefan Roth, and Bernt Schiele.
\newblock The cityscapes dataset for semantic urban scene understanding.
\newblock In {\em Proceedings of the IEEE conference on computer vision and
  pattern recognition}, pages 3213--3223, 2016.

\bibitem{dai2021coatnet}
Zihang Dai, Hanxiao Liu, Quoc~V Le, and Mingxing Tan.
\newblock Coatnet: Marrying convolution and attention for all data sizes.
\newblock {\em Advances in Neural Information Processing Systems},
  34:3965--3977, 2021.

\bibitem{deng2009imagenet}
Jia Deng, Wei Dong, Richard Socher, Li-Jia Li, Kai Li, and Li Fei-Fei.
\newblock Imagenet: A large-scale hierarchical image database.
\newblock In {\em 2009 IEEE conference on computer vision and pattern
  recognition}, pages 248--255. Ieee, 2009.

\bibitem{dosovitskiy2020image}
Alexey Dosovitskiy, Lucas Beyer, Alexander Kolesnikov, Dirk Weissenborn,
  Xiaohua Zhai, Thomas Unterthiner, Mostafa Dehghani, Matthias Minderer, Georg
  Heigold, Sylvain Gelly, et~al.
\newblock An image is worth 16x16 words: Transformers for image recognition at
  scale.
\newblock {\em arXiv preprint arXiv:2010.11929}, 2020.

\bibitem{Everingham10}
M. Everingham, L. Van~Gool, C.~K.~I. Williams, J. Winn, and A. Zisserman.
\newblock The pascal visual object classes (voc) challenge.
\newblock {\em International Journal of Computer Vision}, 88(2):303--338, June
  2010.

\bibitem{ghiasi2021open}
Golnaz Ghiasi, Xiuye Gu, Yin Cui, and Tsung-Yi Lin.
\newblock Open-vocabulary image segmentation.
\newblock {\em arXiv preprint arXiv:2112.12143}, 2021.

\bibitem{gupta2020contrastive}
Tanmay Gupta, Arash Vahdat, Gal Chechik, Xiaodong Yang, Jan Kautz, and Derek
  Hoiem.
\newblock Contrastive learning for weakly supervised phrase grounding.
\newblock In {\em European Conference on Computer Vision}, pages 752--768.
  Springer, 2020.

\bibitem{hamilton2022unsupervised}
Mark Hamilton, Zhoutong Zhang, Bharath Hariharan, Noah Snavely, and William~T
  Freeman.
\newblock Unsupervised semantic segmentation by distilling feature
  correspondences.
\newblock {\em arXiv preprint arXiv:2203.08414}, 2022.

\bibitem{hendrycks2021many}
Dan Hendrycks, Steven Basart, Norman Mu, Saurav Kadavath, Frank Wang, Evan
  Dorundo, Rahul Desai, Tyler Zhu, Samyak Parajuli, Mike Guo, et~al.
\newblock The many faces of robustness: A critical analysis of
  out-of-distribution generalization.
\newblock In {\em Proceedings of the IEEE/CVF International Conference on
  Computer Vision}, pages 8340--8349, 2021.

\bibitem{hendrycks2021natural}
Dan Hendrycks, Kevin Zhao, Steven Basart, Jacob Steinhardt, and Dawn Song.
\newblock Natural adversarial examples.
\newblock In {\em Proceedings of the IEEE/CVF Conference on Computer Vision and
  Pattern Recognition}, pages 15262--15271, 2021.

\bibitem{hwang2019segsort}
Jyh-Jing Hwang, Stella~X Yu, Jianbo Shi, Maxwell~D Collins, Tien-Ju Yang, Xiao
  Zhang, and Liang-Chieh Chen.
\newblock Segsort: Segmentation by discriminative sorting of segments.
\newblock In {\em Proceedings of the IEEE/CVF International Conference on
  Computer Vision}, pages 7334--7344, 2019.

\bibitem{jia2021scaling}
Chao Jia, Yinfei Yang, Ye Xia, Yi-Ting Chen, Zarana Parekh, Hieu Pham, Quoc Le,
  Yun-Hsuan Sung, Zhen Li, and Tom Duerig.
\newblock Scaling up visual and vision-language representation learning with
  noisy text supervision.
\newblock In {\em International Conference on Machine Learning}, pages
  4904--4916. PMLR, 2021.

\bibitem{kamath2021mdetr}
Aishwarya Kamath, Mannat Singh, Yann LeCun, Gabriel Synnaeve, Ishan Misra, and
  Nicolas Carion.
\newblock Mdetr-modulated detection for end-to-end multi-modal understanding.
\newblock In {\em Proceedings of the IEEE/CVF International Conference on
  Computer Vision}, pages 1780--1790, 2021.

\bibitem{kim2021vilt}
Wonjae Kim, Bokyung Son, and Ildoo Kim.
\newblock Vilt: Vision-and-language transformer without convolution or region
  supervision.
\newblock In {\em International Conference on Machine Learning}, pages
  5583--5594. PMLR, 2021.

\bibitem{KrauseStarkDengFei-Fei_3DRR2013}
Jonathan Krause, Michael Stark, Jia Deng, and Li Fei-Fei.
\newblock 3d object representations for fine-grained categorization.
\newblock In {\em 4th International IEEE Workshop on 3D Representation and
  Recognition (3dRR-13)}, Sydney, Australia, 2013.

\bibitem{krizhevsky2009learning}
Alex Krizhevsky, Geoffrey Hinton, et~al.
\newblock Learning multiple layers of features from tiny images.
\newblock 2009.

\bibitem{li2022language}
Boyi Li, Kilian~Q Weinberger, Serge Belongie, Vladlen Koltun, and Ren{\'e}
  Ranftl.
\newblock Language-driven semantic segmentation.
\newblock {\em arXiv preprint arXiv:2201.03546}, 2022.

\bibitem{li2022mask}
Feng Li, Hao Zhang, Shilong Liu, Lei Zhang, Lionel~M Ni, Heung-Yeung Shum,
  et~al.
\newblock Mask dino: Towards a unified transformer-based framework for object
  detection and segmentation.
\newblock {\em arXiv preprint arXiv:2206.02777}, 2022.

\bibitem{caltech101}
Fei-Fei Li, Marco Andreeto, Marc'Aurelio Ranzato, and Pietro Perona.
\newblock Caltech 101, 2022.

\bibitem{li2021align}
Junnan Li, Ramprasaath Selvaraju, Akhilesh Gotmare, Shafiq Joty, Caiming Xiong,
  and Steven Chu~Hong Hoi.
\newblock Align before fuse: Vision and language representation learning with
  momentum distillation.
\newblock {\em Advances in neural information processing systems},
  34:9694--9705, 2021.

\bibitem{liu2022open}
Quande Liu, Youpeng Wen, Jianhua Han, Chunjing Xu, Hang Xu, and Xiaodan Liang.
\newblock Open-world semantic segmentation via contrasting and clustering
  vision-language embedding.
\newblock In {\em European Conference on Computer Vision}, pages 275--292.
  Springer, 2022.

\bibitem{liu2022swin}
Ze Liu, Han Hu, Yutong Lin, Zhuliang Yao, Zhenda Xie, Yixuan Wei, Jia Ning, Yue
  Cao, Zheng Zhang, Li Dong, et~al.
\newblock Swin transformer v2: Scaling up capacity and resolution.
\newblock In {\em Proceedings of the IEEE/CVF Conference on Computer Vision and
  Pattern Recognition}, pages 12009--12019, 2022.

\bibitem{Long_2015_CVPR}
Jonathan Long, Evan Shelhamer, and Trevor Darrell.
\newblock Fully convolutional networks for semantic segmentation.
\newblock In {\em Proceedings of the IEEE Conference on Computer Vision and
  Pattern Recognition (CVPR)}, June 2015.

\bibitem{melas2022deep}
Luke Melas-Kyriazi, Christian Rupprecht, Iro Laina, and Andrea Vedaldi.
\newblock Deep spectral methods: A surprisingly strong baseline for
  unsupervised semantic segmentation and localization.
\newblock In {\em Proceedings of the IEEE/CVF Conference on Computer Vision and
  Pattern Recognition}, pages 8364--8375, 2022.

\bibitem{mottaghi_cvpr14}
Roozbeh Mottaghi, Xianjie Chen, Xiaobai Liu, Nam-Gyu Cho, Seong-Whan Lee, Sanja
  Fidler, Raquel Urtasun, and Alan Yuille.
\newblock The role of context for object detection and semantic segmentation in
  the wild.
\newblock In {\em IEEE Conference on Computer Vision and Pattern Recognition
  (CVPR)}, 2014.

\bibitem{Nilsback08}
Maria-Elena Nilsback and Andrew Zisserman.
\newblock Automated flower classification over a large number of classes.
\newblock In {\em Indian Conference on Computer Vision, Graphics and Image
  Processing}, Dec 2008.

\bibitem{oord2018representation}
Aaron van~den Oord, Yazhe Li, and Oriol Vinyals.
\newblock Representation learning with contrastive predictive coding.
\newblock {\em arXiv preprint arXiv:1807.03748}, 2018.

\bibitem{parkhi12a}
Omkar~M. Parkhi, Andrea Vedaldi, Andrew Zisserman, and C.~V. Jawahar.
\newblock Cats and dogs.
\newblock In {\em IEEE Conference on Computer Vision and Pattern Recognition},
  2012.

\bibitem{PontTuset_eccv2020}
Jordi Pont-Tuset, Jasper Uijlings, Soravit Changpinyo, Radu Soricut, and
  Vittorio Ferrari.
\newblock Connecting vision and language with localized narratives.
\newblock In {\em ECCV}, 2020.

\bibitem{radford2021learning}
Alec Radford, Jong~Wook Kim, Chris Hallacy, Aditya Ramesh, Gabriel Goh,
  Sandhini Agarwal, Girish Sastry, Amanda Askell, Pamela Mishkin, Jack Clark,
  et~al.
\newblock Learning transferable visual models from natural language
  supervision.
\newblock In {\em International Conference on Machine Learning}, pages
  8748--8763. PMLR, 2021.

\bibitem{recht2019imagenet}
Benjamin Recht, Rebecca Roelofs, Ludwig Schmidt, and Vaishaal Shankar.
\newblock Do imagenet classifiers generalize to imagenet?
\newblock In {\em International Conference on Machine Learning}, pages
  5389--5400. PMLR, 2019.

\bibitem{ronneberger2015u}
Olaf Ronneberger, Philipp Fischer, and Thomas Brox.
\newblock U-net: Convolutional networks for biomedical image segmentation.
\newblock In {\em International Conference on Medical image computing and
  computer-assisted intervention}, pages 234--241. Springer, 2015.

\bibitem{sharma2018conceptual}
Piyush Sharma, Nan Ding, Sebastian Goodman, and Radu Soricut.
\newblock Conceptual captions: A cleaned, hypernymed, image alt-text dataset
  for automatic image captioning.
\newblock In {\em Proceedings of ACL}, 2018.

\bibitem{singh2022flava}
Amanpreet Singh, Ronghang Hu, Vedanuj Goswami, Guillaume Couairon, Wojciech
  Galuba, Marcus Rohrbach, and Douwe Kiela.
\newblock Flava: A foundational language and vision alignment model.
\newblock In {\em Proceedings of the IEEE/CVF Conference on Computer Vision and
  Pattern Recognition}, pages 15638--15650, 2022.

\bibitem{thomee2016yfcc100m}
Bart Thomee, David~A Shamma, Gerald Friedland, Benjamin Elizalde, Karl Ni,
  Douglas Poland, Damian Borth, and Li-Jia Li.
\newblock Yfcc100m: The new data in multimedia research.
\newblock {\em Communications of the ACM}, 59(2):64--73, 2016.

\bibitem{van2021unsupervised}
Wouter Van~Gansbeke, Simon Vandenhende, Stamatios Georgoulis, and Luc Van~Gool.
\newblock Unsupervised semantic segmentation by contrasting object mask
  proposals.
\newblock In {\em Proceedings of the IEEE/CVF International Conference on
  Computer Vision}, pages 10052--10062, 2021.

\bibitem{wang2019learning}
Haohan Wang, Songwei Ge, Zachary Lipton, and Eric~P Xing.
\newblock Learning robust global representations by penalizing local predictive
  power.
\newblock In {\em Advances in Neural Information Processing Systems}, pages
  10506--10518, 2019.

\bibitem{wang2020deep}
Jingdong Wang, Ke Sun, Tianheng Cheng, Borui Jiang, Chaorui Deng, Yang Zhao,
  Dong Liu, Yadong Mu, Mingkui Tan, Xinggang Wang, et~al.
\newblock Deep high-resolution representation learning for visual recognition.
\newblock {\em IEEE transactions on pattern analysis and machine intelligence},
  43(10):3349--3364, 2020.

\bibitem{wang2016comprehensive}
Kaiye Wang, Qiyue Yin, Wei Wang, Shu Wu, and Liang Wang.
\newblock A comprehensive survey on cross-modal retrieval.
\newblock {\em arXiv preprint arXiv:1607.06215}, 2016.

\bibitem{wang2022image}
Wenhui Wang, Hangbo Bao, Li Dong, Johan Bjorck, Zhiliang Peng, Qiang Liu, Kriti
  Aggarwal, Owais~Khan Mohammed, Saksham Singhal, Subhojit Som, et~al.
\newblock Image as a foreign language: Beit pretraining for all vision and
  vision-language tasks.
\newblock {\em arXiv preprint arXiv:2208.10442}, 2022.

\bibitem{wang2022tokencut}
Yangtao Wang, Xi Shen, Yuan Yuan, Yuming Du, Maomao Li, Shell~Xu Hu, James~L
  Crowley, and Dominique Vaufreydaz.
\newblock Tokencut: Segmenting objects in images and videos with
  self-supervised transformer and normalized cut.
\newblock {\em arXiv preprint arXiv:2209.00383}, 2022.

\bibitem{wei2022contrastive}
Yixuan Wei, Han Hu, Zhenda Xie, Zheng Zhang, Yue Cao, Jianmin Bao, Dong Chen,
  and Baining Guo.
\newblock Contrastive learning rivals masked image modeling in fine-tuning via
  feature distillation.
\newblock {\em arXiv preprint arXiv:2205.14141}, 2022.

\bibitem{wortsman2022model}
Mitchell Wortsman, Gabriel Ilharco, Samir~Ya Gadre, Rebecca Roelofs, Raphael
  Gontijo-Lopes, Ari~S Morcos, Hongseok Namkoong, Ali Farhadi, Yair Carmon,
  Simon Kornblith, et~al.
\newblock Model soups: averaging weights of multiple fine-tuned models improves
  accuracy without increasing inference time.
\newblock In {\em International Conference on Machine Learning}, pages
  23965--23998. PMLR, 2022.

\bibitem{xian2019semantic}
Yongqin Xian, Subhabrata Choudhury, Yang He, Bernt Schiele, and Zeynep Akata.
\newblock Semantic projection network for zero-and few-label semantic
  segmentation.
\newblock In {\em Proceedings of the IEEE/CVF Conference on Computer Vision and
  Pattern Recognition}, pages 8256--8265, 2019.

\bibitem{xu2022groupvit}
Jiarui Xu, Shalini De~Mello, Sifei Liu, Wonmin Byeon, Thomas Breuel, Jan Kautz,
  and Xiaolong Wang.
\newblock Groupvit: Semantic segmentation emerges from text supervision.
\newblock In {\em Proceedings of the IEEE/CVF Conference on Computer Vision and
  Pattern Recognition}, pages 18134--18144, 2022.

\bibitem{xu2021simple}
Mengde Xu, Zheng Zhang, Fangyun Wei, Yutong Lin, Yue Cao, Han Hu, and Xiang
  Bai.
\newblock A simple baseline for zero-shot semantic segmentation with
  pre-trained vision-language model.
\newblock {\em arXiv preprint arXiv:2112.14757}, 2021.

\bibitem{yu2022coca}
Jiahui Yu, Zirui Wang, Vijay Vasudevan, Legg Yeung, Mojtaba Seyedhosseini, and
  Yonghui Wu.
\newblock Coca: Contrastive captioners are image-text foundation models.
\newblock {\em arXiv preprint arXiv:2205.01917}, 2022.

\bibitem{yuan2021florence}
Lu Yuan, Dongdong Chen, Yi-Ling Chen, Noel Codella, Xiyang Dai, Jianfeng Gao,
  Houdong Hu, Xuedong Huang, Boxin Li, Chunyuan Li, et~al.
\newblock Florence: A new foundation model for computer vision.
\newblock {\em arXiv preprint arXiv:2111.11432}, 2021.

\bibitem{zhang2022dino}
Hao Zhang, Feng Li, Shilong Liu, Lei Zhang, Hang Su, Jun Zhu, Lionel~M Ni, and
  Heung-Yeung Shum.
\newblock Dino: Detr with improved denoising anchor boxes for end-to-end object
  detection.
\newblock {\em arXiv preprint arXiv:2203.03605}, 2022.

\bibitem{zhang2022glipv2}
Haotian Zhang, Pengchuan Zhang, Xiaowei Hu, Yen-Chun Chen, Liunian~Harold Li,
  Xiyang Dai, Lijuan Wang, Lu Yuan, Jenq-Neng Hwang, and Jianfeng Gao.
\newblock Glipv2: Unifying localization and vision-language understanding.
\newblock {\em arXiv preprint arXiv:2206.05836}, 2022.

\bibitem{zhao2017pyramid}
Hengshuang Zhao, Jianping Shi, Xiaojuan Qi, Xiaogang Wang, and Jiaya Jia.
\newblock Pyramid scene parsing network.
\newblock In {\em Proceedings of the IEEE conference on computer vision and
  pattern recognition}, pages 2881--2890, 2017.

\bibitem{zhou2017scene}
Bolei Zhou, Hang Zhao, Xavier Puig, Sanja Fidler, Adela Barriuso, and Antonio
  Torralba.
\newblock Scene parsing through ade20k dataset.
\newblock In {\em Proceedings of the IEEE conference on computer vision and
  pattern recognition}, pages 633--641, 2017.

\end{thebibliography}
}

\clearpage
\appendix

\section{Additional Implementation Details}
\label{app:additional_implementation_details}

In this section, we describe the implementation details for the proposed PACL method. Particularly, in \Cref{app:vision_embedder_architecture}, we describe the architecture of the vision embedder used for training PACL, in \Cref{app:training_details}, we describe specifics of training including hyperparameters and prompt engineering details. Finally, in \Cref{app:training_eval_datasets}, we describe details of image-text datasets used for training as well as segmentation and image classification datasets used for evaluation.

\subsection {Vision Embedder Architecture}
\label{app:vision_embedder_architecture}

\begin{figure}[!t]
    \centering
    \includegraphics[width=\linewidth]{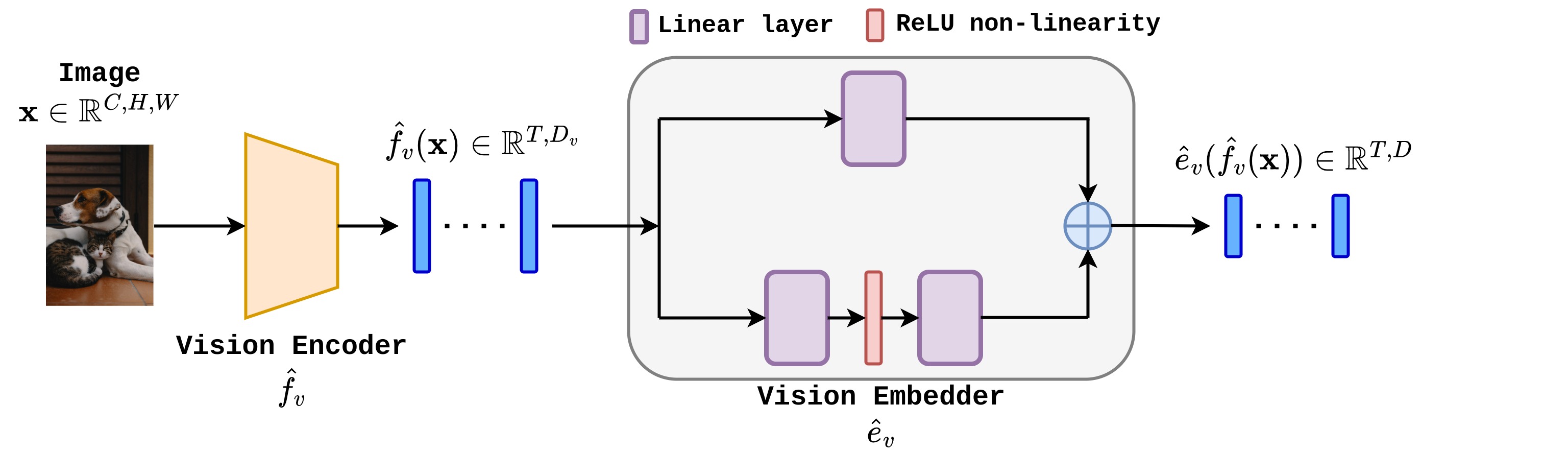}
    \caption{\textbf{Vision embedder $\hat{e}_v$ architecture for PACL.} The image encoder $\hat{f}_v$ produces token/patch-wise representations $\hat{f}_v(\mathbf{x}) \in \mathbb{R}^{T, D_v}$ of an input image $\mathbf{x}$. The vision embedder $\hat{e}_v$ converts the patch-wise representations to the multi-modal shared dimensional space, $\hat{e}_v(\hat{f}_v(\mathbf{x})) \in \mathbb{R}^{T, D}$ where $T$ is the number of tokens or patches.}
    \vspace{-4mm}
    \label{fig:pacl_vision_embedder}
\end{figure}

In \Cref{sec:zeroshot_semantic_segmentation}, we have discussed that the proposed PACL approach is flexible in the sense that PACL can be applied using pre-trained frozen encoders. Particularly, since CLIP's pre-trained vision encoders have desirable properties (see Semantic Coherence in \Cref{sec:semantic_coherence}), we use these pre-trained encoders from CLIP to train PACL to transfer to the task of zero-shot semantic segmentation. This simplifies the training to just a small vision embedder on top of the vision encoder. In this section and in \Cref{fig:pacl_vision_embedder}, we present the architecture of the Vision embedder. In particular, we use a single residual block with two linear layers in the main branch and a single linear layer in the residual branch. There is a ReLU non-linearity between the two linear layers in the main branch. The resulting model requires training a mere $1.1$M parameters whereas the architecture has a total of $150$M parameters for CLIP ViT-B/16. This helps us in scaling up and training on a larger batch size for our experiments as there is no gradient propagation through the frozen image and text encoders.

\subsection {Training details for Vision Embedder}
\label{app:training_details}

In \Cref{app:architecture_hypers}, we describe the architecture of pre-trained encoders as well as the hyperparameters used for training the PACL models. In \Cref{app:prompt_engg}, we provide some details on CLIP's prompt engineering used to derive best results from the text encoder of a pre-trained CLIP model.

\subsubsection{Architecture and Hyperparameters}
\label{app:architecture_hypers}

As mentioned above, we only train PACL using a Vision embedder on top of a pre-trained CLIP vision encoder. This allows us the flexibility not only of using multiple pre-trained vision encoders but also combinations of different vision and text encoders. In \Cref{sec:zeroshot_semantic_segmentation}, we show an ablation with combinations of different pre-trained vision and text encoders. In particular, we use: a) CLIP ViT-B/16 vision and text encoders, b) CLIP ViT-L/14 vision and text encoders and b) DINO ViT-B/16 vision encoder with CLIP ViT-B/16 text encoder. For each of these combinations, we train a vision embedder as discussed in \Cref{app:vision_embedder_architecture} and report zero-shot semantic segmentation results in \Cref{table:zeroshot_seg_ablations} of the main paper.

All our models are trained on a single node with 4 NVIDIA A100 GPUs with a GPU memory of 40GB in each GPU. We use AdamW as the optimizer with beta values $0.9$ and $0.98$, an eps value of $1e-6$ and a weight decay of $0.2$. We use an initial learning rate of $5e-4$ and reduce the learning rate using a Cosine Annealing schedule where the maximum number of iterations is set as the total number of iterations during training (i.e., number of epochs $\times$ number of iterations per epoch). We use a batch size of 4096 (1024 per GPU) and train the model for a total of $10$ epochs on image-text data. \emph{We do not use any segmentation annotations or class-agnostic segmentation masks during training}. We provide details on these image-text datasets in \Cref{app:training_eval_datasets}. When the training dataset is a combination of GCC-3M, GCC-12M and YFCC-15M, the model takes 10 days to train on 4 NVIDIA A100 GPUs.

\subsubsection{Prompt Engineering}
\label{app:prompt_engg}

Since we use CLIP's \cite{radford2021learning} pre-trained text encoders, we follow the prompt engineering guidelines following CLIP's OpenAI repository during inference time. In particular, during inference, we compute the average embedding from the text encoder using a set of 7 prompts: \texttt{itap of a ().}, \texttt{a bad photo of the ().}, \texttt{a origami ().}, \texttt{a photo of the large ().}, \texttt{a () in a video game.}, \texttt{art of the ()}, \texttt{a photo of the small ()}, where we put the name of the class within the parenthesis \texttt{()}. We use the mean of the embeddings from the the prompts for each class in order to compute cosine similarity with the patch representations from the vision encoder. This is similar to the way CLIP performs zero-shot image classification, however CLIP uses only the CLS token from the vision encoder to compute cosine similarity.

\subsection {Training and Evaluation Datasets}
\label{app:training_eval_datasets}

\textbf{Image-text datasets for training: } We use primarily 3 different image-text datasets for training all our models. \textbf{Firstly,} we use \emph{Google Conceptual Captions (GCC) 3M}, which contains approximately 3 million images, each annotated with a caption. The images are scraped from the web and the corresponding captions are obtained from the Al-text HTML data associated with each image from the web. \textbf{Secondly,} we use \emph{Google Conceptual Captions (GCC) 12M}, which is similar to GCC-3M but containing a much larger corpus of image-text pairs with approximately 12 million samples. The primary purpose of GCC-12M is for pre-training whereas GCC-3M is a relatively less noisy dataset meant for fine-tuning pre-trained models. \textbf{Thirdly,} we use \emph{YFCC-15M}, a subset of 15 million samples from the popular YFCC-100M \cite{thomee2016yfcc100m} dataset, which is one of the largest publicly available datasets containing image-text information obtained from Flickr. The subset of approximately 15 million images is defined by CLIP \cite{radford2021learning} by filtering images from YFCC-100M with natural language titles and/or descriptions in English.   

\textbf{Semantic segmentation datasets for zero-shot segmentation: } We use the following semantic segmentation datasets for zero-shot evaluation on the task of semantic segmentation: \textbf{a)} \emph{Pascal VOC} \cite{Everingham10}: it has 20 foreground classes and 1 background class with 1449 validation images. We measure performance only on foreground classes and mask predictions with entropy above $1.5$ as background, \textbf{b)} \emph{Pascal Context} \cite{mottaghi_cvpr14}: it has 59 classes with 5k validation images of indoor and outdoor scenes, \textbf{c)} \emph{COCO Stuff} \cite{caesar2018coco}: it has 172 classes categorised into either ``thing" classes or ``stuff" classes and has 5k validation images, \textbf{d)} \emph{ADE20K} \cite{zhou2017scene}: the version we evaluate on is widely used and has 150 classes with 2k validation images. For all datasets, we report the mean intersection over union (mIoU), the most popular evaluation metric for semantic segmentation.

\textbf{Image classification datasets for zero-shot classification: } We evaluate PACL on a suite of 12 image classification datasets which include ImageNet \cite{deng2009imagenet}, 4 well-known distribution shifts on ImageNet as well as 7 other popular image classification datasets. \emph{ImageNet} is a very popular image classification dataset with 1000 classes relating to concepts contained in the WordNet hierarchy. We use $50000$ validation samples in ImageNet for evaluation. The 4 datasets considered to be popular distribution shifts on ImageNet are: \textbf{a)} \emph{ImageNet-A} \cite{hendrycks2021natural} which contains natural real-world images from 200 classes in ImageNet but which are mostly mis-classified by well-known ResNet classifiers, \textbf{b)} \emph{ImageNet-R} \cite{hendrycks2021many} which contains cartoons, graphics and other art renditions of images from 200 classes in ImageNet, \textbf{c)} \emph{ImageNet-Sketch} \cite{wang2019learning} which contains $50000$ validation images, $50$ from each of the $1000$ ImageNet classes constructed by making the Google search, "sketch of ()" where () is the ImageNet class concerned and \textbf{d)} \emph{ImageNet-V2} \cite{recht2019imagenet} which has $10000$ validation images obtained by following the same collection procedure as ImageNet original images, in order to make the distribution of ImageNet-V2 as similar as possible to ImageNet. The other 7 image classification datasets include: \textbf{a)} \emph{CIFAR-10} \cite{krizhevsky2009learning} having $10000$ test images from 10 classes including different types of automobiles and animals, \textbf{b)} \emph{CIFAR-100} \cite{krizhevsky2009learning} having $10000$ test images from 100 classes instead of 10 obtained in a similar fashion as CIFAR-10, \textbf{c)} \emph{Stanford Cars} \cite{KrauseStarkDengFei-Fei_3DRR2013} with $8041$ test images containing cars of different makes and models, \textbf{d)} \emph{Caltech-101} \cite{caltech101} having 101 categories of images with 40-800 images per class, \textbf{e)} \emph{Food-101} \cite{bossard14} containing 101 classes of food items organized by the type of food, with approximately $25000$ test images, \textbf{f)} \emph{Oxford-IIIT Pets} \cite{parkhi12a}, a dataset with $37$ categories of pets with approximately $200$ images per class, and \textbf{g)} \emph{Flower} dataset \cite{Nilsback08} having 102 different categories of flowers with between 40 and 258 images for each class.

\section{Additional Results}
\label{app:additional_results}

\subsection{Semantic Coherence in CLIP}
\label{app:semantic_coherence}

\begin{figure}[!t]
    \centering
    \begin{subfigure}{0.18\linewidth}
        \centering
        \includegraphics[width=\linewidth]{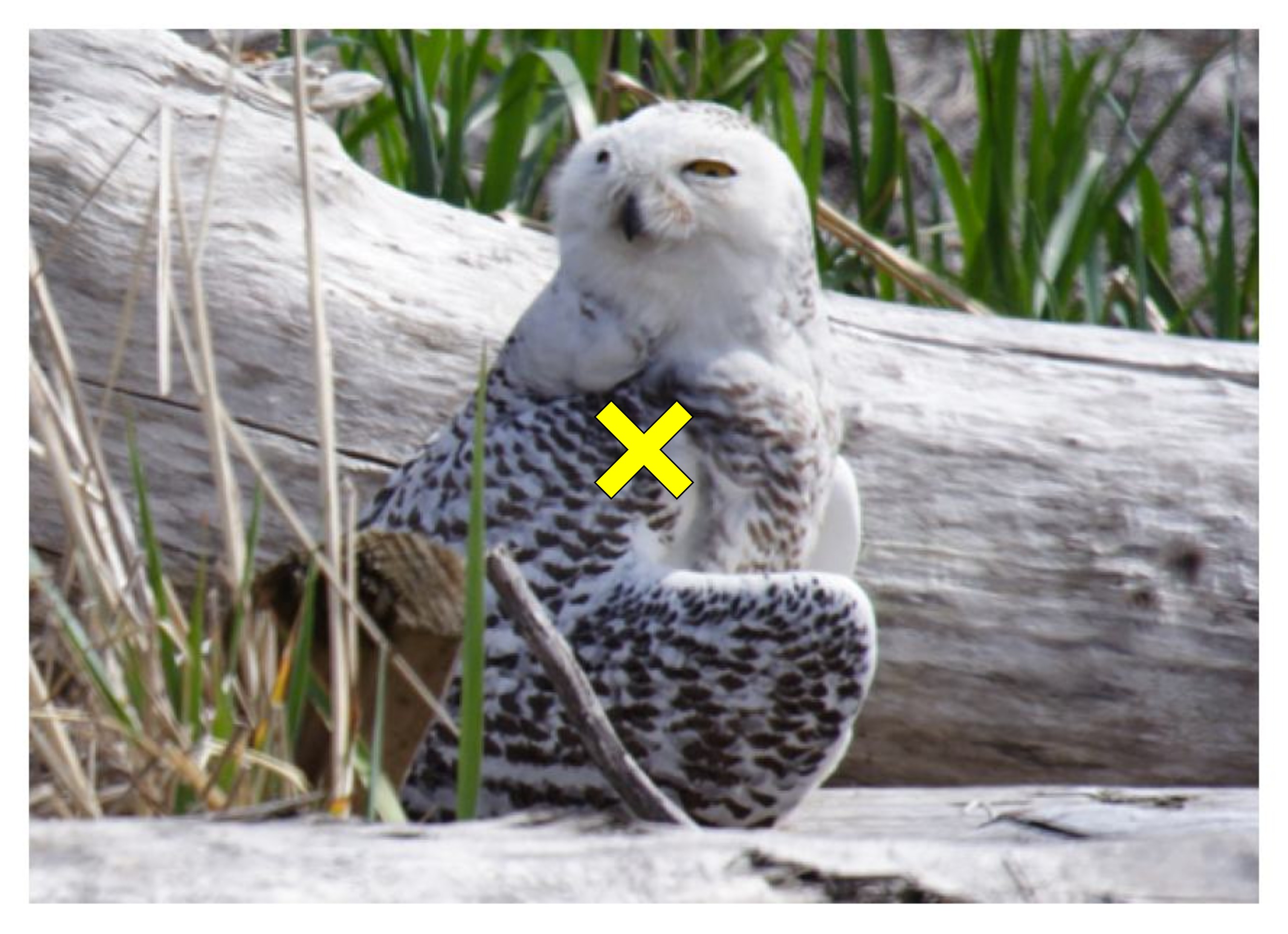}
    \end{subfigure}
    \begin{subfigure}{0.18\linewidth}
        \centering
        \includegraphics[width=\linewidth]{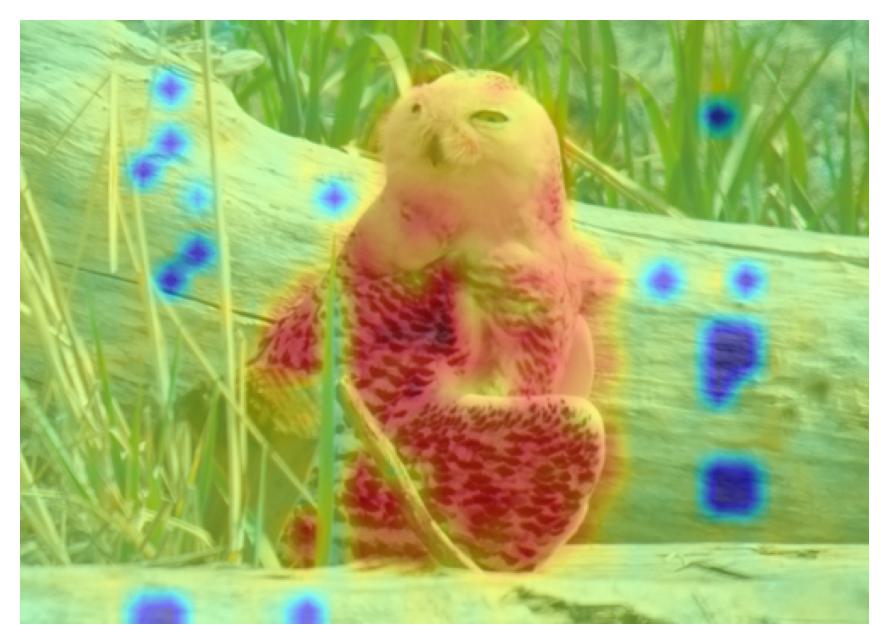}
    \end{subfigure}
    \begin{subfigure}{0.17\linewidth}
        \centering
        \includegraphics[width=\linewidth]{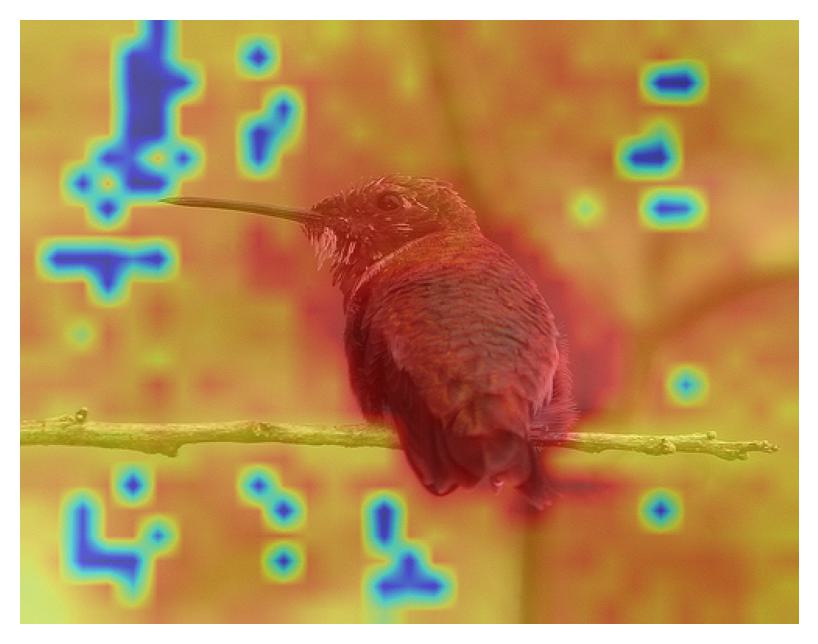}
    \end{subfigure}
    \begin{subfigure}{0.18\linewidth}
        \centering
        \includegraphics[width=\linewidth]{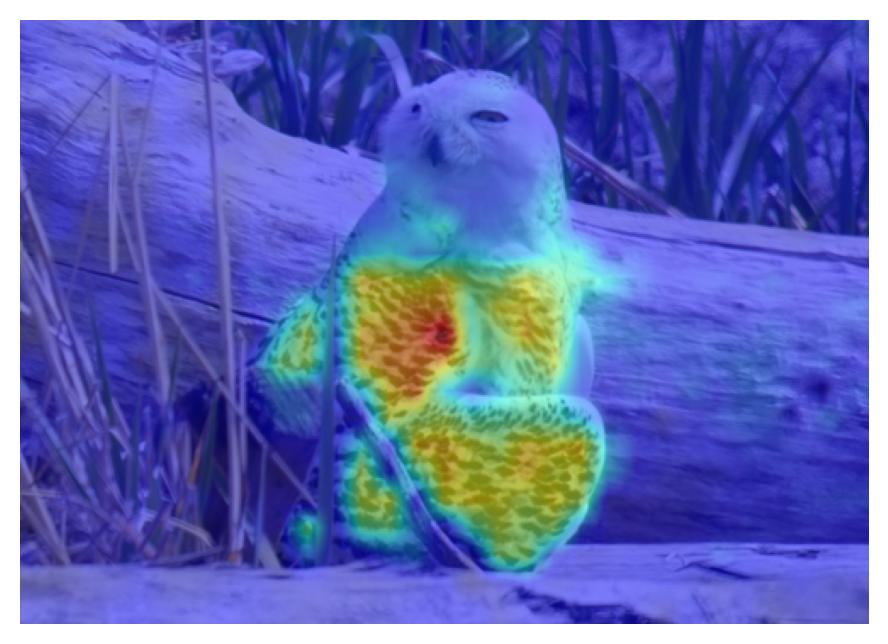}
    \end{subfigure}
    \begin{subfigure}{0.17\linewidth}
        \centering
        \includegraphics[width=\linewidth]{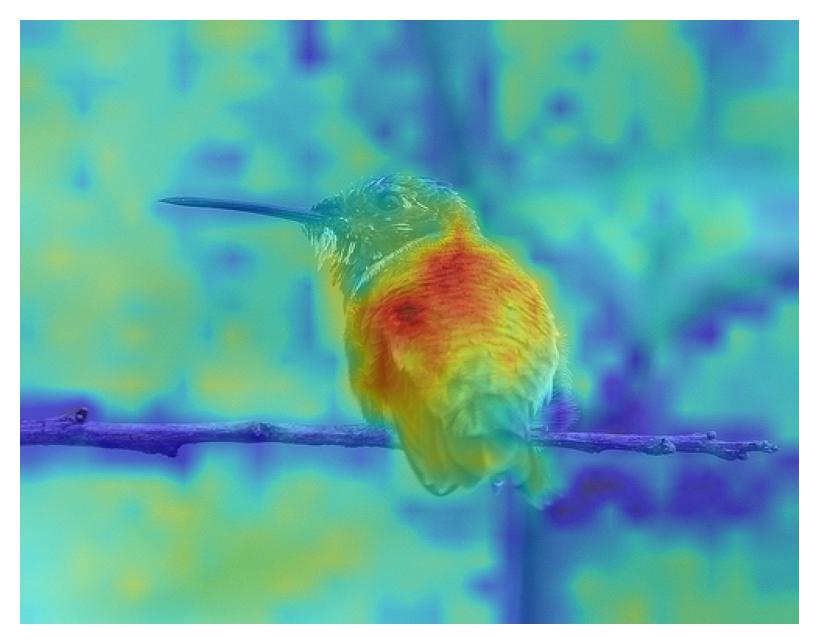}
    \end{subfigure}

    \begin{subfigure}{0.18\linewidth}
        \centering
        \includegraphics[width=\linewidth]{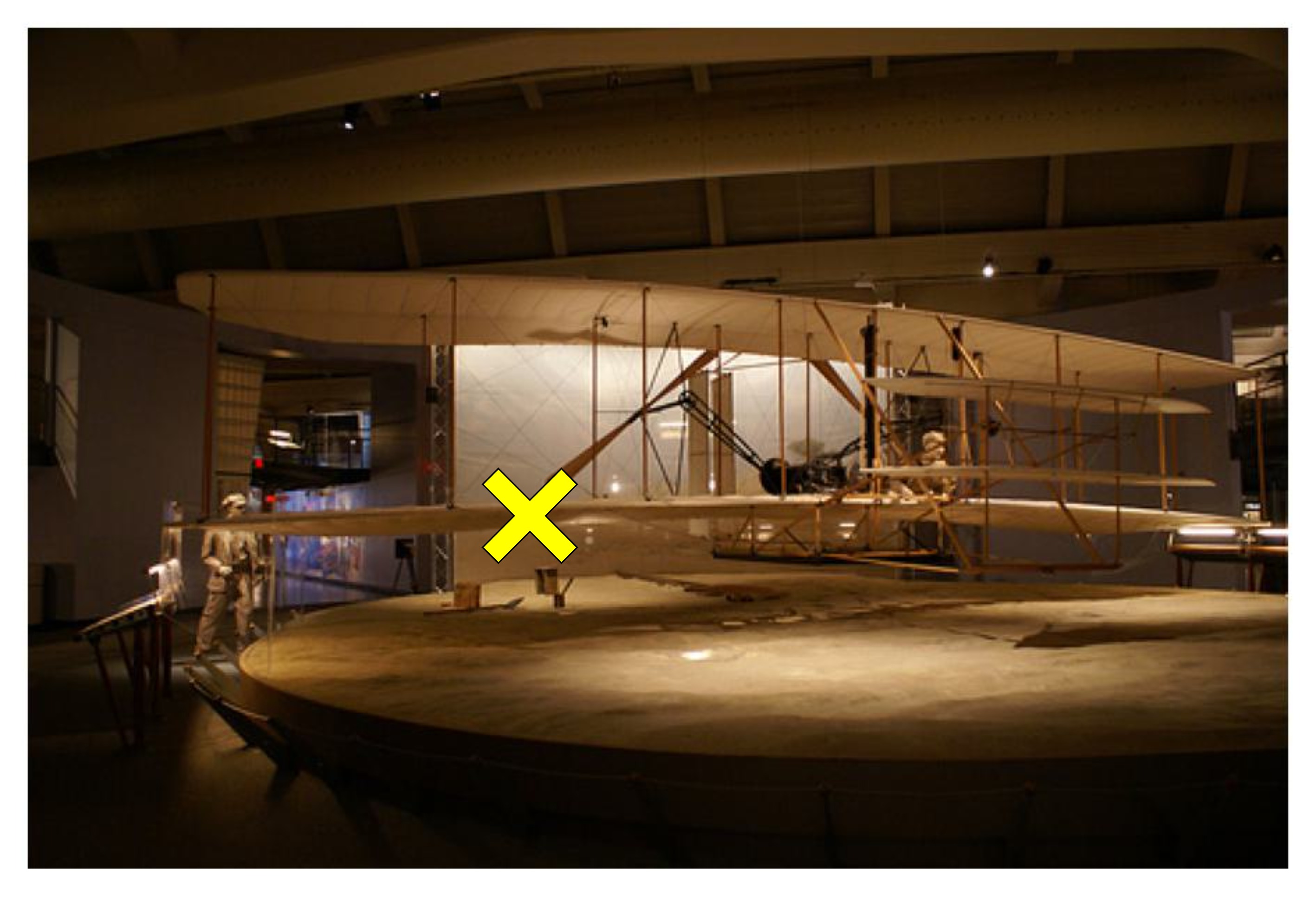}
    \end{subfigure}
    \begin{subfigure}{0.18\linewidth}
        \centering
        \includegraphics[width=\linewidth]{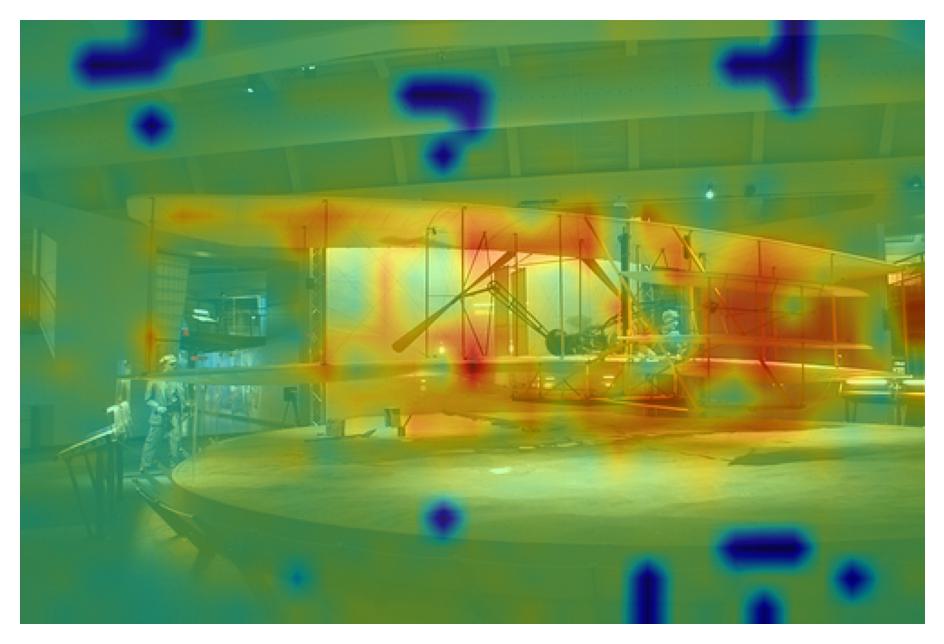}
    \end{subfigure}
    \begin{subfigure}{0.17\linewidth}
        \centering
        \includegraphics[width=\linewidth]{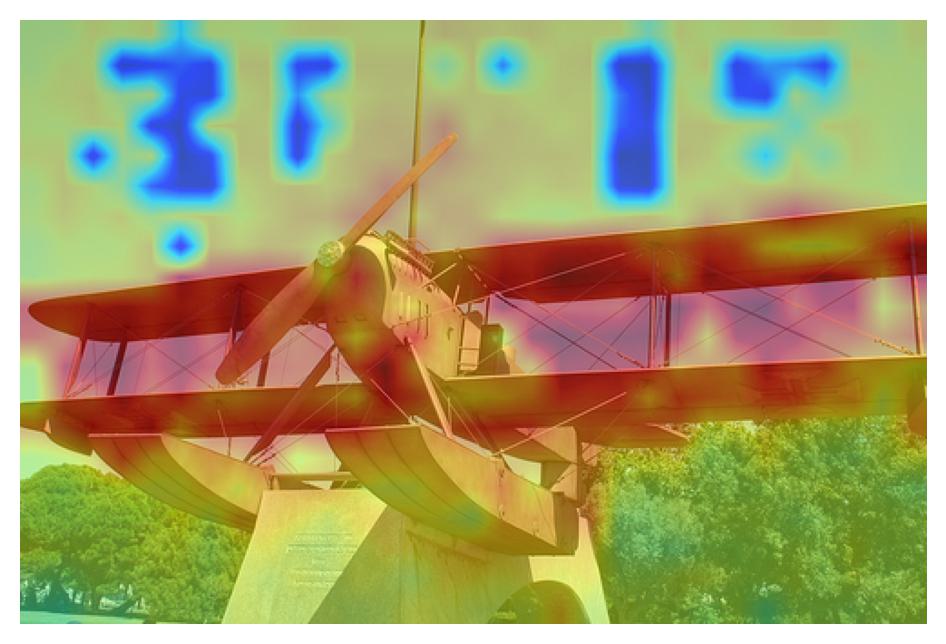}
    \end{subfigure}
    \begin{subfigure}{0.18\linewidth}
        \centering
        \includegraphics[width=\linewidth]{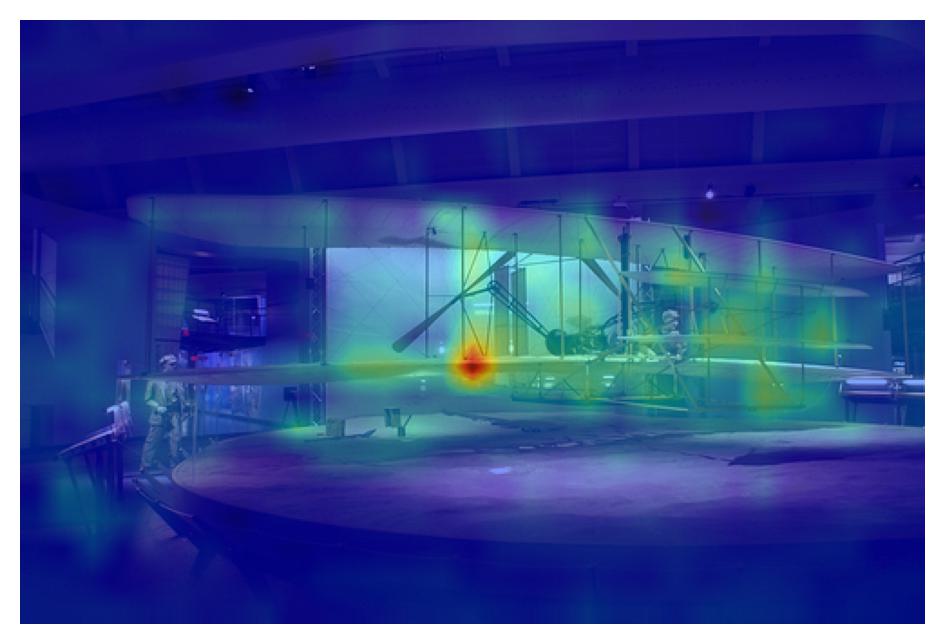}
    \end{subfigure}
    \begin{subfigure}{0.17\linewidth}
        \centering
        \includegraphics[width=\linewidth]{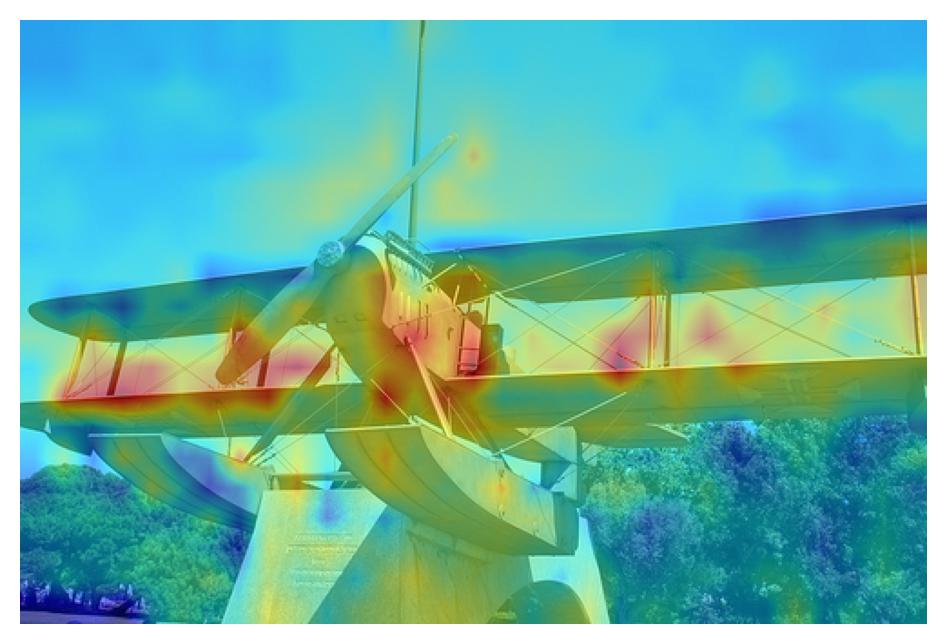}
    \end{subfigure}

    \begin{subfigure}{0.18\linewidth}
        \centering
        \includegraphics[width=\linewidth]{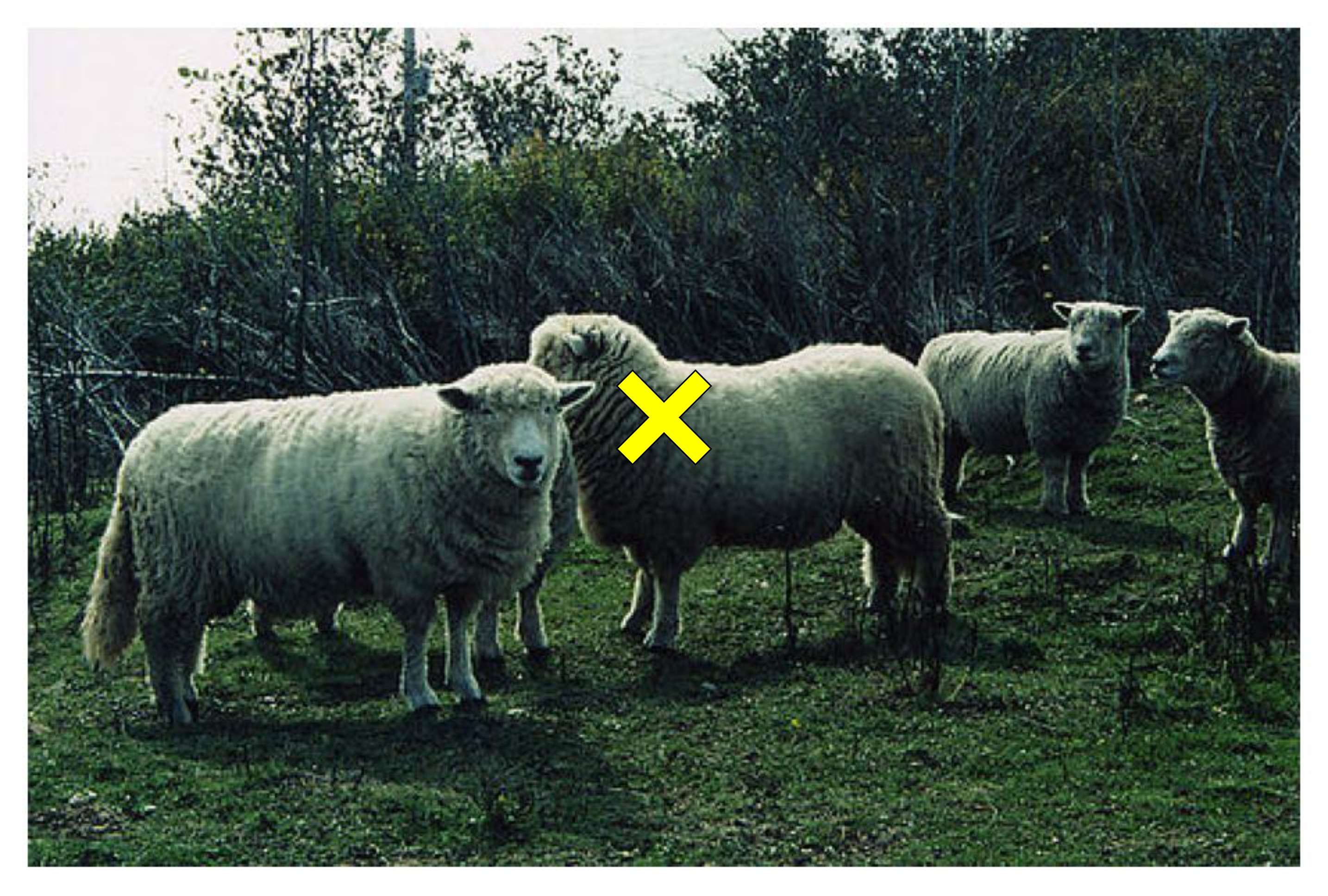}
        \caption{}
        \vspace{-2mm}
    \end{subfigure}
    \begin{subfigure}{0.18\linewidth}
        \centering
        \includegraphics[width=\linewidth]{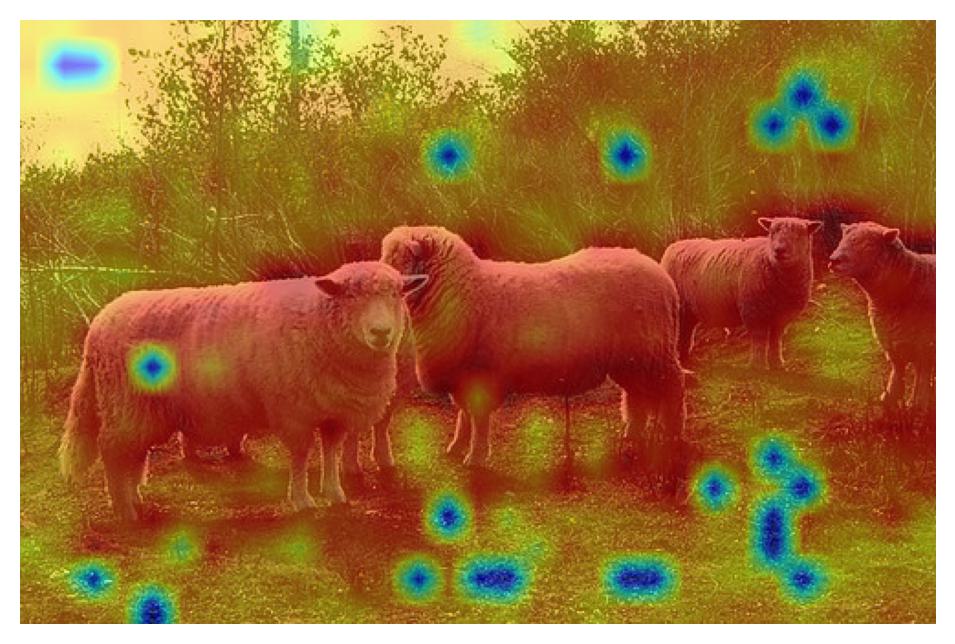}
        \caption{}
        \vspace{-2mm}
    \end{subfigure}
    \begin{subfigure}{0.17\linewidth}
        \centering
        \includegraphics[width=\linewidth]{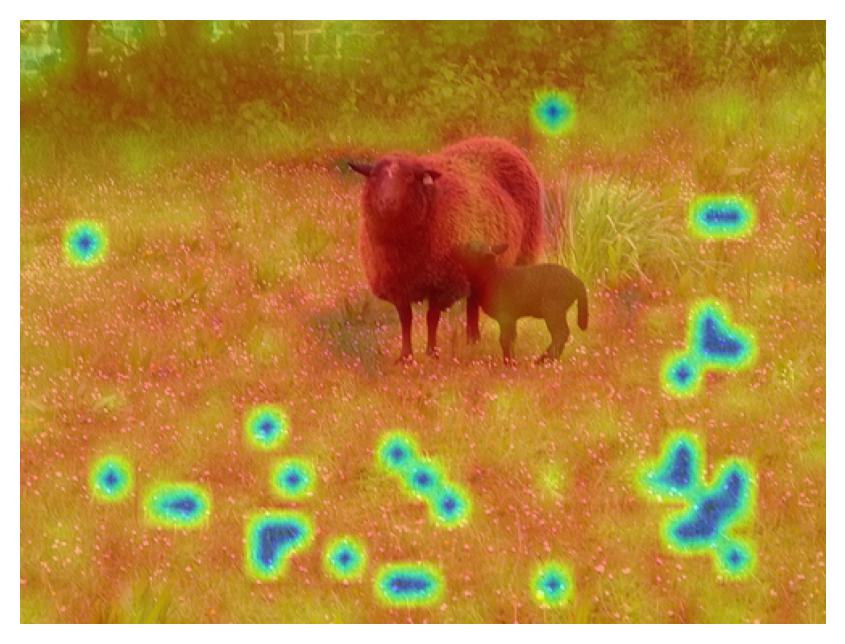}
        \caption{}
        \vspace{-2mm}
    \end{subfigure}
    \begin{subfigure}{0.18\linewidth}
        \centering
        \includegraphics[width=\linewidth]{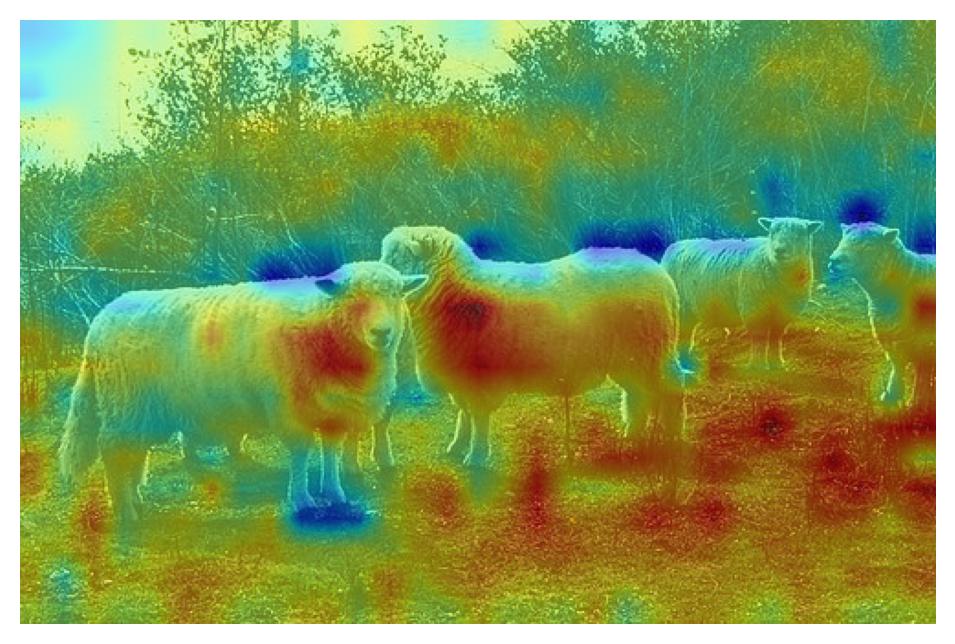}
        \caption{}
        \vspace{-2mm}
    \end{subfigure}
    \begin{subfigure}{0.17\linewidth}
        \centering
        \includegraphics[width=\linewidth]{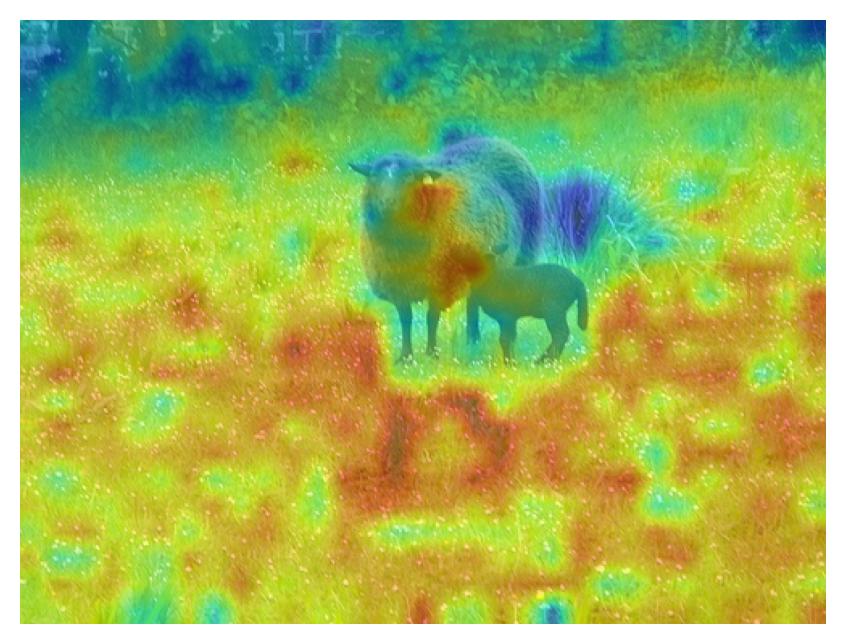}
        \caption{}
        \vspace{-2mm}
    \end{subfigure}
    \caption{\textbf{Additional qualitative results on semantic coherence between CLIP and DINO ViT-B/16.} \textbf{a)}: we show the original image of a class (bird in top row, aeroplane in middle row and sheep in bottom row) with the patch marker (yellow X near the centre). \textbf{b, c)}: we show CLIP vision encoder cosine similarity across all patches for the same and a different image of the same class. \textbf{d, e)}: we show the same for DINO.}
    \vspace{-4mm}
    \label{fig:app_coherence_qualitative}
\end{figure}

Semantic coherence is a property of ViT based vision encoders where semantically similar regions of the image have similar patch/token level representations in the feature space of the vision encoder. In \Cref{sec:semantic_coherence} and \Cref{fig:coherence_qualitative}, we have shown both quantitative and qualitative results comparing the semantic coherence of a CLIP and a DINO ViT-B/16 vision encoders. Particularly, we had seen that CLIP's ViT-B/16 vision encoder performs better than DINO. In \Cref{fig:app_coherence_qualitative}, we present additional qualitative examples to further corroborate our observations in \Cref{sec:semantic_coherence}. We show qualitative examples from the Bird, Plane and Sheep classes in Pascal VOC and plot the patch level similarity between a selected patch from the original image (marked using a yellow cross) and all patches from the same image as well as a different image. The similarity is shown using a heatmap where yellow and red shades indicate high similarity and blue shades indicate low similarity. Our observations are similar to the ones in \Cref{sec:semantic_coherence}, and we see that CLIP performs competitively or better than DINO. While DINO seems to cover semantically meaningful regions in the images, it doesn't cover the entirety of the relevant object. CLIP seems to be doing a better job at covering all the patches for the object as highly similar to the marked patch, thereby indicating better semantic coherence.

\begin{table*}[!t]
    \centering
    \scriptsize
    \resizebox{0.95\linewidth}{!}
    {
    \begin{tabular}{c|c|cccccccc|cccc}
    \toprule
    & & \multicolumn{12}{c}{\textbf{Datasets}} \\
    \textbf{Model} & \textbf{Vision Encoder} & \multicolumn{8}{c}{\textbf{\textit{Image Classification}}} & \multicolumn{4}{c}{\textbf{\textit{ImageNet Shifts}}} \\
    & & \textit{ImageNet} \cite{deng2009imagenet} & \textit{C10} \cite{krizhevsky2009learning} & \textit{C100} \cite{krizhevsky2009learning} & \textit{Cars} \cite{KrauseStarkDengFei-Fei_3DRR2013} & \textit{Caltech101} \cite{caltech101} & \textit{Food101} \cite{bossard14} & \textit{Pets} \cite{parkhi12a} & \textit{Flowers102} \cite{Nilsback08} & \textit{ImageNet-A} \cite{hendrycks2021natural} & \textit{ImageNet-R} \cite{hendrycks2021many} & \textit{ImageNet-Sketch} \cite{wang2019learning} & \textit{ImageNet-V2} \cite{recht2019imagenet} \\
    \midrule
    \multirow{2}{*}{CLIP} & ViT-B/16 & $68.73$ & $91.18$ & $67.88$ & $63.50$ & $85.69$ & $87.52$ & $88.44$ & $61.12$ & $38.88$ & $76.83$ & $48.36$ & $62.21$ \\
                          & ViT-L/14 & $75.96$ & $95.85$ & $76.94$ & $76.9$ & $86.38$ & $92.69$ & $92.91$ & $69.13$ & $55.44$ & $87.32$ & $59.71$ & $70.26$ \\
    \midrule
    \multirow{2}{*}{CLIP + PACL (Ours)} & ViT-B/16 & $73.61$ & $92.3$ & $69.11$ & $60.7$ & $84.8$ & $89.12$ & $90.1$ & $62.3$ & $42.10$ & $78.1$ & $50.14$ & $65.4$ \\
                                 & ViT-L/14 & $78.2$ & $95.13$ & $74.43$ & $74.2$ & $86.25$ & $93.2$ & $93.05$ & $69.7$ & $59.13$ & $85.6$ & $63.23$ & $72.88$ \\
    \bottomrule
    \end{tabular}
    }
    \vspace{-3mm}
    \caption{\textbf{Zero-shot Image Classification on 12 different datasets.} We compare PACL's performance with vanilla CLIP for both ViT-B/16 and ViT-L/14 encoders. The first 8 datasets are standard image classification datasets: ImageNet, CIFAR-10, CIFAR-100, Stanford Cars, Caltech101, Food101, OxfordIIITPets, and Flowers102. The remaining 4 datasets are standard distribution shifts on ImageNet: ImageNet-A, ImageNet-R, ImageNet-Sketch and ImageNet-V2. PACL + CLIP broadly outperforms vanilla CLIP on most of the classification datasets.}
    \label{table:zeroshot_image_classification}
\end{table*}

\subsection {Qualitative Segmentation Results}
\label{app:qualitative_segmentation_results}

\begin{figure}[!t]
    \centering
    \begin{subfigure}{0.135\linewidth}
        \centering
        \includegraphics[width=\linewidth]{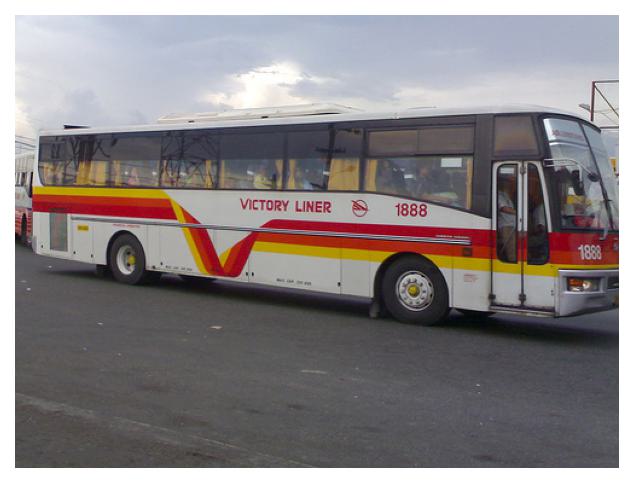}
    \end{subfigure}
    \begin{subfigure}{0.135\linewidth}
        \centering
        \includegraphics[width=\linewidth]{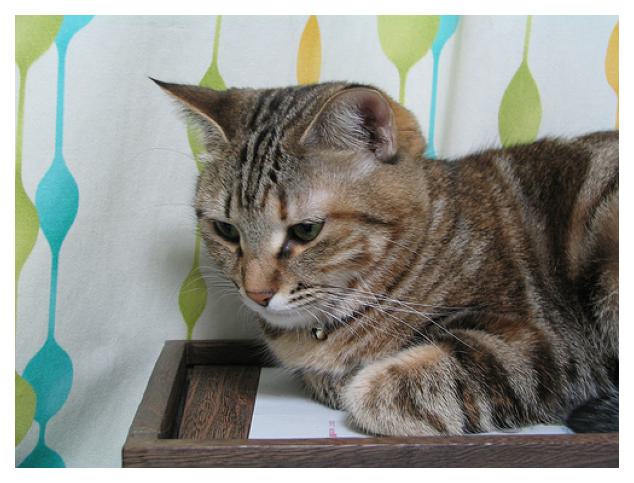}
    \end{subfigure}
    \begin{subfigure}{0.135\linewidth}
        \centering
        \includegraphics[width=\linewidth]{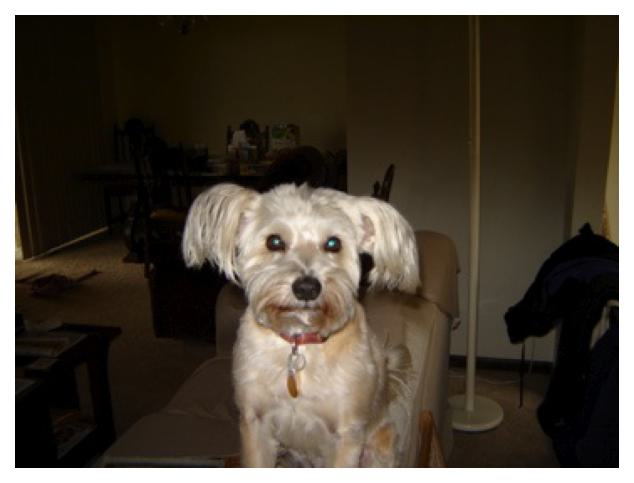}
    \end{subfigure}
    \begin{subfigure}{0.135\linewidth}
        \centering
        \includegraphics[width=\linewidth]{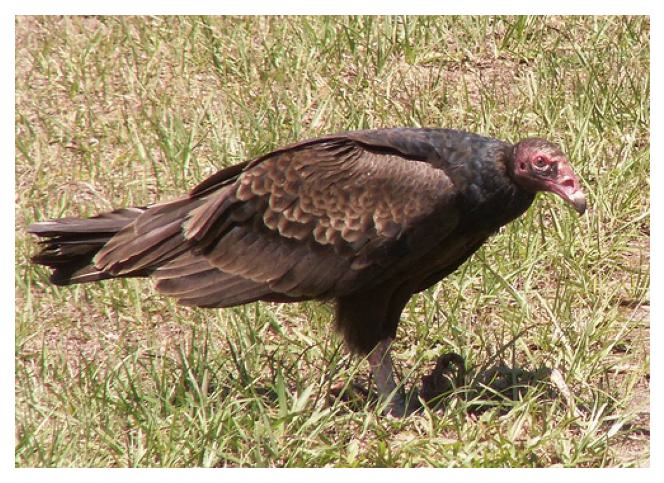}
    \end{subfigure}
    \begin{subfigure}{0.085\linewidth}
        \centering
        \includegraphics[width=\linewidth]{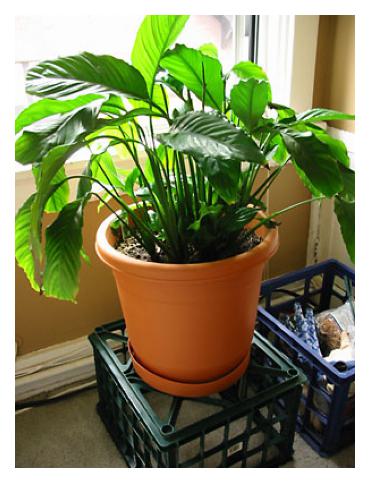}
    \end{subfigure}
    \begin{subfigure}{0.085\linewidth}
        \centering
        \includegraphics[width=\linewidth]{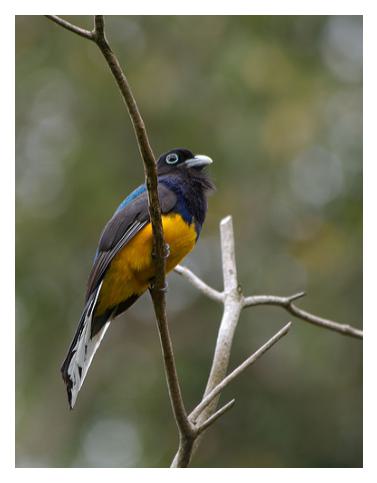}
    \end{subfigure}
    \begin{subfigure}{0.075\linewidth}
        \centering
        \includegraphics[width=\linewidth]{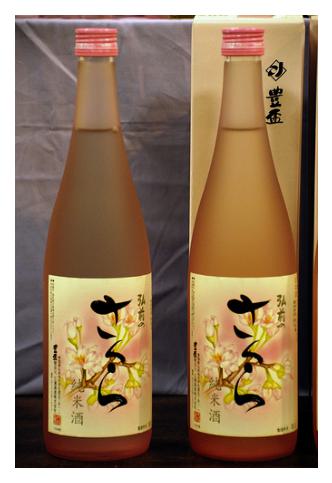}
    \end{subfigure}
    \begin{subfigure}{0.135\linewidth}
        \centering
        \includegraphics[width=\linewidth]{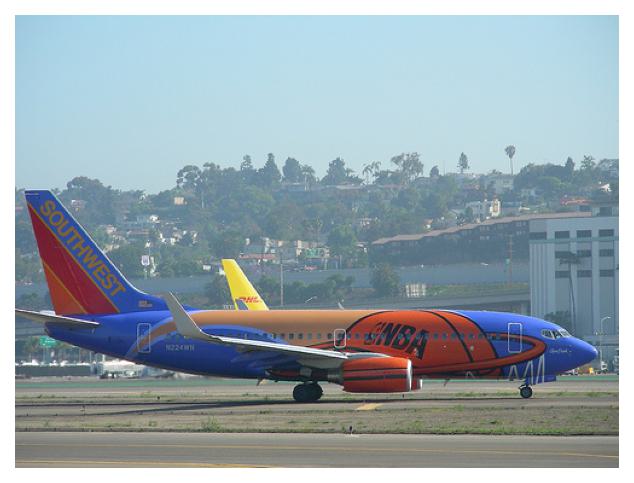}
    \end{subfigure}

    \begin{subfigure}{0.135\linewidth}
        \centering
        \includegraphics[width=\linewidth]{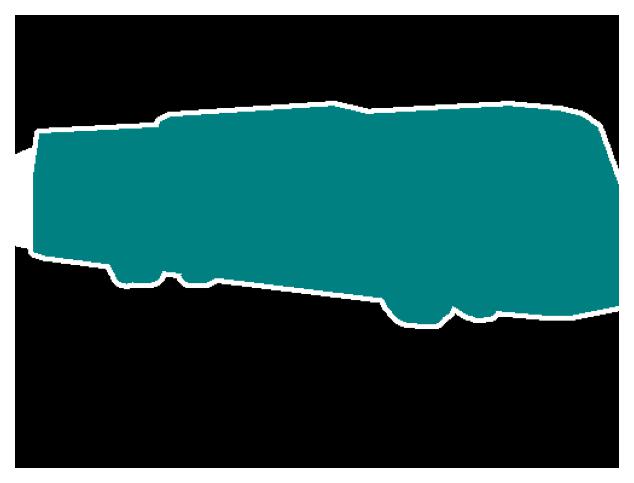}
    \end{subfigure}
    \begin{subfigure}{0.135\linewidth}
        \centering
        \includegraphics[width=\linewidth]{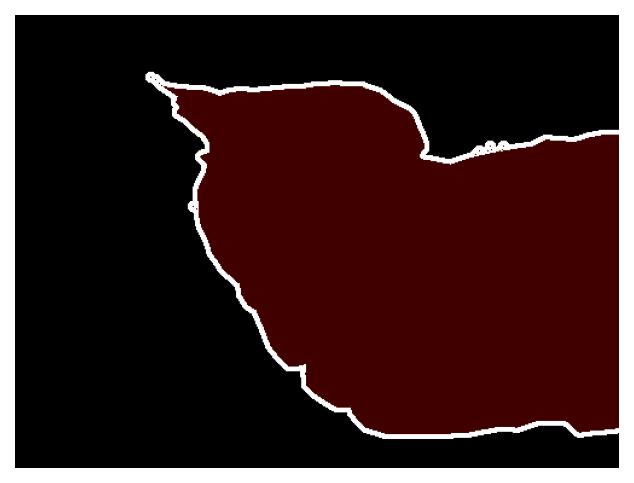}
    \end{subfigure}
    \begin{subfigure}{0.135\linewidth}
        \centering
        \includegraphics[width=\linewidth]{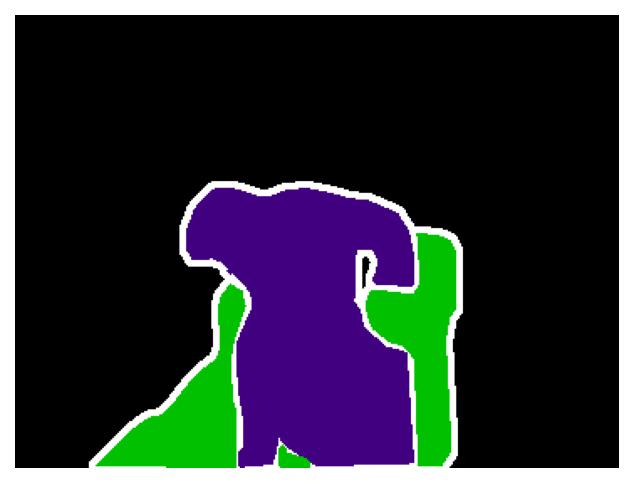}
    \end{subfigure}
    \begin{subfigure}{0.135\linewidth}
        \centering
        \includegraphics[width=\linewidth]{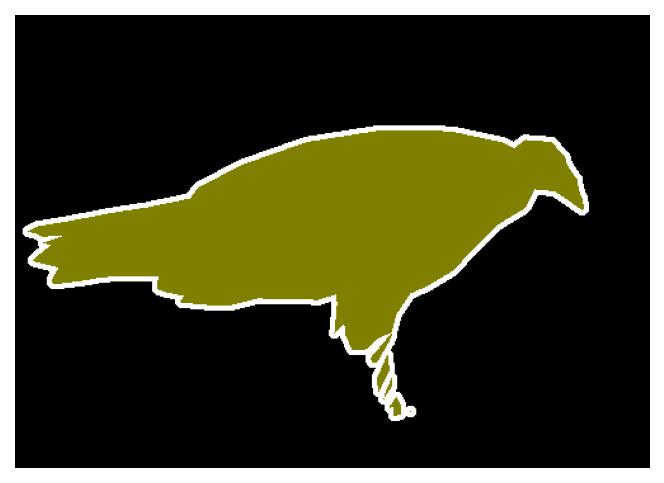}
    \end{subfigure}
    \begin{subfigure}{0.085\linewidth}
        \centering
        \includegraphics[width=\linewidth]{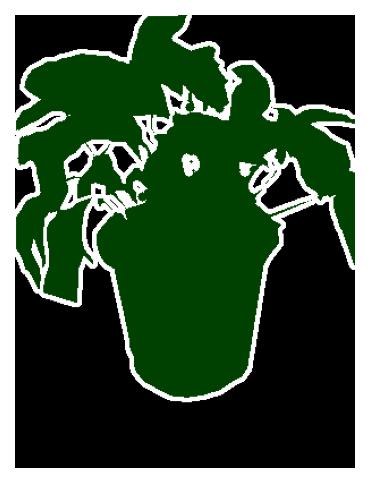}
    \end{subfigure}
    \begin{subfigure}{0.085\linewidth}
        \centering
        \includegraphics[width=\linewidth]{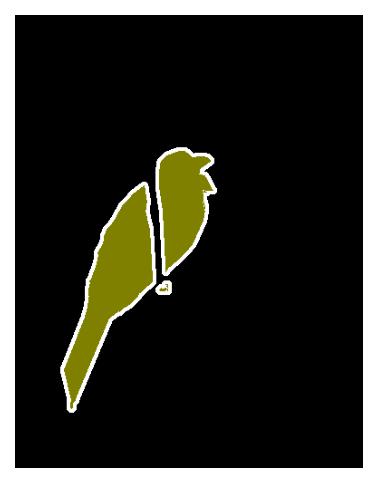}
    \end{subfigure}
    \begin{subfigure}{0.075\linewidth}
        \centering
        \includegraphics[width=\linewidth]{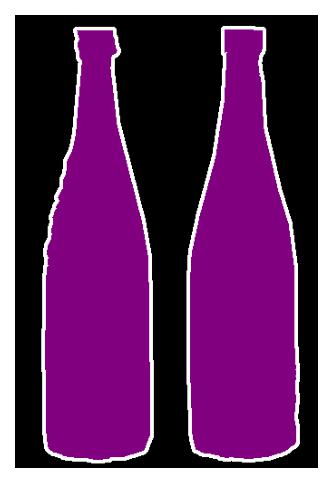}
    \end{subfigure}
    \begin{subfigure}{0.135\linewidth}
        \centering
        \includegraphics[width=\linewidth]{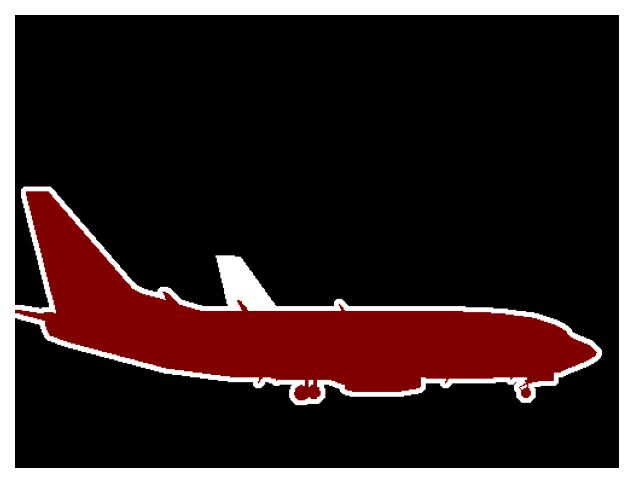}
    \end{subfigure}

    \begin{subfigure}{0.135\linewidth}
        \centering
        \includegraphics[width=\linewidth]{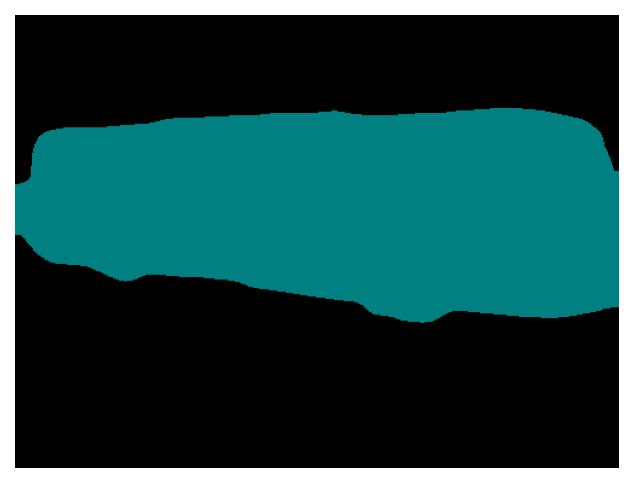}
    \end{subfigure}
    \begin{subfigure}{0.135\linewidth}
        \centering
        \includegraphics[width=\linewidth]{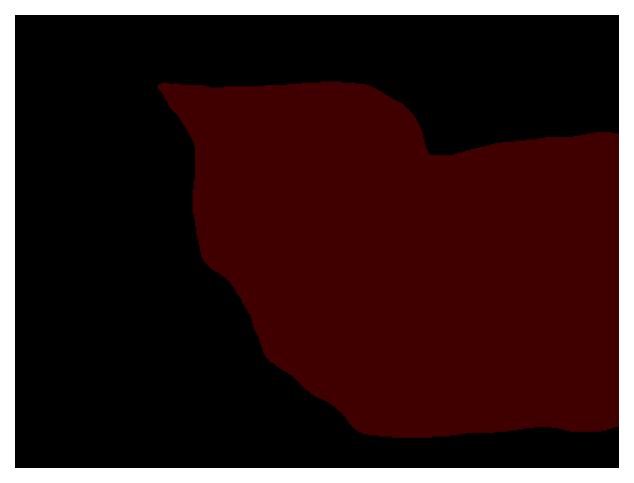}
    \end{subfigure}
    \begin{subfigure}{0.135\linewidth}
        \centering
        \includegraphics[width=\linewidth]{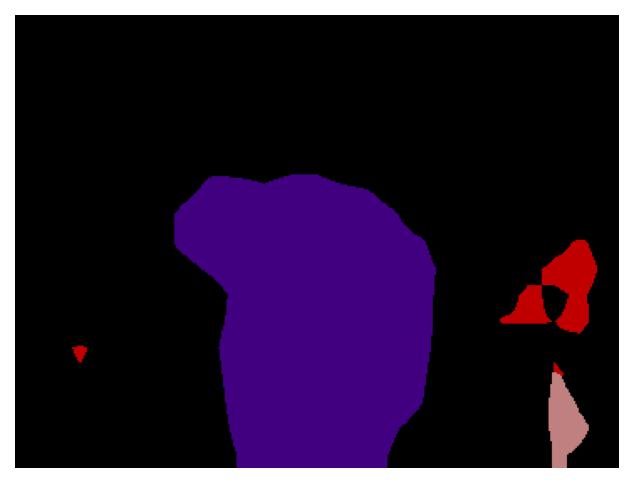}
    \end{subfigure}
    \begin{subfigure}{0.135\linewidth}
        \centering
        \includegraphics[width=\linewidth]{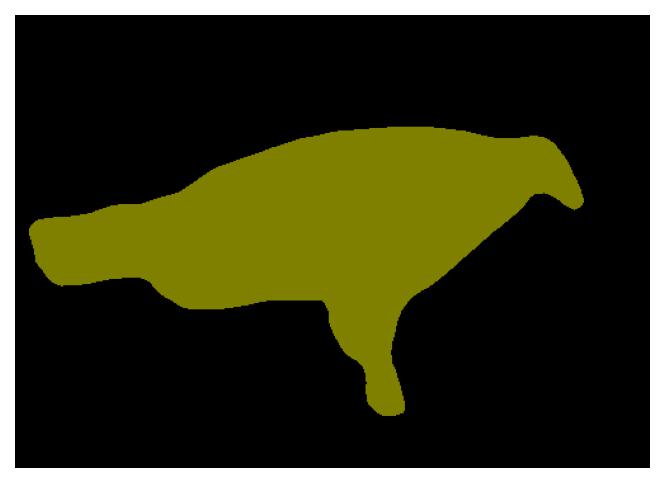}
    \end{subfigure}
    \begin{subfigure}{0.085\linewidth}
        \centering
        \includegraphics[width=\linewidth]{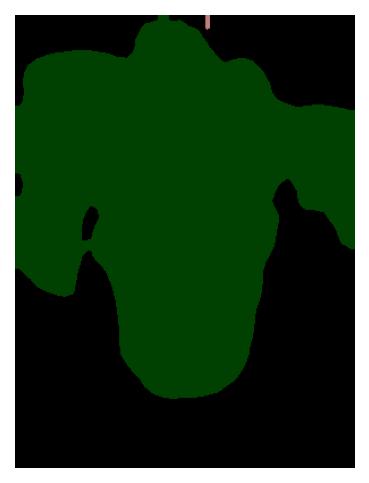}
    \end{subfigure}
    \begin{subfigure}{0.085\linewidth}
        \centering
        \includegraphics[width=\linewidth]{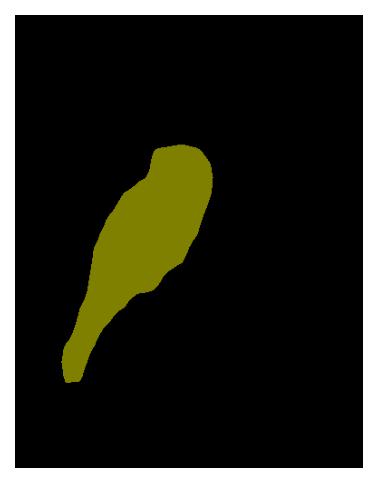}
    \end{subfigure}
    \begin{subfigure}{0.075\linewidth}
        \centering
        \includegraphics[width=\linewidth]{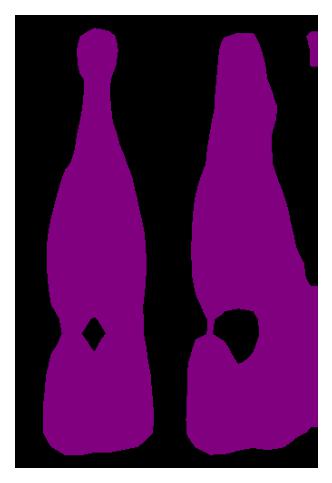}
    \end{subfigure}
    \begin{subfigure}{0.135\linewidth}
        \centering
        \includegraphics[width=\linewidth]{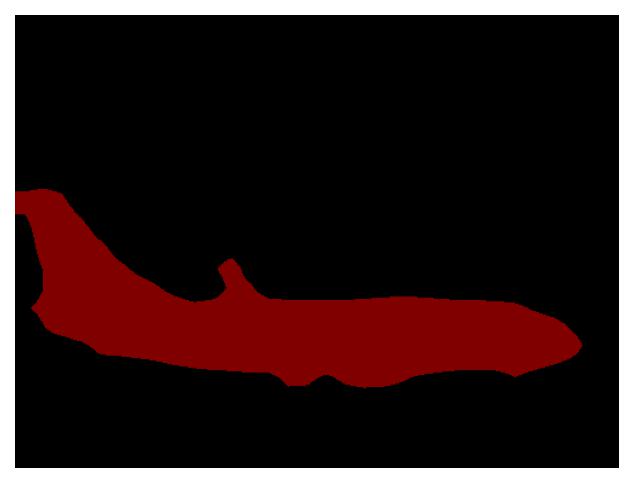}
    \end{subfigure}

    \caption{\textbf{Additional qualitative results on zero-shot semantic segmentation on Pascal VOC.} The first row shows the original images, the second row shows the corresponding ground-truth labels and the third row shows the predictions from our best performing model, i.e., PACL trained using a pre-trained CLIP ViT-B/16 encoder on GCC-3M + GCC-12M + YFCC-15M.}
    \vspace{-4mm}
    \label{fig:app_segmentation_voc}
\end{figure}

\begin{figure}[!t]
    \centering
    \begin{subfigure}{0.12\linewidth}
        \centering
        \includegraphics[width=\linewidth]{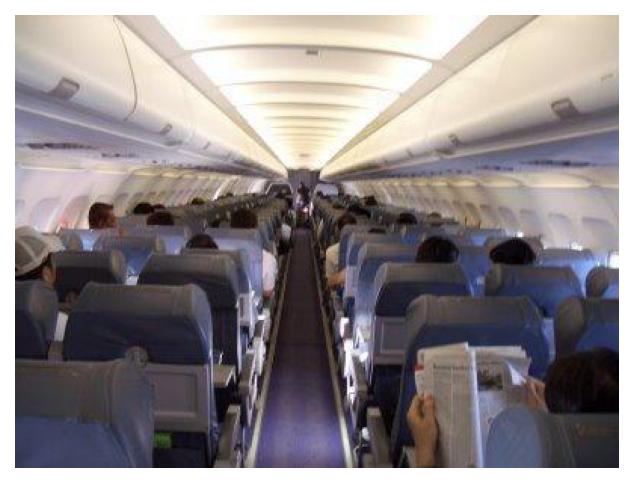}
    \end{subfigure}
    \begin{subfigure}{0.12\linewidth}
        \centering
        \includegraphics[width=\linewidth]{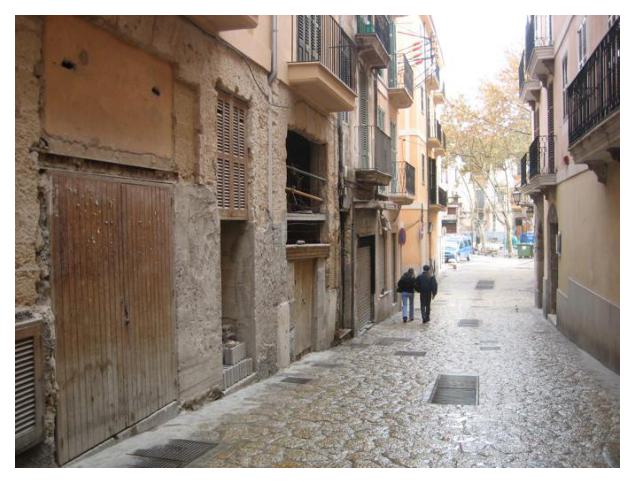}
    \end{subfigure}
    \begin{subfigure}{0.12\linewidth}
        \centering
        \includegraphics[width=\linewidth]{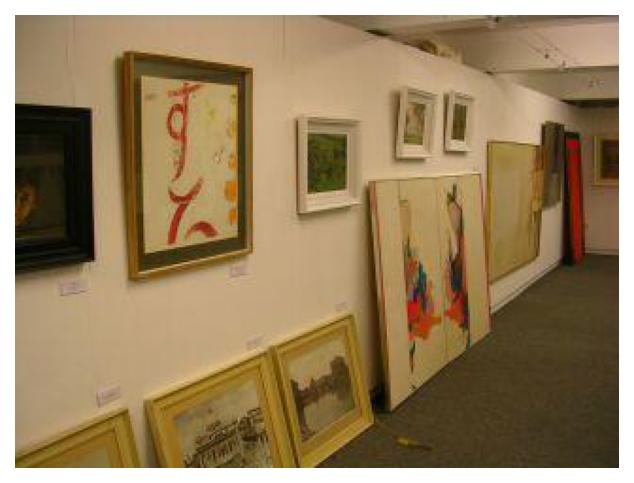}
    \end{subfigure}
    \begin{subfigure}{0.15\linewidth}
        \centering
        \includegraphics[width=\linewidth]{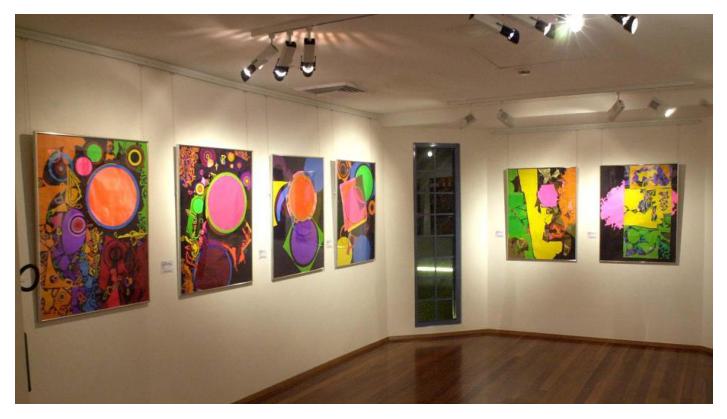}
    \end{subfigure}
    \begin{subfigure}{0.135\linewidth}
        \centering
        \includegraphics[width=\linewidth]{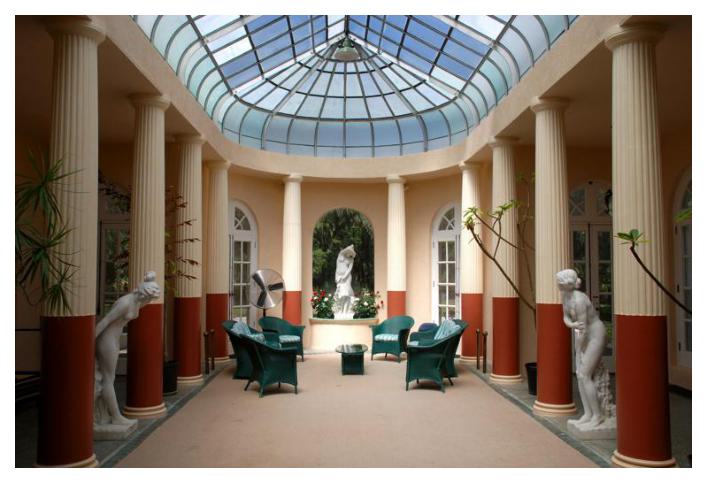}
    \end{subfigure}
    \begin{subfigure}{0.12\linewidth}
        \centering
        \includegraphics[width=\linewidth]{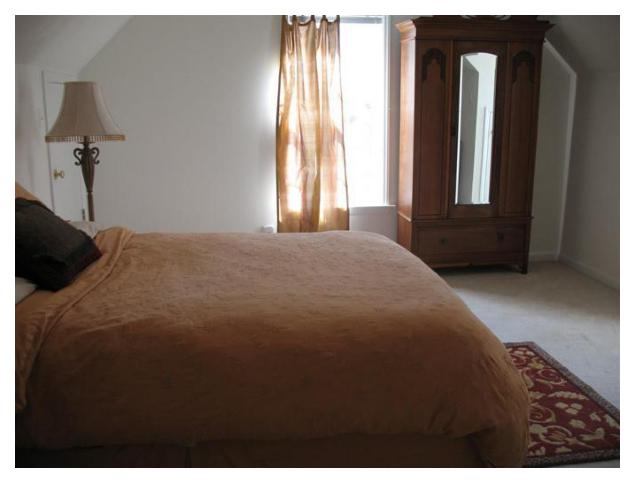}
    \end{subfigure}
    \begin{subfigure}{0.16\linewidth}
        \centering
        \includegraphics[width=\linewidth]{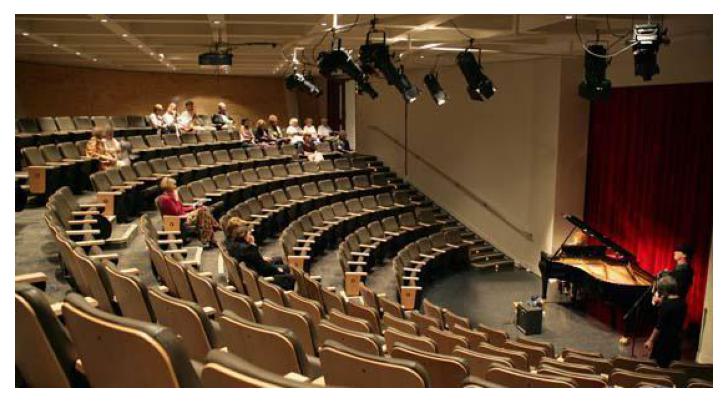}
    \end{subfigure}

    \begin{subfigure}{0.12\linewidth}
        \centering
        \includegraphics[width=\linewidth]{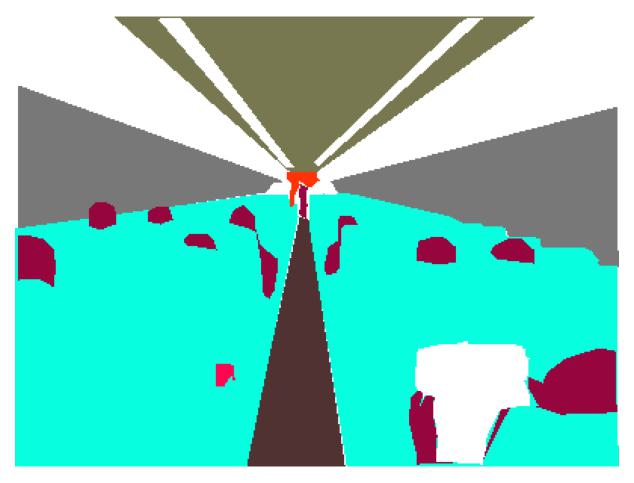}
    \end{subfigure}
    \begin{subfigure}{0.12\linewidth}
        \centering
        \includegraphics[width=\linewidth]{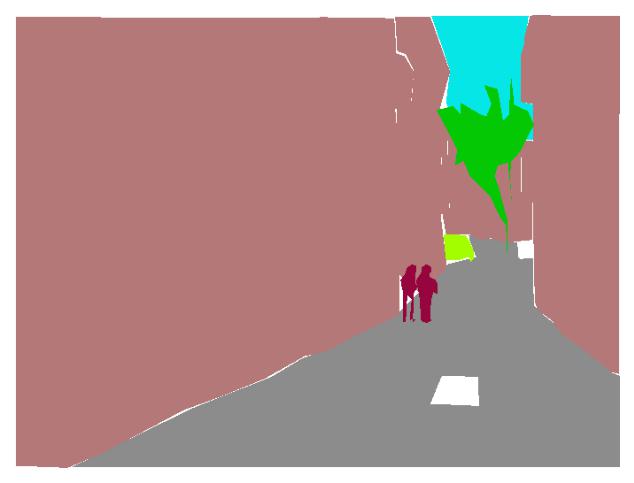}
    \end{subfigure}
    \begin{subfigure}{0.12\linewidth}
        \centering
        \includegraphics[width=\linewidth]{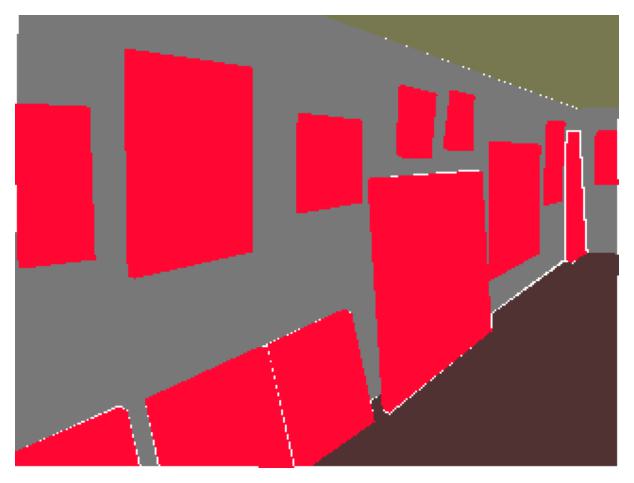}
    \end{subfigure}
    \begin{subfigure}{0.15\linewidth}
        \centering
        \includegraphics[width=\linewidth]{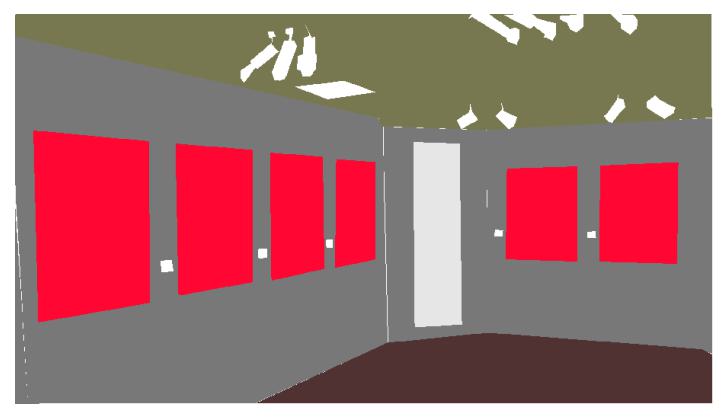}
    \end{subfigure}
    \begin{subfigure}{0.135\linewidth}
        \centering
        \includegraphics[width=\linewidth]{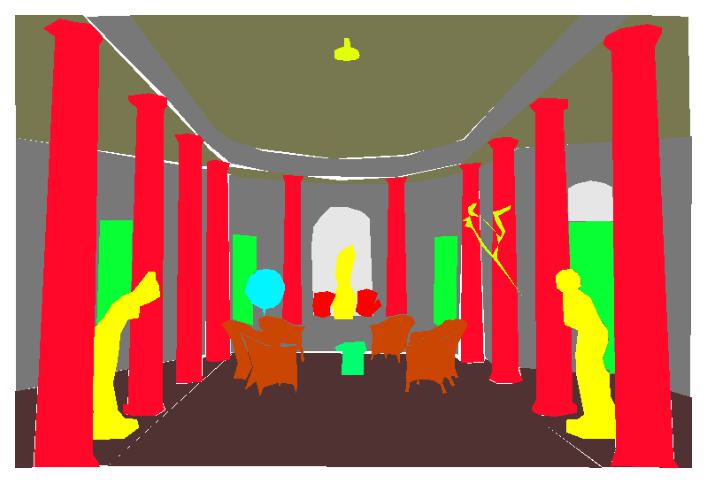}
    \end{subfigure}
    \begin{subfigure}{0.12\linewidth}
        \centering
        \includegraphics[width=\linewidth]{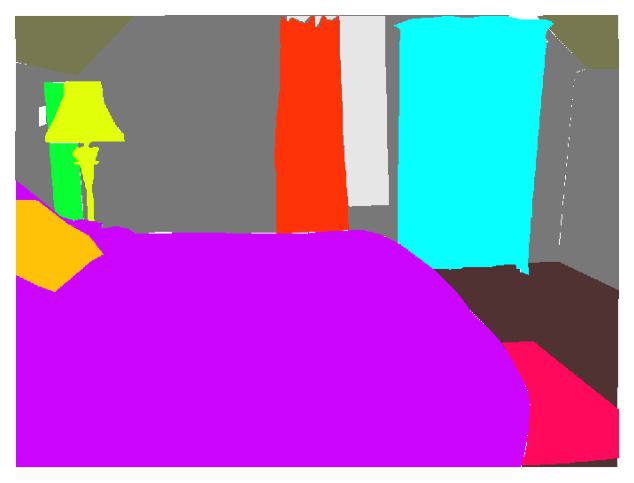}
    \end{subfigure}
    \begin{subfigure}{0.16\linewidth}
        \centering
        \includegraphics[width=\linewidth]{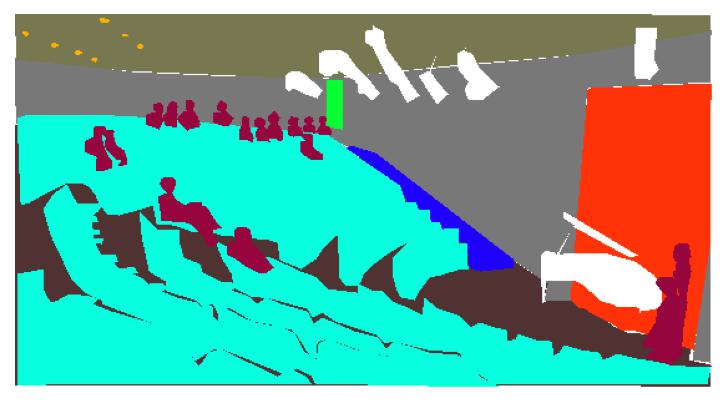}
    \end{subfigure}

    \begin{subfigure}{0.12\linewidth}
        \centering
        \includegraphics[width=\linewidth]{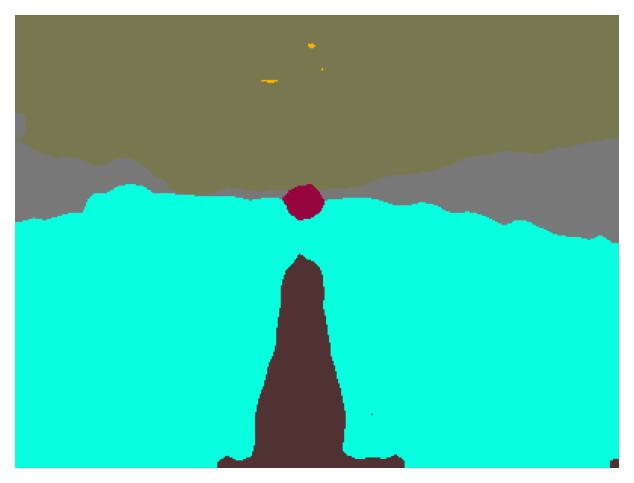}
    \end{subfigure}
    \begin{subfigure}{0.12\linewidth}
        \centering
        \includegraphics[width=\linewidth]{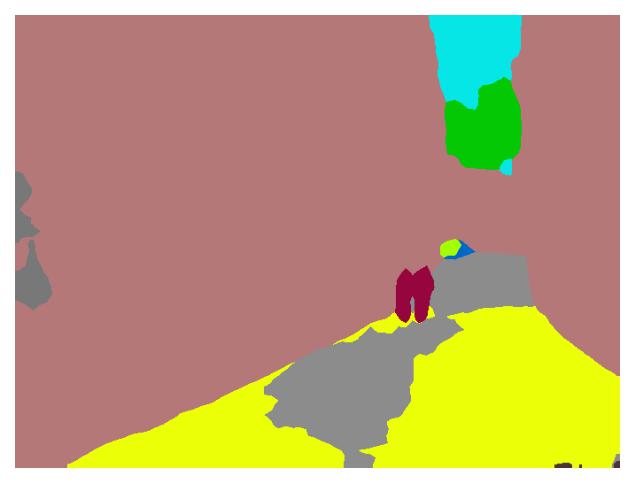}
    \end{subfigure}
    \begin{subfigure}{0.12\linewidth}
        \centering
        \includegraphics[width=\linewidth]{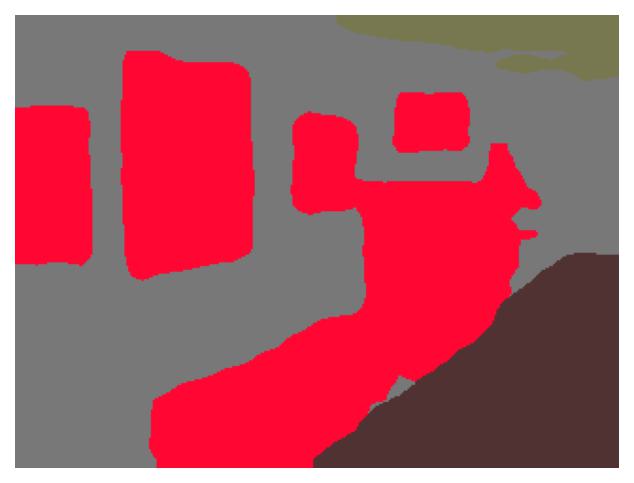}
    \end{subfigure}
    \begin{subfigure}{0.15\linewidth}
        \centering
        \includegraphics[width=\linewidth]{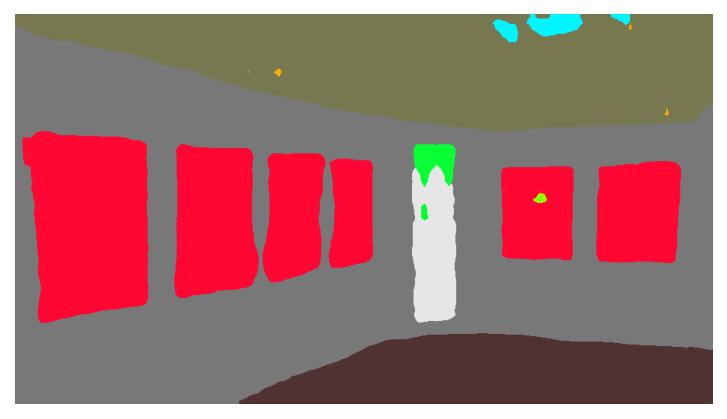}
    \end{subfigure}
    \begin{subfigure}{0.135\linewidth}
        \centering
        \includegraphics[width=\linewidth]{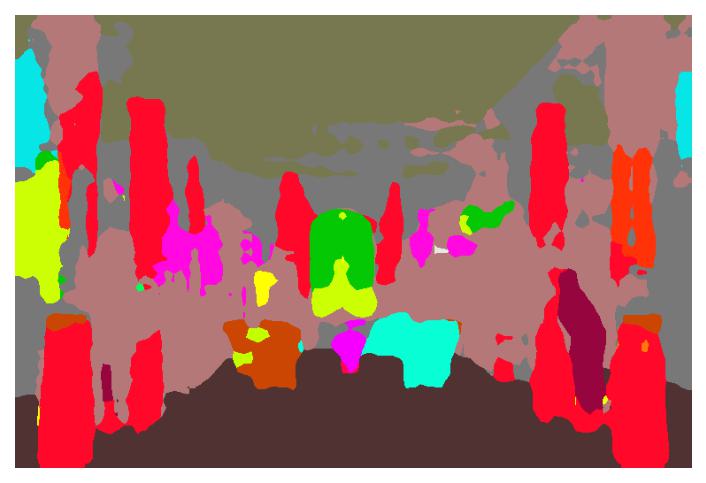}
    \end{subfigure}
    \begin{subfigure}{0.12\linewidth}
        \centering
        \includegraphics[width=\linewidth]{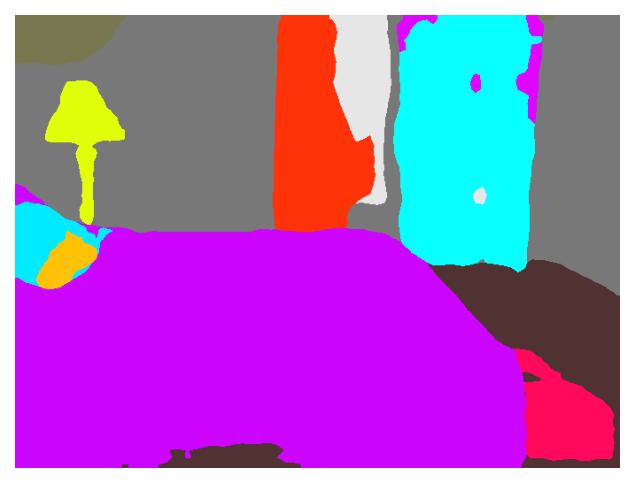}
    \end{subfigure}
    \begin{subfigure}{0.16\linewidth}
        \centering
        \includegraphics[width=\linewidth]{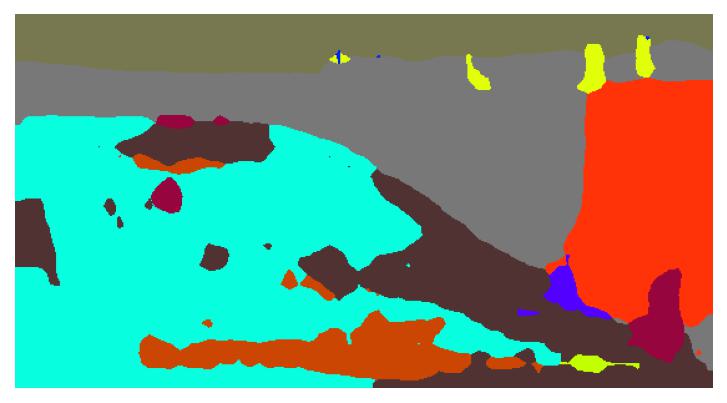}
    \end{subfigure}

    \caption{\textbf{Additional qualitative results on zero-shot semantic segmentation on ADE-20K.} The first row shows the original images, the second row shows the corresponding ground-truth labels and the third row shows the predictions from our best performing model, i.e., PACL trained using a pre-trained CLIP ViT-B/16 encoder on GCC-3M + GCC-12M + YFCC-15M.}
    \vspace{-4mm}
    \label{fig:app_segmentation_ade20k}
\end{figure}

In \Cref{fig:zeroshot_segmentation_qualitative}, we showed qualitative results for the task of zero-shot semantic segmentation on both Pascal VOC and ADE20K datasets. In this section, we present more qualitative results on the same. Particularly, in \Cref{fig:app_segmentation_voc}, we show additional qualitative results on 8 images from Pascal VOC covering different concepts including bus, cat, dog, bird, potted plant, bottle and plane. Similarly, in \Cref{fig:app_segmentation_ade20k}, we provide qualitative segmentation results from various indoor and outdoor scenes from ADE20K. Similar to our observations in the main paper, we find that the zero-shot segmentation results are decent and our models can recognise a large variety of concepts without ever having been trained on segmentation annotations or masks for any of them. This shows the potential of using the scale of large image-text datasets for zero-shot transfer to semantic segmentation. 

\subsection{Zero-shot Image Classification}
\label{app:zeroshot_image_classification}

In \Cref{fig:diff_plot_classification} of the main paper, we showed the difference in zero-shot classification accuracies for PACL models trained with CLIP backbones as compared to vanilla CLIP models on a suite of 12 image classification tasks including ImageNet, 4 datasets considered to be distribution shifts on ImageNet and 7 other well-known image classification datasets. In this section, we provide the exact classification accuracies for all the models on each of the datasets. We present these results in \Cref{table:zeroshot_image_classification}. As mentioned in the main paper, the PACL models outperform vanilla CLIP on 10 out of 12 datasets for the ViT-B/16 model and 7 out of 12 datasets for the ViT-L/14 backbone, thereby broadly outperforming CLIP on zero-shot image classification.

\end{document}